\documentclass{article}

\usepackage{microtype}
\usepackage{graphicx}
\usepackage{subfigure}
\usepackage{booktabs} 

\usepackage{hyperref}

\usepackage[accepted]{icml2019}
\usepackage{amsmath}
\usepackage{amssymb}
\usepackage{mathtools}
\usepackage{breqn}
\usepackage{todonotes}
\usepackage{amsthm} 
\usepackage{placeins}

\renewcommand{\S}{\mathcal{S}}
\newcommand{\A}{\mathcal{A}}
\newcommand{\T}{\mathcal{T}}

\newcommand{\Real}{\mathbb{R}}

\newcommand{\E}{\mathbb{E}}

\newcommand{\deq}{\mathrel{\mathop{:}}=}
\newcommand{\commentout}[1]{}

\hypersetup{
	plainpages=false,
		colorlinks=true,              	linkcolor=blue,               	anchorcolor=blue,             	citecolor=blue,               	filecolor=blue,               	pagecolor=blue,               	urlcolor=blue,                	pdfview=FitH,                 	pdfstartview=FitH,            	pdfpagelayout=SinglePage      }

\newtheorem{theorem}{Theorem}

\icmltitlerunning{TD ($\Delta$) }

\begin{document}

\twocolumn[
\icmltitle{Separating value functions across time-scales}

\icmlsetsymbol{equal}{*}

\begin{icmlauthorlist}
\icmlauthor{Joshua Romoff}{equal,mg,fb}
\icmlauthor{Peter Henderson}{equal,sf}
\icmlauthor{Ahmed Touati}{udem,fb}
\icmlauthor{Emma Brunskill}{sf}
\icmlauthor{Joelle Pineau}{mg,fb}
\icmlauthor{Yann Ollivier}{fb}
\end{icmlauthorlist}

\icmlaffiliation{mg}{MILA, McGill University}
\icmlaffiliation{fb}{Facebook AI Research}
\icmlaffiliation{sf}{Stanford University}
\icmlaffiliation{udem}{MILA, Universit\'e de Montr\'eal}
\icmlcorrespondingauthor{Joshua Romoff}{joshua.romoff@mail.mcgill.ca}
\icmlcorrespondingauthor{Peter Henderson}{phend@stanford.edu}

\icmlkeywords{Machine Learning, ICML}

\vskip 0.3in
]

\printAffiliationsAndNotice{\icmlEqualContribution} 
\begin{abstract}
In many finite horizon episodic reinforcement learning (RL) settings, it is desirable to optimize for the undiscounted return --
in settings like Atari, for instance, the goal is to collect the most points while staying alive in the long run. 
Yet, it may be difficult (or even intractable) mathematically to learn with this target. As such, temporal discounting is often applied to optimize over a shorter effective planning horizon.
This comes at the risk of potentially biasing the optimization target away from the undiscounted goal. 
In settings where this bias is unacceptable -- where the system \emph{must} optimize for longer horizons 
at higher discounts 
-- the target of the value function approximator may increase in variance leading to difficulties in learning.
We present an extension of temporal difference (TD) learning, which we call TD($\Delta$), that breaks down a value function into a series of components based on the differences between value functions with smaller discount factors. 
The separation of a longer horizon value function into these components has useful properties in scalability and performance. 
We discuss these properties and show theoretic and empirical improvements over standard TD learning in certain settings.
\end{abstract}

\section{Introduction}
\label{intro}

The goal of reinforcement learning (RL) algorithms is to learn a policy that optimizes the cumulative reward (return) provided by the environment. 
A discount factor $0 \leq \gamma < 1$ can be used to optimize an exponentially decreasing function of the future return.
Discounting is often used as a biased proxy for optimizing the cumulative reward to reduce variance and make use of convenient theoretical convergence properties, making learning more efficient and stable~\cite{bertsekas1995neuro,prokhorov1997adaptive,even2003learning}.
However, in many of the complex tasks used for evaluating current state-of-the-art reinforcement learning systems  ~\cite{mnih2013playing,OpenAI_dota}, it is more desirable to optimize for performance over long horizons. The optimal choice of discount factor, which balances asymptotic policy performance with learning ability, is often difficult, and solutions have ranged from scheduled curricula~\cite{OpenAI_dota,prokhorov1997adaptive,franccois2015discount} to meta-gradient learning of the discount factor~\cite{xu2018meta}.

\citet{OpenAI_dota}, for example, start with a small discount factor and gradually increase it to bootstrap the learning process. 
Rather than explicitly tackling the problem of discount selection, we make the observation that for any arbitrary discount factor, the discounted value function already encompasses all smaller time-scales (discounts). This simple observation allows us to derive a novel method of generating separable value functions. That is, we can separate the value function into a number of partial estimators, which we call delta estimators, which approximate the difference $W_z=V_{\gamma_z}-V_{\gamma_{z-1}}$ between value functions. Importantly, each delta estimators is learnable by itself, because it satisfies a Bellman-like equation based on the $W$s of shorter horizons. 
Thus, these delta estimators can then be summed to yield the same discounted value function, and any subset of estimators from the series of smaller $\gamma_z$ values.
The use of difference methods (the delta between two value functions at different s) leads us to call our method TD($\Delta$).

The separable nature of the full TD($\Delta$) estimator allows for each component to be learned in a way that is optimal for that part of the overall value function. This means that, for example, the learning rate (and similarly other parameters) can be adjusted for each component, yielding overall faster convergence. 
Moreover, the components corresponding to smaller effective horizons can converge faster, bootstrapping larger horizon components (at the risk of some bias).
Our method provides a simple drop-in way to separate value functions in any TD-like algorithm to increase performance in a variety of settings, particularly in MDPs with dense rewards. 

We provide an intuitive method for setting intermediary $\gamma$ values which yields performance gains, in most cases, without additional tuning. Yet, we also show that this method affords the option of further fine-tuning for further performance improvement and note that our method is compatible with adaptive $\gamma$ selection methods \cite{xu2018meta}. 
We demonstrate these benefits theoretically and highlight performance gains in a simple ring MDP -- used by \citet{kearns2000bias} for a similar bias-variance analysis -- by adjusting the $k$-step returns used to update each delta estimator. 
We also show how this method can be combined with TD($\lambda$)~\cite{sutton1984temporal} and Generalized Advantage Estimation (GAE)~\cite{schulman2015high}, leading to empirical gains in dense reward Atari games. 

\section{Related work}
\label{sec:related}

Many recent works approach discount factors in different ways. To our knowledge, the closest work to our own is that of \citet{fedus2019hyper}, \citet{sherstan2018generalizing}, \citet{sutton2011horde}, \citet{sutton1995td}, \citet{feinberg1994markov}, and \citet{reinke2017average}, which learn ensembles of value functions at different s to form a generalized value function. In the case of \citet{reinke2017average}, they do so for imitating the average return estimator. \citet{feinberg1994markov} examine an optimal policy for the mixture of two value functions with different discount factors. Similarly, \citet{sutton1995td} present learning value functions across different levels of temporal abstraction through mixing functions. In the case of \citet{sherstan2018generalizing} and \citet{sutton2011horde}, they train a value function such that it can be queried for a given set of time-scales. Finally, concurrent to this work, \citet{fedus2019hyper} re-weight multiple value functions across different discount factors to form a hyperbolic value function. However, we note that none of the aforementioned works utilize short term estimates to train the longer term value functions. Thus, while our method can similarly be used as a generalized value function, the ability to query smaller time-scales is a side-benefit to the performance increases yielded by separating value functions into different time-scales via TD($\Delta$). 

Some recent work has investigated how to precisely select the discount factor choice~\cite{franccois2015discount,xu2018meta}. \citet{franccois2015discount} suggest a particular scheduling mechanism, seen similarly in \citet{OpenAI_dota} and \citet{prokhorov1997adaptive}. \citet{xu2018meta} propose a meta-gradient approach which learns the discount factor (and $\lambda$ value) over time.  All of these methods can be applied to our own as we do not necessarily prescribe a final overall $\gamma$ value to be used. 

Finally, another broad category of work relates to our own in a somewhat peripheral way. Indeed, hierarchical reinforcement learning methods often decompose value functions or reward functions into a number of smaller systems which can be optimized somewhat separately~\cite{dietterich2000hierarchical,henderson2018optiongan,hengst2002discovering,reynolds1999decision,menache2002q,russell2003q, vanseijen2017hra}. These works learn hierarchical policies, paired with the decomposed value functions, which reflect the structure of the goals.

\section{Background and notation}

Consider a fully observable Markov Decision Process (MDP) ~\cite{bellman1957markovian} $(\S, \A, P, r)$ with state space $\S$, action space $\A$, transition probabilities $P : \S \times \A \rightarrow (\S \rightarrow [0,1])$ mapping state-action pairs to distributions over next states, and reward function $r : (\S \times \A) \rightarrow \Real$. At every timestep $t$, an agent is in a state $s_t$, can take an action $a_t$, receive a reward $r_t = r(s_t, a_t)$, and transition to its next state in the system $s_{t+1} \sim P(\cdot \mid s_t, a_t)$.

In the usual MDP setting, an agent optimizes the discounted return: $V_\gamma^\pi(s) = \left[\sum_{t=0}^\infty \gamma^{t} r_t | s_0 = s, \pi \right]$, where $\gamma$ is the discount factor and $\pi : \S \rightarrow (\A \rightarrow [0, 1])$ is the policy that the agent follows.
$V_\gamma^\pi$ can be obtained as the fixed point of the Bellman operator over the action-value function $\T^\pi V^\pi = r^\pi + \gamma P^\pi V^\pi$ where $r^\pi$ and $P^\pi$ are respectively the expected immediate reward and transition probabities operator induced by the policy $\pi$. In the rest of the paper, we drop the superscript $\pi$ to avoid clutter in the formulas.

The value estimate, $\hat V_\gamma$ may approximate the true value function $V_\gamma$ via temporal difference (TD) learning~\citep{sutton1984temporal}. Given a transition $(s_t, a_t, r_t, s_{t+1})$ we can update our value function using the one-step TD error: $ \delta_{t}^\gamma=r_{t} + \gamma \hat{V}_\gamma(s_{t+1}) -\hat{V}_\gamma(s_{t})$. 
Alternatively, 
given an entire trajectory, we can instead use the discounted sum of one-step TD errors, which is commonly referred to as either the $\lambda$-return \cite{sutton1984temporal} or equivalently the generalized advantage estimator (GAE) \cite{ schulman2015high}: 
\begin{equation}
\label{eq:gae}
A(s_t) = \sum_{k=0}^{\infty} (\lambda \gamma)^k \delta_{t+k}^\gamma,
\end{equation}
where the $\lambda$ controls the bias-variance trade-off.

With function approximation we use a parameterized value function $\hat V_\gamma(\cdot;\theta)$ and then update our value function via the following loss:
\begin{equation}
\label{eq:ppovalue}
    \mathcal{L} (\theta) = \mathbb{E}\left[ \left( \hat V_\gamma(s; \theta) - \left (\hat V_\gamma(s) +  A(s) \right) \right)^2 \right].
\end{equation}
In actor-critic methods~\cite{sutton2000policy, konda2000actor, mnih2016asynchronous}, the value function is updated per equation~\ref{eq:ppovalue}, and a stochastic parameterized policy (actor, $\pi_\omega(a | s)$) is learned from this value estimator via the advantage function where the loss is: 
\begin{equation}
    \mathcal{L}(\omega) = \mathbb{E} \left[- \log \pi(a ,s; \omega) A (s) \right].
\end{equation}
Building on top of actor-critic methods, Proximal Policy Optimization (PPO)~\cite{schulman2017proximal} constrains the policy update to a given optimization region (a trust region) in the form of a clipping objective between the current and old parameters, $\omega$ and $\omega_{old}$:
\begin{equation}
     \mathcal{L} (\omega) = \mathbb{E} \left[ \min \left ( \rho(\omega) A(s), \psi (\omega) A(s)\right)\right], 
     \label{eq:ppo}
\end{equation}
where $\rho = \frac{\pi_\omega (a | s)}{\pi_{\omega{old}}(a|s)}$ is the likelihood ratio, $\psi(\omega) = \text{clip} \left (\rho , 1- \epsilon, 1+\epsilon \right )$ is the clipped likelihood ratio, and $\epsilon < 1$ is some small factor applied to constrain the update.

\section{TD($\Delta$)}

In this section, we introduce TD($\Delta$), along with several variations, including: Multi-step TD, TD($\lambda$), and GAE.
\subsection{Single-step TD($\Delta$)}
\label{sec:singleTD}
Consider learning with $Z+1$ different discount factors
$\Delta \deq \gamma_0,\gamma_1,\ldots,\gamma_{Z}$. Each of these define a corresponding value function $V_{\gamma_z}$. We define the \emph{delta functions} $W_z$ by
\begin{equation}
W_z\coloneqq V_{\gamma_z}-V_{\gamma_{z-1}}, \qquad W_0\coloneqq V_{\gamma_0}.
\end{equation}
This results in $Z+1$ delta functions such
that the desired $V_{\gamma_z}$ is simply the sum of the delta functions:
\begin{equation}
V_{\gamma_z}(s)=\sum_{i=0}^{z} W_i(s).
\end{equation}

We can derive a Bellman-like equation for the delta functions $W$.
Indeed, $W_0=V_0$ satisfies the Bellman equation
\begin{equation}
W_0 (s_t ) = \mathbb{E} \left [r_t + \gamma_0W_0 (s_{t+1} )\right ],
\end{equation}
while the delta functions at larger s satisfy:
\begin{align}
\label{eq:onestep}
&W_z(s_t)  = V_{\gamma_z}(s_t)-V_{\gamma_{z-1}}(s_t)  \nonumber\\
& = \mathbb{E} \biggl[ \left(r_t+\gamma_z
V_{\gamma_z}(s_{t+1})\right)  - \left ( r_t+\gamma_{z-1}V_{\gamma_{z-1}}(s_{t+1})\right)\biggr]  \nonumber\\
& = \mathbb{E} \biggl[\gamma_z
\left(W_z(s_{t+1})+V_{\gamma_{z-1}}(s_{t+1})\right) -\gamma_{z-1}V_{\gamma_{z-1}}(s_{t+1}) \biggr]  \nonumber\\
& = \mathbb{E} \biggl[(\gamma_z-\gamma_{z-1})
V_{\gamma_{z-1}}(s_{t+1})+\gamma_z W_z(s_{t+1})\biggr].
\end{align}
This is a Bellman-type equation for $W_z$, with decay factor $\gamma_z$
and rewards $V_{\gamma_{z-1}}(s_{t+1})$. Thus, we can use it to define
the expected TD update for $W_z$.
Note that in this expression,
$V_{\gamma_{z-1}}(s_{t+1})$ can be expanded as the sum of $W_i(s_{t+1})$
for $i\leq z-1$, so that the Bellman equation for $W_z$ depends on the values
of all delta functions $W_i$, $i\leq z-1$.

This way, the delta value function at a given time-scale appears as an
autonomous reinforcement learning problem with rewards coming from the
value function of the immediately lower time-scale.
Thus, for a target discounted value function $V_{\gamma_z}(s)$, we
can train all the delta components in parallel according to
this TD update, bootstrapping off of the old value of all the
estimators. Of course, this requires assuming a sequence of $\gamma_z$ values, including a largest and smallest discount $\gamma_0$ and $\gamma_Z$. We will see in Section~\ref{sec:tuning} that these can affect results, further allowing tuning. However, to avoid the addition of a number of hyperparameters, we assume a simple sequence where we double the effective horizon of the $\gamma_z$ values until the final $\gamma_Z$ value is reached. This simple sequence of $\gamma$'s, without tuning, yields performance gains in many settings as seen in Section~\ref{sec:deepexps}. 

\subsection{Multi-step TD($\Delta$)}
\label{sec:multiTD}
In many scenarios, it has been shown that multi-step TD is more efficient than single-step TD \cite{sutton1998introduction}. We can easily extend TD($\Delta$) to the multi-step case as follows. To begin, since $W_{0}\coloneqq V_{\gamma_0}$, the multi-step target for $W_0$ is identical to the standard multi-step target with $\gamma=\gamma_0$. For all other $W$s, we can unroll both the bootstrap term and the rewards from the previous value function in Section~\ref{sec:singleTD}: 
\begin{align}
    &W_0(s_t) = \mathbb{E} \biggl [ \sum_{i=0}^{k_0-1} \gamma_0^{i}  r_{t+i} + \gamma_0^{k_0} W_0 (s_{t+k})\biggr ], \nonumber \\
\label{eq:multistepupdate}
&W_z(s_t)  =\mathbb{E} \biggl [ (\gamma_z - \gamma_{z-1}) V_{\gamma_{z-1}}(s_{t+1}) + \gamma_z W_{z}(s_{t+1}) \biggr ] \nonumber\\
    & = \mathbb{E} \biggl[ (\gamma_z - \gamma_{z-1}) r_{t+1} + \gamma_{z-1} (\gamma_z - \gamma_{z-1}) V_{\gamma_{z-1}}(s_{t+2}) \nonumber\\
    & \quad + \gamma_z (\gamma_z - \gamma_{z-1}) V_{\gamma_{z-1}}(s_{t+2}) + \gamma_z^2 W_{z}(s_{t+2}) \biggr] \nonumber\\
    & = \mathbb{E} \biggl [ (\gamma_z - \gamma_{z-1}) r_{t+1} +  (\gamma_z^2 - \gamma_{z-1}^2) V_{\gamma_{z-1}}(s_{t+2})  \nonumber\\
    & \quad + \gamma_z^2 W_{z}(s_{t+2}) \biggr ]  \nonumber \\
        & = \mathbb{E} \biggl [\sum_{i=1}^{k_z-1} ( \gamma_z^{i} - \gamma_{z-1}^{i}) r_{t+i} + ( \gamma_z^{k_z} - \gamma_{z-1}^{k_z}) V_{\gamma_{z-1}}(s_{t+k})  \nonumber \\
    & \quad +  \gamma_z^{k_z} W_z (s_{t+k}) \biggr] .
        \end{align}
Thus, each $W_z$ receives a fraction of the rewards from the environment up to time-step $k_z-1$. Additionally, each $W$ bootstraps off of its own value function as well as the value at the previous time-scale. A version of this algorithm based on $k$-step bootstrapping from \citet{sutton1998introduction} can be seen in Algorithm~\ref{alg:tab}. We also note that while Alorithm~\ref{alg:tab} has quadratic complexity w.r.t. $Z$, we can make the algorithm linear in implementation for large $Z$ by storing $\hat{V}$ values at each time-scale $\gamma_z$.

\begin{algorithm}[t]
  \caption{Multi-step TD($\Delta$)}
  \label{alg:tab}
\begin{algorithmic}
  \STATE Inputs $\left(\gamma_0, \gamma_1, ..., \gamma_Z\right)$, $\left(k_0, k_1, ..., k_Z\right)$, $\left(\alpha_0, \alpha_1, ..., \alpha_Z\right)$
  \STATE Initialize $\hat W_z(\cdot) = 0 \quad \forall z $
    \FOR{$t=0,1,2...$}
  \STATE Take step according to policy and store $(s_t, r_t, s_{t+1})$
  \IF{$t \geq k_Z$}
  \STATE $\tau \leftarrow t - k_Z + 1$
  \FOR{$z \in 0, 1, ..., Z$}
  \IF{$z=0$}
  \STATE $G^0_\tau \leftarrow \sum_{i=\tau}^{\tau + k_0 - 1} \gamma_0^{i-\tau}  r_i + \gamma^{k_0}_0 \hat W_0(s_{\tau+k_0})$
  \ELSE
  \STATE $G^z_\tau \leftarrow \sum_{i=\tau+1}^{\tau + k_z - 1} (\gamma_z^{i-\tau} - \gamma_{z-1}^{i-\tau}) r_i +$ 
  \STATE $\quad ( \gamma^{k_z}_z - \gamma_{z-1}^{k_z}) \sum_{g=0}^{z-1} \hat W_{g} (s_{\tau+k_z}) +$
  \STATE $\quad \gamma^{k_z}_z \hat W_z(s_{\tau+k_z})$
  \ENDIF
  \ENDFOR
  \FOR{$z \in 0,1, ..., Z$}
  \STATE $\hat W_z(s_\tau) \leftarrow \hat W_z(s_\tau) + \alpha_z \left[ G^z_\tau - \hat W_z (s_\tau) \right]$
  \ENDFOR
  \ENDIF
  \ENDFOR
\end{algorithmic}
\end{algorithm}

\subsection{TD($\lambda,\Delta$)}
\label{sec:TD(lambda)}
The traditional TD($\lambda$)~\cite{sutton1984temporal} uses the following $\lambda$-return as a target for its update rules:
\begin{equation}\label{eq: TD(lambda) return}
    G^{\gamma, \lambda}_t = \hat{V}_\gamma(s_t) + \sum_{k=0}^\infty (\lambda \gamma)^{k} \delta^{\gamma}_{t+k}.
\end{equation}
The underlying TD($\lambda$) operator can be written: 
\begin{equation}\label{eq: TD operator}
    T_\lambda V = V + (I - \lambda \gamma P)^{-1}(T V -V) 
\end{equation}
Similarly, for each $W_z$ we can define a $\lambda$ return:
\begin{equation}\label{eq: TD(Delta) return}
    G^{z, \lambda_z}_t \deq \hat{W}_z(s_t) + \sum_{k=0}^\infty (\lambda_z \gamma_z)^{k} \delta^{z}_{t+k},
\end{equation}
where $\delta^{0}_{t} \deq \delta^{\gamma_0}_t$ and $\delta^{z}_{t} \deq (\gamma_z - \gamma_{z-1}) \hat{V}_{\gamma_{z-1}}(s_{t+1}) + \gamma_z \hat{W}_z(s_{t+1}) - \hat{W}_z(s_t)$ are the TD-errors.

\subsection{TD($\lambda, \Delta$) with GAE}
\label{sec:GAE}

Since GAE is used in powerful policy gradient baselines \cite{schulman2017proximal}, we propose a simple extension of TD($\Delta$) that leverages GAE. Specifically, to train the policy we use the following generalized advantage estimator:
\begin{equation}
\label{eq:deltagae}
    A^{\Delta} (s_t) \deq \sum_{k=0}^{T-1} (\lambda_Z \gamma_Z)^k \delta_{t+k}^{\Delta},
\end{equation}
where  $\delta_{t+k}^{\Delta} \coloneqq r_t + \gamma_Z \sum_{z=0}^Z \hat{W}_z(s_{t+1}) -  \sum_{z=0}^Z \hat{W}_z(s_{t})$. 

Thus, we use $\gamma_Z$ as our discount factor and the sum of all our $W$ estimators as a replacement for $V_{\gamma_Z}$.  This objective can easily be applied to PPO by using the policy update from Eq.~\ref{eq:ppo} and replacing $A$ with $A^{\Delta}$. Similarly, to train each $W_z$, we use a truncated version of their respective $\lambda$-return defined in Equation~\ref{eq: TD(Delta) return}. See Algorithm~\ref{alg:ppodelta} for details. 

\begin{algorithm}[!htbp]
  \caption{PPO-TD($\lambda, \Delta$)}
  \label{alg:ppodelta}
\begin{algorithmic}
\STATE Initialize policy $\omega$ and values $\theta^{z} \quad \forall z$
\FOR{$t$ = 0, 1, 2, \dots}
\STATE Take step according to $\pi_\omega$ and store $(s_t, a_t, r_t, s_{t+1})$
\IF{$t \geq T$}
\STATE $G^{z,\lambda_z} \leftarrow \hat{W}_z(s_{t-T}) + \sum_{k=0}^{T-1} (\lambda_z \gamma_z)^{k} \delta^{z}_{t-T+k} \quad \forall z$
\STATE $A^{\Delta} = \sum_{k=0}^{T-1} (\lambda_Z \gamma_Z)^k \delta_{t-T+k}^{\Delta}$
\STATE Update $\theta^{z}$ with TD (Eq.~\ref{eq:ppovalue}) using $G^z$  $\quad \forall z$ 
\STATE Update $\omega$ with PPO (Eq.~\ref{eq:ppo}) using $A^{\Delta}$
\ENDIF
\ENDFOR
\label{alg}
\end{algorithmic}
\end{algorithm}

\section{Analysis}
We now analyze our estimators more formally. The goal is that our estimator will provide favorable bias-variance trade-offs under some circumstances (as we shall see experimentally). To shed light on this,  
we first start by illustrating when our estimator is identical to the single estimator $\hat V_\gamma$ (Theorem~\ref{prop: equivalence linear}) which gives insight into the important quantities of our estimator that can determine when we may achieve benefits over the standard $\hat V_\gamma$ estimator. Then motivated by these results and prior work by \citet{kearns2000bias}, we bound the error of our estimator in terms of a variance and bias term (Theorem~\ref{prop: bound TD(Delta)}) that also yields insight into how to trade-off this quantities to achieve the best result. 

\subsection{Equivalence settings and improvement}
\label{sec:proofs}

In some cases, we can show that our TD($\Delta$) update and its variations are equivalent to the non-delta estimator $V_\gamma$ when recomposed into a value function. In particular, we focus here on linear function approximation of the form:
\begin{equation*}
 \hat V_\gamma(s) \deq \langle \theta^{\gamma},  \phi(s) \rangle \quad \text{and}
\quad \hat W_z(s) \deq \langle \theta^{z},  \phi(s) \rangle, \forall z
\end{equation*}
where $\theta^{\gamma}$ and $\{ \theta^{z} \}_{z}$ are weight vectors in $\Real^d$ and $\phi: \S \rightarrow \Real^d$ is a feature map from a state to a given $d$-dimensional feature space. Let $\theta^{\gamma}$ be updated using TD($\lambda$) as follows:
\begin{equation}\label{eq: liner TD(lambda)}
    \theta^{\gamma}_{t+1} = \theta^{\gamma}_{t} + \alpha \left(G^{\gamma, \lambda}_t - \hat V_\gamma(s_t)\right) \phi(s_t),
\end{equation}
where $G^{\gamma, \lambda}_t$ is the TD($\lambda$) return defined in equation \ref{eq: TD(lambda) return}.

Similarly, each $\hat W_z$ is updated using TD($\lambda_z$, $\Delta$) as follows:
\begin{equation}\label{eq: liner TD(lambda, Delta)}
    \theta^{z}_{t+1} = \theta^{z}_{t} + \alpha_z \left(G^{z, \lambda_z}_t - \hat W_z(s_t)\right) \phi(s_t),
\end{equation}
where $G^{z, \lambda_z}_t$ is TD($\Delta$) return defined in equation \ref{eq: TD(Delta) return}.
Here, $\alpha$ and $\{ \alpha_z \}_z$ are positive learning rates. The following theorem establishes the equivalence of the two algorithms.
\begin{theorem}\label{prop: equivalence linear}
If $\alpha_z = \alpha, \lambda_z \gamma_z = \lambda \gamma, \forall z$ and if we pick the initial conditions such that $\sum_{z=0}^{Z} \theta_0^{z} = \theta_0^{\gamma}$, then the iterates produced by TD($\lambda$) (Eq.~\ref{eq: liner TD(lambda)}) and TD($\lambda$, $\Delta$) (Eq.~\ref{eq: liner TD(lambda, Delta)}) with linear function approximation satisfy: 
\begin{equation}
    \sum_{z=0}^{Z} \theta_t^{z} = \theta_t^{\gamma}, \forall t,
\end{equation}
(The proof is provided in the Supplemental).
\end{theorem}
Note that the equivalence is achieved when $\lambda_z \gamma_z = \lambda \gamma, \forall z$. When $\lambda$ is close to $1$ and $\gamma_z < \gamma$, the latter condition implies that $\lambda_z = \frac{\lambda \gamma}{\gamma_z}$ could potentially be larger than one. One would conclude that the TD($\lambda_z$) could diverge. Fortunately, we show in the next theorem that the TD($\lambda$) operator defined in equation \ref{eq: TD operator} is a contraction mapping for $1 \leq \lambda < \frac{1+\gamma}{2\gamma}$ which implies that $\lambda \gamma < 1$. 

\begin{theorem}\label{prop: contraction}
$\forall \lambda \in [0, \frac{1+\gamma}{2\gamma}[$, the operator $T_\lambda$ defined as $T_\lambda V = V + (I - \lambda \gamma P)^{-1}(T V -V), \forall V \in \Real^{|\S|}$ is well defined. Moreover, $T_\lambda V$ is a contraction with respect to the max norm and its contraction coefficient is equal to $\frac{\gamma |1-\lambda|}{1-\lambda \gamma}$  
(The proof is provided in the Supplemental).
\end{theorem}
Similarly, we can consider learning each $W_z$ using $k_z$-step TD($\Delta$) instead of TD($\lambda$, $\Delta$). In this case, the analysis of Theorem~\ref{prop: equivalence linear} could be extended to show that with linear function approximation, standard multi-step TD and multi-step TD($\Delta$) are equivalent if  $k_z = k, \forall z$.  

However, we note that the equivalence with unmodified TD learning is the exception rather than the rule. For one, in order to achieve equivalence we require the same learning rate across every . This is a strong restriction as intuitively the shorter time-scales can be learned faster than the longer ones. Further, adaptive optimizers are typically used in the nonlinear approximation setting~\cite{henderson2018did,schulman2017proximal}. 
Thus, the effective rate of learning can differ depending on the properties of each delta estimator and its target. 
In principle, the optimizer can automatically adapt the learning to be different for the shorter and longer s.

Besides for the learning rate, such a decomposition allows for some particularly helpful properties not afforded to the non-delta estimator. In particular, every $W_z$ delta component need not use the same $k$-step return (or $\lambda$-return) as the non-delta estimator (or the higher $W_z$ components). Specifically, if $k_z < k_{z+1}, \forall z$ (or $\gamma_z\lambda_z < \gamma_{z+1}\lambda_{z+1} , \forall z$), then there is the possibility for variance reduction (at the risk of some bias introduction).

\subsection{Analysis for reducing $k_z$ values}
\label{sec:analysis}
To see intuitively how our method differs from the single estimator case, let us consider the tabular \textit{phased} version of k-step TD studied by  \citet{kearns2000bias}. In this setting, starting from each state $s \in \S$, we generate $n$ trajectories $\{s^{(j)}_0 =s, a_0, r_0, \ldots, s^{(j)}_k, a^{(j)}_k, r^{(j)}_k, s^{(j)}_{k+1}, \ldots \}_{1\leq j \leq n}$ following policy $\pi$. For each iteration $t$, called also phase $t$, the value function estimate for $s$ is defined as follows: 
\begin{equation}
    \hat V_{\gamma, t}(s) = \frac{1}{n} \sum_{j=1}^n \left( \sum_{i=0}^{k-1} \gamma^i r^{(j)}_i + \gamma^k \hat V_{\gamma, t-1}(s^{(j)}_{k}) \right)
\end{equation}
The following theorem from \citet{kearns2000bias} provides an upper bound on the error in the value function estimates defined by 
$\Delta^{\hat{V}_\gamma}_t \deq  \max_s \{|\hat{V}_{\gamma, t}(s) - V_\gamma(s)| \}$.
\begin{theorem} \citep{kearns2000bias} \label{prop: bound TD(lambda)}
for any $0 < \delta < 1$, let $\epsilon = \sqrt{\frac{2 \log(2k/\delta)}{n}}$. with probability $1-\delta$, 
\begin{equation}\label{eq: TD error bound}
    \Delta_{t}^{\hat{V}_{\gamma}} \leq \underbrace{\epsilon \left( \frac{1-\gamma^{k}}{1-\gamma} \right)}_{\text{variance term}} + \underbrace{\gamma^{k} \Delta_{t-1}^{\hat{V}_{\gamma}}}_{\text{bias term}},
\end{equation}
(The proof is provided in the Supplemental).
\end{theorem}
The first term $\epsilon ( \frac{1-\gamma^{k}}{1-\gamma})$, in the bound in Eq.~\ref{eq: TD error bound}, is a variance term arising from sampling transitions. In particular, $\epsilon$ bounds the deviation of the empirical average of rewards from the true expected reward. The second term is a bias term due to bootstrapping off of the current value estimate.

Similarly, we consider a phased version of multi-step TD($\Delta$). For each phase $t$, we update each $W$ as follows:
\begin{align}\label{eq: phased TD(Delta)}
    &\hat W_{z, t}(s) = \frac{1}{n} \sum_{j=1}^n \bigg ( \sum_{i=1}^{k-1} ( \gamma_z^{i} - \gamma_{z-1}^{i}) r^{(j)}_{i} + \nonumber \\  &( \gamma_z^{k_z} - \gamma_{z-1}^{k_z}) V_{\gamma_{z-1}}(s^{(j)}_{t+k})   + \gamma_z^{k_z} \hat W_z (s^{(j)}_{t+k}) \bigg ).
\end{align}
We now establish an upper bound on the error of phased TD($\Delta$) defined as the sum of error incurred by each W components $\sum_{z=0}^Z \Delta^{z}_t$, where $\Delta^{z}_t = \max_s \{|\hat{W}_z(s) - W_z(s)| \}$
\begin{theorem}\label{prop: bound TD(Delta)}
Assume that $\gamma_0 \leq \gamma_1 \leq \ldots \gamma_Z = \gamma$ and $k_0 \leq k_1 \ldots \leq k_Z = k$, for any $0 < \delta < 1$, let $\epsilon = \sqrt{\frac{2 \log(2k/\delta)}{n}}$, with probability $1-\delta$,
\begin{align}
\sum_{z=0}^Z  \Delta^{z}_t 
    & \leq \epsilon \frac{1-\gamma^{k}}{1-\gamma} + \underbrace{\epsilon \sum_{z=0}^{Z-1} \frac{\gamma_z^{k_{z+1}}-\gamma_z^{k_z}}{1-\gamma_z}}_{ \text{variance reduction}} \label{eq: bound TD(Delta)} \\
    & 
    + \underbrace{\sum_{z=0}^{Z-1} (\gamma_z^{k_z} - \gamma_{z}^{k_{z+1}} ) \sum_{u=0}^{z} \Delta^{u}_{t-1}}_{ \text{bias introduction}} + \gamma^{k}\sum_{z=0}^{Z} \Delta^{z}_{t-1} \nonumber
\end{align}
(The proof is provided in the Supplemental).
\end{theorem}
Comparing the bound for phased TD($\lambda$) in Theorem~\ref{prop: bound TD(lambda)} with the one for phased TD($\Delta$) in Theorem~\ref{prop: bound TD(Delta)}, we see that the latter allows for a variance reduction equal to $\epsilon \sum_{z=0}^{Z-1} \frac{\gamma_z^{k_{z+1}}-\gamma_z^{k_z}}{1-\gamma_z} \leq 0$ but it suffers from a potential bias introduction equal to $\sum_{z=0}^{Z-1} (\gamma_z^{k_z} - \gamma_{z}^{k_{z+1}} ) \sum_{u=0}^{z} \Delta^{u}_{t-1} \geq 0$. This is due to the compounding bias from all shorter-horizon estimates. 
We note that in the case that $k_z$ are all equal we obtain the same upper bound for both algorithms. 
It is a well known and often used result that the expected discounted return over $T$ steps is close to the infinite-horizon discounted expected return after $T \approx \frac{1}{1-\gamma}$ \citet{kearns2002near}. Thus, we can conveniently reduce $k_z$ for any $\gamma_z$ such that $k_z \approx \frac{1}{1-\gamma_z}$ so that we follow this rule. Thus, if we have $T$ samples, we can have an excellent bias-variance compromise on all time-scales $<<T$ by choosing $k_z=\frac{1}{(1-\gamma_z)} $, so that $\gamma_z^{k_z}$ is bound by a constant (since $\gamma_z^{\frac{1}{1-\gamma_z}} \leq \frac{1}{e}$) for all $z$.
This provides intuitive ways to set both $\gamma_z$ and $k_z$ values (as well as all other parameters) without necessarily searching. We can double the effective horizon at each increasing $W_z$ (to keep a logarithmic number of value functions with respect to the horizon) and similarly adjust all other parameters for estimation.

\section{Experiments}
\label{sec:experiments}
All hyperparameter settings, extended details, and the reproducibility checklist for machine learning research~\cite{repro_checklist} can be found in the Supplemental\footnote{Link to Code: \href{https://github.com/facebookresearch/td-delta}{ github.com/facebookresearch/td-delta}}.

\subsection{Tabular}

\begin{figure}[!htbp]
    \centering
    \includegraphics[width=.235\textwidth]{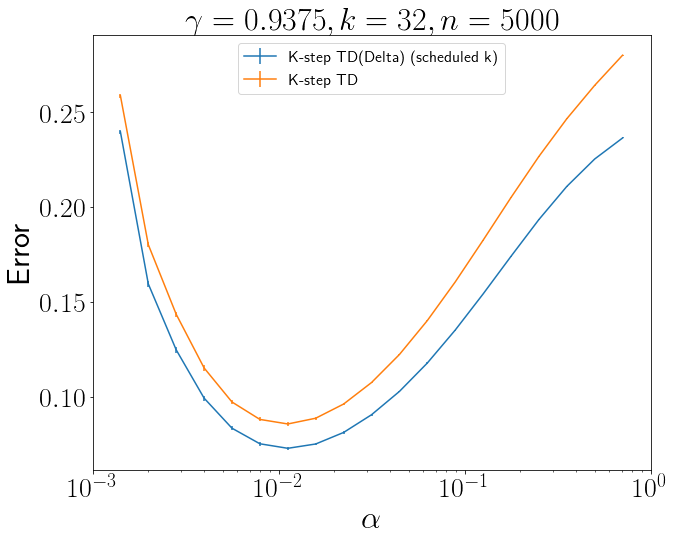}
    \includegraphics[width=.235\textwidth]{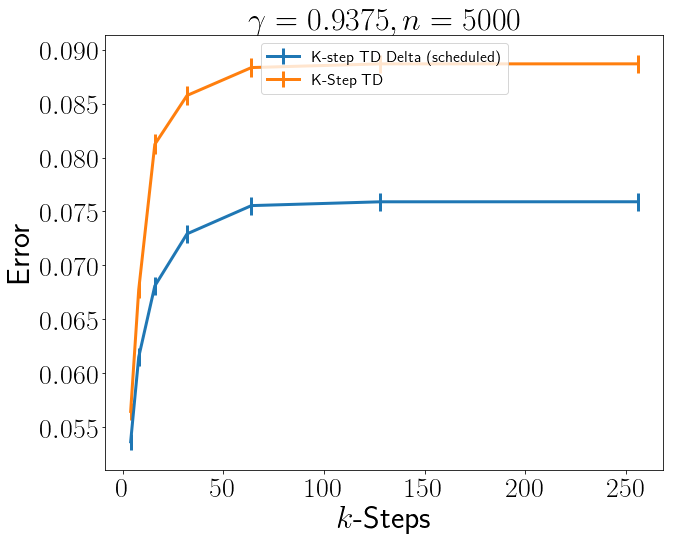}
    \caption{(Left) $\gamma_Z = .9375$, 250 random seeds on the 5-state ring MDP. Error denotes the absolute error against the true discounted value function (pre-computed ahead of time using Value Iteration) averaged across the entire learning trajectory (5000 timesteps). Error bars denote standard error across random seeds. (Right) The average absolute error for the optimal learning rate at each $k$-step return up to the effective planning horizon of $\gamma_Z$.}
    \label{fig:example}
\end{figure}
\begin{figure*}[!htbp]
    \centering
    \includegraphics[width=.49\textwidth]{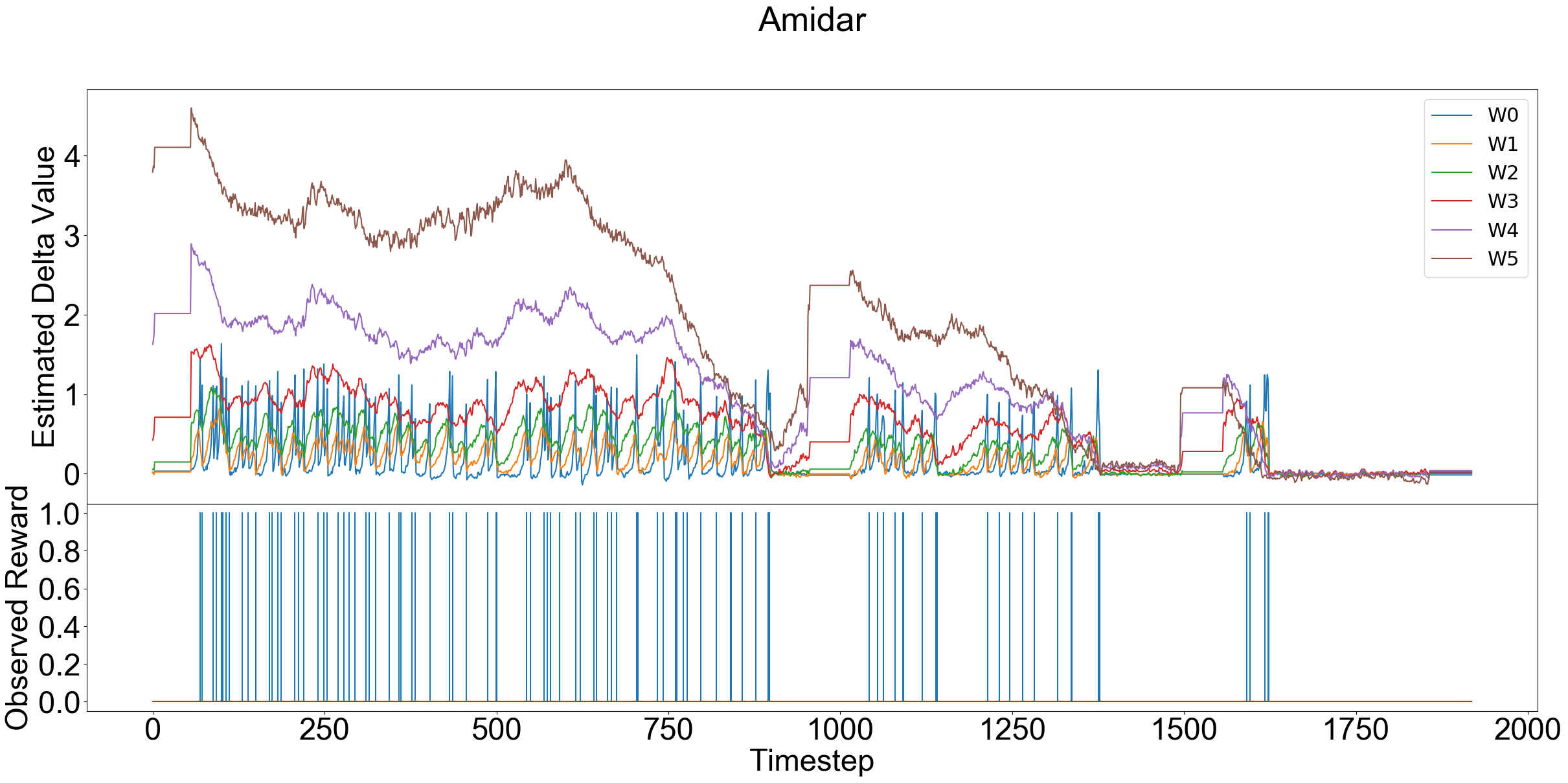}
     \includegraphics[width=.49\textwidth]{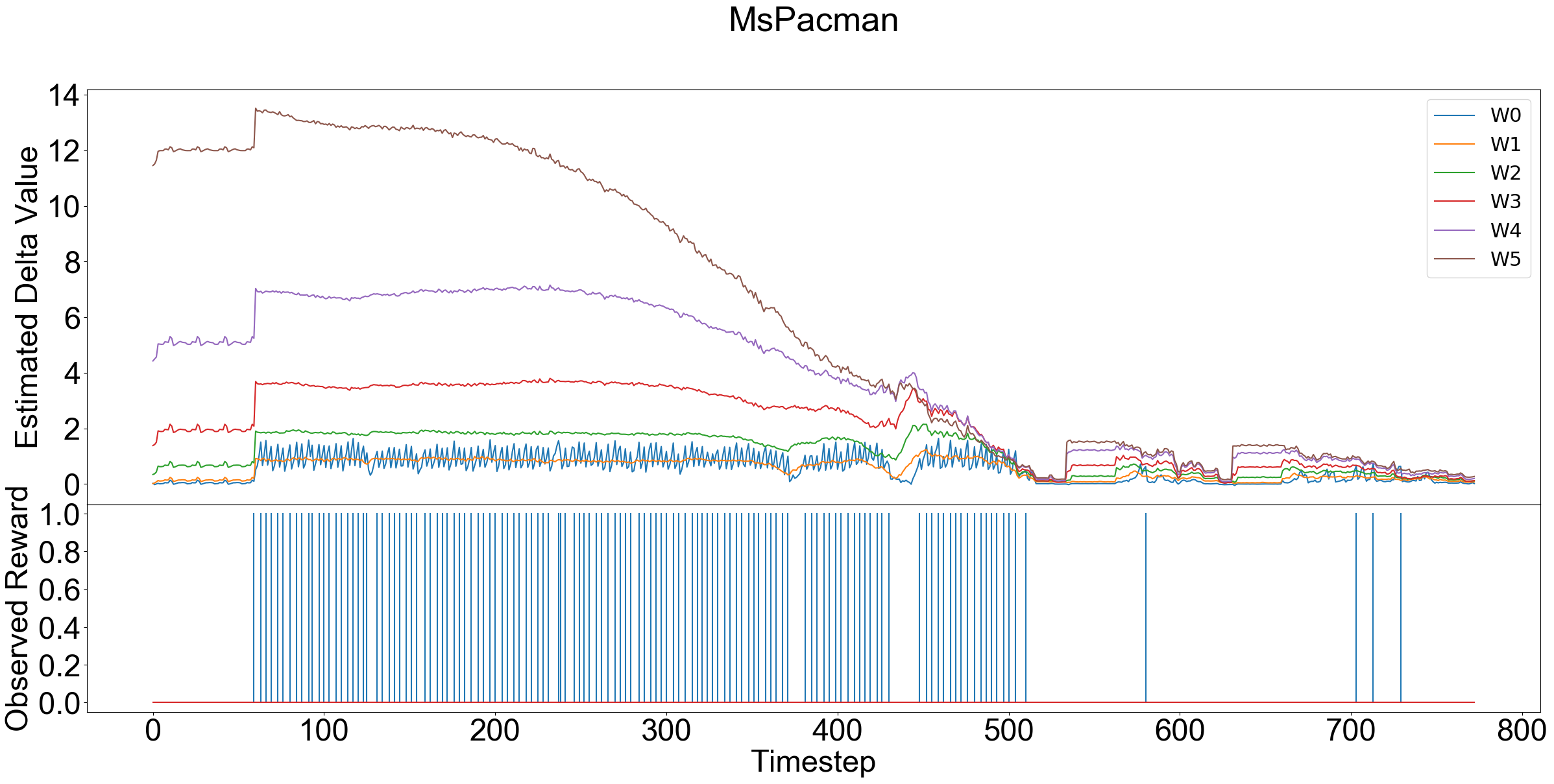}
    \caption{The $W_z$ estimators versus the reward over a single episode in two games - the drops in value align with a lost life. This is done on a single rollout trajectory of the trained PPO-TD($\hat{\lambda}, \Delta$) agent using random seed $1153780$.}
    \label{fig:trajectory}
\end{figure*}
We use the same 5-state ring MDP as in \citet{kearns2000bias} -- a diagram of which is available in the Supplemental for clarity -- to demonstrate performance gains under decreasing $k$-step regimes as described in Section~\ref{sec:proofs}. For all experiments we provide a variable number of gammas starting with $0$ and increasing according to $\gamma_{z+1} = \frac{\gamma_z + 1}{2}$ until the maximum desired $\gamma_Z$ is reached. Similarly, $k_z := \frac{1}{1-\gamma_z}, \forall z$ as described earlier. The baseline is a single estimator with $\gamma=\gamma_Z, k=k_Z$. We run a grid of various $\gamma_Z$ and $k_Z$ values and use standard TD-style updates~\cite{sutton1988learning} for our experiments.

We compare against the true error which can be calculated ahead of time using value iteration (VI)~\cite{bellman1957markovian}.
In the case where we do not tailor $k$ (all $k_z$ are equal), as predicted by the theory in Section~\ref{sec:proofs}, the performance is exactly equal to the single estimator case. We compute the average error from the VI pre-computed optimal value function across the entire training trajectory and plot a sample of these results in Figure~\ref{fig:example}. We supply all results in the supplemental across a set of 7 different $\gamma$ values corresponding to effective planning horizons of ($4, 8, 16, 32, 64, 125, 250$).  We note that performance gains tend to increase with larger $\gamma$ and $k$ values as discussed further in the supplemental. However, consistent with the theory, in all cases we still perform about equal to (statistically) or significantly better than the single estimator setting.

\subsection{Dense reward Atari}
\label{sec:deepexps}

We further demonstrate performance gains in Atari using the PPO-based version of TD($\Delta$). We directly update PPO with TD($\lambda, \Delta$), using the code of \citet{pytorchrl}. We compare against the standard PPO baseline with hyperparameters as found in \cite{schulman2017proximal, pytorchrl}. Our architecture differs slightly from the PPO baseline as the value function now outputs $Z+1$ outputs ($1$ for each $W$). For complete fairness, we also add another neural network architecture which replicates the parameters of TD($\Delta$). That is, we use a neural network value function that outputs $Z+1$ values which are summed together before computing the value loss (we call this PPO+). 
We run two versions of TD($\Delta$). The first version, as described in Section~\ref{sec:GAE}, uses a similar set of $\gamma_z$ sequence as in the ring MDP experiments (starting at $\gamma_Z = 0.99$ and halving the horizon) where $\lambda_z$ is set for each lower $\gamma_z$ such that $\gamma_z \lambda_z = \gamma_Z \lambda_Z$ as per Theorem~\ref{prop: equivalence linear}. However, we note that due to the use of adaptive optimizers, performance may improve as parameters are honed for each delta estimator. Just as in the tabular setting where $k_z$ can be reduced for lower delta estimators, in this setting as well, parity with the baseline model is not necessary and $\lambda$ can effectively be reduced. To this end, we introduce a second version of our method, labelled PPO-TD($\hat\lambda, \Delta$), where we limit $\lambda_z \le 1$. 

We run experiments on the $9$ games defined in \cite{Bellemare2016unifying} as `Hard' with dense rewards. We chose `Hard' games as these games are most likely to need algorithmic improvements to solve. We chose dense reward tasks since we do not tackle the problem of exploration here (needed for tackling sparse reward settings), but rather modeling of complex value functions which dense reward settings are likely to benefit from.
As seen in Table~\ref{tab:asymptotic} (with average return across training and on hold-out no-op starts in the Supplemental), PPO-TD($\lambda, \Delta$) performs (statistically) significantly better in a certain class of games roughly related to the frequency of non-zero rewards (the density). In both versions of TD($\Delta$), the algorithms perform worse asymptotically than the baselines  in two games, Zaxxon and Wizard of Wor, which belong to a class of games with lower density. 
Though PPO-TD($\hat \lambda, \Delta$) performs somewhat better in both cases, as we will see in Section~\ref{sec:tuning}, it is still possible to improve performance further in these games by tuning the number and scale of $\gamma_Z$ factors. 

One may wonder why performance improves in increasingly dense reward settings. 
There is a basic intuition that TD($\Delta$) would allow for quick learning of short-term phenomena, followed by slower learning of long-term dependencies. Such a decomposition is reflected in a rolled out trajectory using the learned policy in Figure~\ref{fig:trajectory}. There, the long-term $W_Z$ value declines early according to a consistent gradient towards a lost life in the game, while short-term phenomena continue to be captured in the smaller components like $W_0$.
\begin{table*}[!htbp]
    \centering
    \resizebox{\textwidth}{!}{    \begin{tabular}{|c|c|c|c|c|c|c|c|c|c|}
    \hline
        Algorithm & Zaxxon & WizardOfWor & \textbf{Qbert} & \textbf{MsPacman} & \textbf{Hero} & \textbf{Frostbite} & \textbf{BankHeist} & \textbf{Amidar} & \textbf{Alien} \\
         \hline
          PPO-TD$\left (\lambda, \Delta \right)$ & 396 $\pm$ 210 &  2118 $\pm$ 138 &13428 $\pm$ 333 $\dagger$ &2273 $\pm$ 67 $\dagger$&29074 $\pm$ 512 $\dagger$&292 $\pm$ 7&1183 $\pm$ 13& 731 $\pm$ 30 $\dagger$&1606 $\pm$ 112$^*$ \\
          \hline
          PPO-TD$\left (\hat \lambda, \Delta \right)$  & 3291 $\pm$ 812&2440 $\pm$ 89&13092 $\pm$ 430 $\dagger$&2241 $\pm$ 78 $\dagger$&29014 $\pm$ 764 $\dagger$&304 $\pm$ 21&1166 $\pm$ 5 &672 $\pm$ 45&1663 $\pm$ 113$^*$\\
          \hline
          PPO+ & 7006 $\pm$ 211 $\dagger$ &2870 $\pm$ 218 $\dagger$&10594 $\pm$ 335&1876 $\pm$ 89&23511 $\pm$ 843&299 $\pm$ 2&1199 $\pm$ 5&611 $\pm$ 34& 1374 $\pm$ 85\\
          \hline
          PPO & 7366 $\pm$ 223 $\dagger$& 3408 $\pm$ 193  $\dagger$ &11735 $\pm$ 387&1888 $\pm$ 111&21038 $\pm$ 972&294 $\pm$ 5&1190 $\pm$ 3&575 $\pm$ 54& 1315 $\pm$ 70\\
          \hline
          Reward Density &1.15&1.07&12.26&13.27&13.46&5.04&6.3&4.63&11.33\\
          \hline
    \end{tabular}}
    \caption{Asymptotic Atari performance (across last 100 episodes) with the mean across 10 seeds and the standard error. $\dagger$ denotes significantly better results over our algorithm in the case of baselines or over the best baseline in the case of our algorithm using Welch's t-test with a significance level of $.05$ and bootstrap confidence intervals~\cite{colas2018many,henderson2018deep}. $^*$ indicates significant using bootstrap CI, but not t-test. Bold algorithms are where we perform as well as or significantly better than the baselines. Reward Density is frequency of rewards per $100$ time-steps averaged over $10k$ timesteps under learned policy using baseline (PPO). Notice how the task `Zaxxon' has a much lower frequency than the largest frequency task (Hero).
    More information in Supplemental.
        }
    \label{tab:asymptotic}
\end{table*}
\begin{figure}[!htbp]
    \centering
    \includegraphics[width=.23\textwidth]{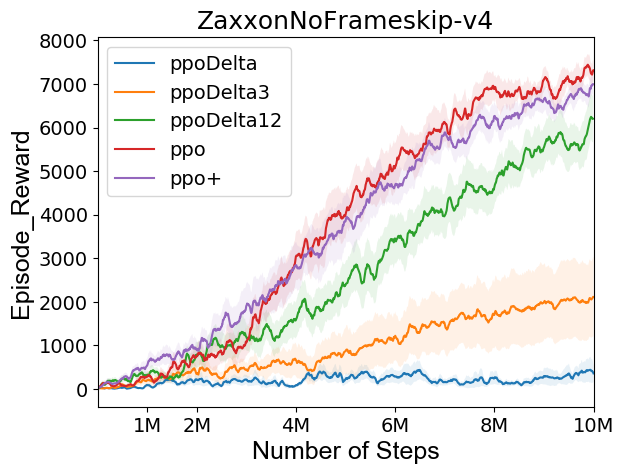}
    \includegraphics[width=.23\textwidth]{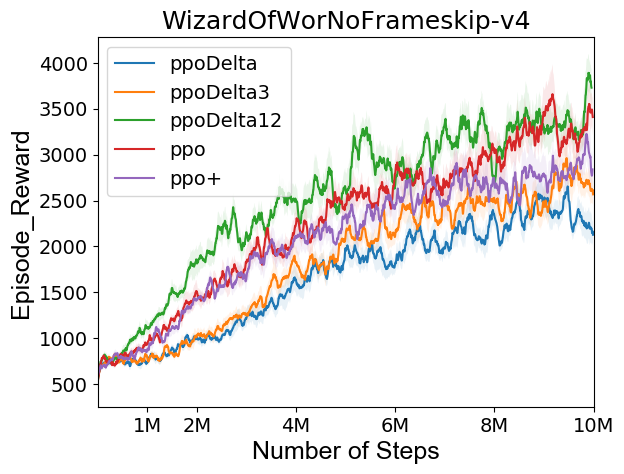}
    \caption{Performance of TD($\Delta$) variations vs. the baselines on Zaxxon and WizardOfWor. ppo+ refers to ppo with an augmented architecture. ppoDelta refers to setting $\gamma_z\lambda_z=\gamma\lambda$ $\forall z$. ppoDelta3 and ppoDelta12 only use two value functions with horizons $(3,100)$ and $(12,100)$ respectively.
    Shaded region is standard error across $10$ random seeds. }
    \label{fig:zaxxon}
    \end{figure}

\subsection{Tuning and Ablation}
\label{sec:tuning}
In the previous section we demonstrated how using a fixed set of $\gamma, \lambda$ tailored to an intuitive set of progressively large horizons, we could yield performance gains in a number of environments over the single estimator case. However, a performance drop was seen in the case of Zaxxon and WizardOfWor. Due to our bias-variance trade-off in bootstrapping from smaller delta estimators, a curriculum based on smaller horizons may effectively slow learning in some cases. However, the benefit of separating value functions in a flexible way, as we propose here, is that they can be tuned. In Figure~\ref{fig:zaxxon} (with full results in the Supplemental), we show how different $\gamma$ values can be used to improve asymptotic performance to match the baseline. 
By increasing the lowest effective horizon ($\gamma_0$) of $W_0$, we bias the algorithm less toward myopic settings and increase the rate of learning comparable to the baselines.
Further tuning of the number of components and their parameters ($\gamma_z, \lambda_z$, learning rate, etc.) may further improve performance.

\section{Discussion}
\label{sec:conclusion}
In this work we explore temporal decomposition of the value function. More concretely, we proposed a novel way for decomposing value estimators via a Bellman update based on the difference between two value estimators with different discount factors. This has convenient theoretical and practical properties which help improve performance in certain settings. 
These properties have additional benefits: they allow for a natural way to distribute and parallelize training, easy inspection of performance at different discount factors, and the possibility of lifelong learning by adding or removing components. 
Moreover, we have also highlighted the limitations of this method (introduced bias toward myopic returns) when using the simple parameter settings we propose. However, these limitations can be overcome with the additional ability to tune parameters at different time-scales.   
We briefly discuss the added benefits of TD($\Delta$) below.

\textbf{Scalability:}
While we have not pursued it experimentally here, another benefit of separating value functions in this way is that 
this reflects a natural way of distributing updates across systems for large scale problems. In fact, prior work has sought different ways to scale RL algorithms through partitioning methods (though typically through other means like dividing the state space)~\cite{wingate2004solving,wingate2004p3vi}. Our work provides another such method for scaling RL systems in a different way. A TD($\Delta$) update can be spread across many machines, such that each $W_z$ is updated separately (as long as weights are synced across machines after a parallel update).

\textbf{Additional tuning ability:} Many of the performance improvements seen here come not necessarily from the decomposition method itself, but from the ability to set certain parameters differently for each component. 
The fine-grained nature of the decomposition of the value function allows for further improvement by tuning the number of delta estimators and the $\gamma_z$ values which correlate with them. In the future, a meta-gradient method as \citet{xu2018meta} proposed could be used to automatically scale delta estimators to time-scales which require more computational complexity.
However, the default method for tailoring $\gamma_z$ and $k_z$ and $\lambda_z$ values as described above (doubling effective horizons until the maximum horizon is reached), still yields improvements in most games tested here, without additional tuning.

\textbf{Interpreting performance at different time-scales:} 
As we mention in Section~\ref{sec:related}, another benefit of TD($\Delta$) is the ability to examine the value function at different s after a single pass of learning. That is, we can compose value functions from $\gamma_0, ..., \gamma_Z$ and understand the differences between different time-scales. This has implications for real-world uses with similar motivations as \citet{sherstan2018generalizing} describe. Take for example an MDP where the bulk of rewards are in some central region, requiring following a policy $\pi$ for some number of timesteps before reaching the dense reward region. By examining each $W_z$ component as we do in Figure~\ref{fig:trajectory}, a practitioner could understand how far into a trajectory $\pi$ must be followed before the dense reward region is reached. This adds some layer of interpretability to the value function which is missing in the single estimator case. Similarly, this may have the benefit in determining an optimal stopping point for the policy. In production systems where there is a cost to running a policy (time, money, or energy resources), yet the policy can be run indefinitely, a practitioner may use $W_z$ components to determine if the discounted return at a larger horizon is worth the cost.

\textbf{TD($\Delta$) as an (almost) anytime algorithm:} Throughout this work, we emphasize this algorithm as a complement to selection of a final $\gamma_Z$. The longest horizon discount factor can be chosen according to other methods (hyperparameter optimization or meta-gradient methods). However, an added benefit of our method not explored in this work is its functionality as an almost anytime algorithm.
While longer time horizons will take longer to converge, at any point in time the sum of all horizons which have converged are a suitable approximation for the value function at that intermediary point.
Therefore, with enough resources, TD($\Delta$) could potentially  at anytime add one further time-scale $Z\leftarrow Z+1$ (initialized to $W_{Z+1}=0$ which preserves the current $V$ estimate).
This has implications for methods which already extend discount factors through a curriculum~\cite{OpenAI_dota}.

\textbf{Other extensions:}
Our method should also extend easily to any TD-like methods such as Sarsa($\lambda$) and Q-learning with few adjustments. We leave this to future work.

\textbf{Conclusion: }
We believe that TD($\Delta$) is a important drop-in addition to any TD-based training methods that can be applied to a number of existing model-free RL algorithms. 
We especially highlight the value of this method for performance tuning. We show that a simple sequence of $\gamma_z$ values based on doubling horizon values can yield performance gains especially in dense settings, but this performance can be enhanced further with tuning. As the complexity of modeling and training long-horizon problems increases, TD($\Delta$) may be another tool for scaling and honing production systems for optimal performance.

\newpage

\section*{Acknowledgements}
We thank Alexandre Pich\'e, Vincent Fran\c{c}ois-Lavet, and Harsh Satija for many helpful discussions about the work.

\bibliography{example_paper}
\bibliographystyle{icml2019}

\appendix

\onecolumn

\section{Reproducibility Checklist}

We follow the reproducibility checklist~\cite{repro_checklist} and point to relevant sections explaining them here.

For all algorithms presented, check if you include:
\begin{itemize}
    \item \textbf{A clear description of the algorithm.} See Algorithm 1 in the main paper.
    \item \textbf{An analysis of the complexity (time, space, sample size) of the algorithm.} See Proofs section in Supplemental Material for bias-variance trade-offs and some rudimentary complexity analysis. Experimentally, we demonstrate similarity or improvements on sample complexity as discussed in main paper. In terms of computation time (for Deep-RL experiments) the newly proposed algorithm is almost identical to the baseline due to its parallelizable nature.
    \item \textbf{A link to a downloadable source code, including all dependencies.} See experimental section in Appendix and main paper, the code is linked in the experimental section also.
    \end{itemize}

For any theoretical claim, check if you include:
\begin{itemize}
    \item \textbf{A statement of the result.} See analysis section in the main paper and the supplemental Proof section.
    \item \textbf{A clear explanation of any assumptions.} See supplemental Proof section.
    \item \textbf{A complete proof of the claim.} See supplemental Proof section.
\end{itemize}

For all figures and tables that present empirical results, check if you include:
\begin{itemize}
    \item \textbf{A complete description of the data collection process, including sample size.} We use standard benchmarks provided in OpenAI Gym~\cite{gym,Bellemare2016unifying}.
    \item \textbf{A link to downloadable version of the dataset or simulation environment.} See: \href{https://github.com/openai/gym}{github.com/openai/gym} for OpenAI Gym benchmarks and for Ring MDP see included code in experimental section.
    \item \textbf{An explanation of how samples were allocated for training / validation / testing.} We do not use a split because we are examining the optimization performance. Atari environments we use are deterministic, so we run 10 random seeds where the randomness stems from the initialization of the neural networks and policy sampling. As described in \citet{henderson2018deep,colas2018many} we perform statistical analysis on this seed distribution to determine the range of performance expected of an algorithm for a deterministic environment as compared to the single estimator case. We also run on a hold out set of random starts (we use 1-30 no-ops at the start of training as in \citet{mnih2013playing} and show those results in the Supplemental Material).
    \item \textbf{An explanation of any data that were excluded.} We exclude the other Atari games because of time constraints and because we hypothesize that our method will help in dense and complex games as defined in \citet{Bellemare2016unifying}.
    \item \textbf{The range of hyper-parameters considered, method to select the best hyper-parameter configuration, and specification of all hyper-parameters used to generate results.} For parity with the baseline, we use similar hyperparameters in the deep learning case as recommended in \citet{schulman2017proximal,pytorchrl} (described in the experimental section of the Supplemental). In the tabular case we run a grid. For choosing $\gamma_z$ values for our own method we use a schedule as described in the main paper for simplicity and demonstrate how performance can be improved by tuning these values.
    \item \textbf{The exact number of evaluation runs.} 10 seeds for Atari experiments, 3000 episodes per random seed for 200 random seeds for tabular.
    \item \textbf{A description of how experiments were run.} See Experiments section in Supplemental.
    \item \textbf{A clear definition of the specific measure or statistics used to report results.} Undiscounted return for last 100 episodes in asymptotic results and across all training episodes for average results. Welch's t-test used for significance testing using script from \citet{colas2018many} across random seeds.
    \item \textbf{Clearly defined error bars.} Standard error used in all cases.
    \item \textbf{A description of results with central tendency (e.g. mean) and variation (e.g. stddev).} We use standard error, but results are seen in main paper.
    \item \textbf{A description of the computing infrastructure used.} We distribute all runs across 1 CPUs and 1 GPU per run for Atari experiments. GPU used: GP100. Both the baseline and our methods take approximately 8 hours to run.
\end{itemize}

\section{Proofs}

\subsection{Equivalence to TD($\lambda$) with linear function approximation: theorem \ref{prop: equivalence linear}}

The proof is by induction. We first need to show that the base case holds - at $t=0$. This is trivially true given the assumption on initialization - we note this is trivial to do with zero-initialization. 

For a given times-step $t$, we assume that the statement holds i.e
$\sum_{z=0}^Z \theta^{z}_{t} = \theta^{\gamma}_t$
and let's show that it holds at next time-step $t+1$.

\begin{align}
    \sum_{z=0}^Z \theta^{z}_{t+1} &= \sum_{z=0}^Z \left( \theta^{z}_{t} + \alpha_z \left( G^{z, \lambda_z}_t - \hat W_z(s_t)\right) \phi(s_t)\right) \\
    & = \theta^{\gamma}_t + \sum_{z=0}^Z \alpha_z \left( \sum_{k=t}^\infty (\lambda_z \gamma_z)^{k-t} \delta_k^{z} \right) \phi(s_t) \quad \text{by induction assumption} \\
    & = \theta^{\gamma}_t + \alpha \sum_{k=t}^\infty (\lambda \gamma)^{k-t} \underbrace{\left( \sum_{z=0}^Z \delta_k^{z} \right) }_{(\star)}\phi(s_t) \quad \text{thanks to }\alpha_z = \alpha, \lambda_z \gamma_z = \lambda \gamma,  \forall z
\end{align}
To prove that $\sum_{z=0}^Z \theta^{z}_{t+1} = \theta^{\gamma}_{t+1}$, we need to prove that term $(\star) = \sum_{z=0}^Z \delta_k^{z}$ is equal to the standard TD error $\delta^{\gamma}_k$.

\begin{align}
    \sum_{z=0}^Z \delta^{z}_{k} & = r_k + \gamma_0 \hat V_{\gamma_0}(s_{k+1}) - \hat V_{\gamma_0}(s_k) + \sum_{z=1}^Z \left( (\gamma_{z} - \gamma_{z-1}) \sum_{u=0}^{z-1} \langle \theta_{t}^{u}, \phi(s_{k+1}) \rangle + \gamma_z \langle \theta_t^{z},  \phi(s_{k+1}) \rangle - \langle \theta_t^{z}, \phi(s_k) \rangle \right) \nonumber \\
    & = r_k + \gamma_0 \hat V_{\gamma_0}(s_{k+1})
    + \langle \sum_{z=1}^Z \left( \gamma_z \sum_{u=0}^{z} \theta_{t}^{u} - \gamma_{z-1} \sum_{u=0}^{z-1} \theta_{t}^{u} \right), \phi(s_{k+1}) \rangle - \langle \sum_{z=0}^Z\theta_t^{z}, \phi(s_t) \rangle  \nonumber \\
    & = r_k + \gamma_0 \hat V_{\gamma_0}(s_{k+1}) + 
    \gamma_Z \langle \sum_{z=0}^{Z} \theta_{t}^{z}, \phi(s_{k+1}) \rangle - \gamma_0  \langle \theta_t^{0}, \phi(s_{t+1}) \rangle - V_{\gamma}(s_k) \nonumber \\
    & = r_k + \gamma_0 \hat V_{\gamma_0}(s_{k+1}) - \gamma \hat V_{\gamma}(s_{k+1}) - \gamma_0 \hat V_{\gamma_0}(s_{k+1}) - \hat V_{\gamma}(s_k) \nonumber \\
    & = r_k + \gamma \hat V_{\gamma}(s_{t+1}) - \hat V_{\gamma}(s_k) = \delta^{\gamma}_k
\end{align}

\subsection{Proof that $\lambda$ can be $>1$ as long as $\lambda < \frac{1 + \gamma}{2\gamma} $: theorem \ref{prop: contraction}}
The Bellman operator is defined as follows:
\begin{equation}
    T = r + \gamma P  
\end{equation}
where $r$ and $P$ are respectively the expected reward function and the transition probability operator induced by the policy $\pi$.
The TD($\lambda$) operator is defined as a geometric weighted sum of $T$, as follows:
\begin{equation}
    T_{\lambda} = (1-\lambda) \sum_{k=0}^\infty \lambda^k (T)^{k+1}
\label{eq: lambda-sum}
\end{equation}
One could conclude that we need $\lambda \in [0,1]$ so that the above sum is finite. An equivalent definition of $T_\lambda$ is as follows: for all function $W$ 
\begin{equation}
        T_\lambda W=  W + \left( I - \lambda \gamma P \right) ^{-1} \left ( T W - W \right).
\label{eq: lambda-inverse}
\end{equation}
The latter formula is well defined if $0 \leq \lambda \gamma < 1$ to guarantee that the spectral norm of the operator $\lambda \gamma P$ is smaller that one and hence $I - \lambda \gamma P$ is invertible. However, by considering values of $\lambda$ greater than one, we loose the equivalence between equations \ref{eq: lambda-sum} and \ref{eq: lambda-inverse}. This is not an issue since in practice we use TD error for training which correspond in expectation to the definition of $T_{\lambda}$ given in \ref{eq: lambda-inverse}.

Now, let's study the contraction property of the operator $T_{\lambda}$. First, \ref{eq: lambda-inverse} can be rewritten as : 
\begin{equation}
    T_{\lambda} = \left( I - \lambda \gamma P \right) ^{-1} \left ( T W -\lambda \gamma P W\right).
\end{equation}
For any functions $W_1$ and $W_2$, we have:
\begin{align}
       T_\lambda W_1 - T_\lambda W_2 &= \left( I - \lambda \gamma P \right) ^{-1} \left(T W_1 - T W_2 - \lambda \gamma P \left(W_1 - W_2 \right) \right) \nonumber \\
       &= \left( I - \lambda \gamma P \right)^{-1} \left (\gamma P \left (W_1 - W_2 \right )  - \lambda \gamma P \left(W_1 - W_2 \right) \right) \nonumber \\
       &= \left( I - \lambda \gamma P \right)^{-1} \left ( \gamma \left ( 1 - \lambda \right ) P  \left( W_1 - W_2 \right) \right)
\end{align}

We obtain then: 
\begin{equation}
\lVert T_\lambda W_1 - T_\lambda W_2 \rVert \leq \frac{\gamma \lvert 1- \lambda \rvert}{1-\lambda \gamma}\lVert W_1 - W_2 \rVert 
\end{equation}
We know that $0 \leq \lambda \leq 1$ is a contraction, for $\lambda > 1$ we need:
\begin{align}
\frac{\gamma(\lambda -1)}{1-\lambda \gamma} &< 1 \Rightarrow
\gamma\lambda -\gamma < 1-\lambda \gamma  \nonumber  \\
\Rightarrow 2 \gamma \lambda &< 1 + \gamma 
\Rightarrow
\lambda < \frac{1 + \gamma}{2\gamma} 
\end{align}

Therefore, $T_\lambda$ is a contraction if $0 \leq \lambda < \frac{1+\gamma}{2\gamma}$. This directly implies that for $
\gamma<1$, $\gamma \lambda < 1$.

\subsection{Expanded Bias-variance comparison proof: theorem \ref{prop: bound TD(lambda)}}

Hoeffding inequality guarantees for a variable that is bounded between $[-1,+1]$, 
that 
\begin{equation}
    P(\left|\frac{1}{n}\sum_{j=1}^n r^{(j)}_i - \mathbb{E}[r_i] \right| \geq \epsilon) \leq 2 \exp \left(\frac{-2 n^2 \epsilon^2}{n 2^2}\right) 
\end{equation}
If we assume $n$ and the probability of exceeding an $\epsilon$ value is fixed to be no more than $\delta$, we can solve for the resulting value of $\epsilon$:
\begin{equation}
2 \exp(\frac{- n^2 \epsilon^2}{2 n}) = \delta \Longrightarrow 
\frac{-n \epsilon^2}{2} = \log(\delta/2)  \Longrightarrow 
\epsilon = \sqrt{ \frac{2 \log {(2/ \delta )}}{n}}
\end{equation}
So by Hoeffding, if we have $n$ samples, with probability at least $1-\delta$, 
\begin{equation}
\left| \frac{1}{n} \sum_j r^{(j)}_{i} - \mathbb{E}[r_i] \right| \leq \epsilon = \sqrt{ \frac{2 \log {(2/\delta )}}{n}}. 
\end{equation}

Since we want each of the $k$ $\E[r_i]$ reward terms to all be estimated up to $\epsilon$ accuracy with high probability, we can use a union bound to ensure that the probability that we fail to estimate any of these $k$ expected reward terms is $\delta$ by requiring that the probability we fail to estimate any of them is at most $\delta/k$. Substituting this into the above equation for $\epsilon$, we obtain that with probability at least $1-\delta$, each of the  $\E[r_i]$ terms are estimated to within $\epsilon = \sqrt{ \frac{2 \log {\frac{2k}{\delta}}}{n}}$ accuracy. 

This is a slightly different expression than \citet{kearns2000bias} obtain in terms of constants, it is unclear which inequality was used to come to their method as their proof was omitted.

We can now assume that all $E[r_i]$ terms are estimated to at least $\epsilon$ accuracy. Substituting this back into
the definition of the k-step TD update we get 

\begin{align}
\hat{V}_{t+1}(s) - V(s) &= \frac{1}{n} \sum_{j=1}^n \left( r^{(j)}_0 + \gamma r^{(j)}_1 + \ldots \gamma^{k-1}r^{(j)}_{k-1} + \gamma^k V_t (s^{(j)}_k) \right) - V(s) \nonumber \\
&= \sum_{i=0}^{k-1} \gamma^i \left( \frac{1}{n} \sum_{i=1}^n r^{(j)}_i - \E[r_i]\right) + \gamma^k \left( \frac{1}{n} \sum_{i=1}^n  V_t(s_k^i) - \E [V(s_k)] \right)
\end{align}
where in the second line we re-expressed the value in terms of a sum of $k$ rewards. We now upper bounded the difference from $E[r_i]$ by $\epsilon$ to get
\begin{align}
\hat{V}_{t+1}(s) - V(s) 
&\leq  \sum_{l=0}^{k-1} \gamma^l  \epsilon +  \gamma^k \left( \frac{1}{n} \sum_{i=1}^n  V_t(s_k^i) - \E [V(s_k)] \right) \nonumber \\
&\leq  \epsilon \frac{(1-\gamma^k)}{1-\gamma} + \gamma^k \left( \frac{1}{n} \sum_{i=1}^n  V_t(s_k^i) - \E [V(s_k)] \right)
\end{align}

and then the second term is bounded by $\Delta^{\gamma}_{t-1}$ by assumption. Hence, we obtain:
\begin{eqnarray}
\Delta^{\gamma}_{t} \leq \epsilon \frac{(1-\gamma^k)}{1-\gamma} + \gamma^k \Delta^{\gamma}_{t-1}
\end{eqnarray}

\subsection{Proving error bound for phased TD($\Delta$): theorem \ref{prop: bound TD(Delta)}}

Phased TD($\Delta$) update rules for $z \geq 1$:
\begin{dmath}\label{eq: phased TD(Delta)2}
    \hat W_{z, t}(s) = \frac{1}{n} \sum_{j=1}^n ( \sum_{i=1}^{k_z-1} ( \gamma_z^{i} - \gamma_{z-1}^{i}) r^{(j)}_{i} + ( \gamma_z^{k_z} - \gamma_{z-1}^{k_z}) \hat V_{\gamma_{z-1}}(s^{(j)}_{k}) + \gamma_z^{k_z} \hat W_z (s^{(j)}_{k}) )
\end{dmath}
We know that according to the multi-step update rule  \ref{eq:multistepupdate} for $z \geq 1$:
\begin{equation}
    W_z(s) = \mathbb{E} \left [\sum_{i=1}^{k_z-1} ( \gamma_z^{i} - \gamma_{z-1}^{i}) r_{i} + ( \gamma_z^{k_z} - \gamma_{z-1}^{k_z}) V_{\gamma_{z-1}}(s_{k}) + \gamma_z^{k_z} W_z (s_{k}) \right]
\end{equation}
Then, subtracting the two expressions gives for $z \geq 1$:
\begin{align}
    \hat W_{z, t}(s) - W_z(s) & = \sum_{i=1}^{k_z-1} ( \gamma_z^{i} - \gamma_{z-1}^{i})  \left( \frac{1}{n} \sum_{j=1}^n r^{(j)}_{i} - \E [r_i] \right) + (\gamma_z^{k_z} - \gamma_{z-1}^{k_z}) \left(\sum_{u=0}^{z-1} \hat W_u(s^{(j)}_k) - \E [W_z(s_k)] \right) \nonumber \\
    & \quad + \gamma_z^{k_z} \left(W_z(s^{(j)}_k) - \E [W_z(s_k)] \right)
\end{align}
Assume that $k_0 \leq k_1 \leq \ldots k_Z=k$, the W estimates share at most $k_Z=k$ reward terms $\frac{1}{n} \sum_{j=1}^n r^{(j)}_{i}$, Using Hoeffding inequality and union bound, we obtain that with probability $1-\delta$, each $k$ empirical average reward $\frac{1}{n} \sum_{j=1}^n r^{(j)}_{i}$ deviates from the true expected reward $\E[r_i]$ by at most $\epsilon = \sqrt{\frac{2 \log(2k/\delta)}{n}}$. Hence, with probability $1-\delta$, $\forall z \geq 1$, we have:
\begin{align}
    \Delta^{z}_t & \leq \epsilon  \sum_{i=1}^{k_z-1} ( \gamma_z^{i} - \gamma_{z-1}^{i}) + (\gamma_z^{k_z} - \gamma_{z-1}^{k_z}) \sum_{u=0}^{z-1} \Delta^{u}_{t-1} + \gamma_z^{k_z} \Delta^{z}_{t-1} \nonumber \\
    & = \epsilon \left( \frac{1- \gamma_z^{k_z}}{1-\gamma_z} - \frac{1- \gamma_{z-1}^{k_z}}{1-\gamma_{z-1}} \right) + (\gamma_z^{k_z} - \gamma_{z-1}^{k_z}) \sum_{u=0}^{z-1} \Delta^{u}_{t-1} + \gamma_z^{k_z} \Delta^{z}_{t-1}
\end{align}
and 
\begin{equation}
    \Delta^{0}_t \leq \epsilon \frac{1-\gamma_{0}^{k_0}}{1-\gamma_{0}} + \gamma_0^{k_0} \Delta^{0}_{t-1}
\end{equation}

Summing the two previous inequalities gives:
\begin{align}
    \sum_{z=0}^Z  \Delta^{z}_t & \leq \epsilon \frac{1-\gamma_{0}^{k_0}}{1-\gamma_{0}} + 
    \epsilon \sum_{z=1}^Z \left( \frac{1- \gamma_z^{k_z}}{1-\gamma_z} - \frac{1- \gamma_{z-1}^{k_z}}{1-\gamma_{z-1}} \right) 
    + \sum_{z=1}^Z (\gamma_z^{k_z} - \gamma_{z-1}^{k_z}) \sum_{u=0}^{z-1} \Delta^{u}_{t-1} + \gamma_z^{k_z} \Delta^{z}_{t-1} \nonumber \\
    & = \underbrace{\epsilon \frac{1-\gamma_Z^{k_Z}}{1-\gamma_Z} + \epsilon \sum_{z=0}^{Z-1} \frac{\gamma_z^{k_{z+1}}-\gamma_z^{k_z}}{1-\gamma_z}}_{(\star) \text{variance term}}
    + \underbrace{\sum_{z=1}^Z (\gamma_z^{k_z} - \gamma_{z-1}^{k_z}) \sum_{u=0}^{z-1} \Delta^{u}_{t-1} + \gamma_z^{k_z} \Delta^{z}_{t-1}}_{(\star \star) \text{bias term}} 
\end{align}
Let's focus now further on the bias term $(\star \star)$:
\begin{align}
    \sum_{z=1}^Z (\gamma_z^{k_z} - \gamma_{z-1}^{k_z}) & \sum_{u=0}^{z-1} \Delta^{u}_{t-1} + \gamma_z^{k_z} \Delta^{z}_{t-1} 
     = \sum_{u=0}^{Z-1} \sum_{z=u+1}^Z (\gamma_z^{k_z} - \gamma_{z-1}^{k_z}) \Delta^{u}_{t-1} + \sum_{z=1}^Z \gamma_z^{k_z} \Delta^{z}_{t-1} \nonumber \\
    & = \sum_{u=0}^{Z-1} \Delta^{u}_{t-1} \left( \sum_{z=u+1}^Z \gamma_z^{k_z} - \sum_{z=u}^{Z-1} \gamma_{z}^{k_{z+1}}\right) + \sum_{z=1}^Z \gamma_z^{k_z} \Delta^{z}_{t-1} \nonumber \\
    & = \sum_{u=0}^{Z-1} \Delta^{u}_{t-1} \left( \sum_{z=u+1}^{Z-1} (\gamma_z^{k_z} - \gamma_{z}^{k_{z+1}} ) + \gamma_Z^{k_Z} - \gamma_u^{k_{u+1}} \right) + \sum_{z=1}^Z \gamma_z^{k_z} \Delta^{z}_{t-1} \nonumber \\
    & = \sum_{u=0}^{Z-1} \sum_{z=u+1}^{Z-1} (\gamma_z^{k_z} - \gamma_{z}^{k_{z+1}} ) \Delta^{u}_{t-1} + \gamma_Z^{k_Z}\sum_{z=0}^{Z} \Delta^{z}_{t-1} + \sum_{z=0}^{Z-1} (\gamma_z^{k_z} - \gamma_{z}^{k_{z+1}}) \Delta^{z}_{t-1} \nonumber \\
    & = \sum_{u=0}^{Z-1} \sum_{z=u}^{Z-1} (\gamma_z^{k_z} - \gamma_{z}^{k_{z+1}} ) \Delta^{u}_{t-1} + \gamma_Z^{k_Z}\sum_{z=0}^{Z} \Delta^{z}_{t-1} \nonumber \\
    & = \sum_{z=0}^{Z-1} (\gamma_z^{k_z} - \gamma_{z}^{k_{z+1}} ) \sum_{u=0}^{z} \Delta^{u}_{t-1} + \gamma_Z^{k_Z}\sum_{z=0}^{Z} \Delta^{z}_{t-1}
\end{align}

Finally, we obtain:

\begin{equation}
    \sum_{z=0}^Z  \Delta^{z}_t 
    \leq \underbrace{\epsilon \frac{1-\gamma^{k}}{1-\gamma} + \epsilon \sum_{z=0}^{Z-1} \frac{\gamma_z^{k_{z+1}}-\gamma_z^{k_z}}{1-\gamma_z}}_{ \text{variance term}}
    + \underbrace{\sum_{z=0}^{Z-1} (\gamma_z^{k_z} - \gamma_{z}^{k_{z+1}} ) \sum_{u=0}^{z} \Delta^{u}_{t-1} + \gamma^{k}\sum_{z=0}^{Z} \Delta^{z}_{t-1}}_{ \text{bias term}} 
\end{equation}

\section{Experiments}

All code can be found in: \href{https://github.com/facebookresearch/td-delta}{github.com/facebookresearch/td-delta}.
\subsection{Tabular}
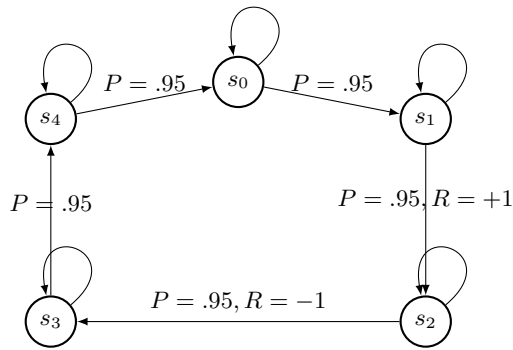
\begin{figure}
\centering
  \begin{tikzpicture}[auto,node distance=8mm,>=latex,font=\small]

    \tikzstyle{round}=[thick,draw=black,circle]

    \node[round] (s0) {$s_0$};
    \node[round,below right=0mm and 20mm of s0] (s1) {$s_1$};
    \node[round,below=20mm of s1] (s2) {$s_2$};
    \node[round,below left=0mm and 20mm of s0] (s4) {$s_4$};
    \node[round,below=20mm of s4] (s3) {$s_3$};

    \draw[->] (s0) -- node [above] {$P=.95$} (s1);
    \draw[->] (s1) -- node [above] {$P=.95, R=+1$} (s2);
    \draw[->] (s2) -- node [above] {$P=.95, R=-1$} (s3);
    \draw[->] (s3) -- node [above] {$P=.95$} (s4);
    \draw[->] (s4) -- node [above] {$P=.95$} (s0);

    \draw[->] (s0) [out=40,in=100,loop] to node [very near start, anchor=center] (m2) {} (s0);
    \draw[->] (s1) [out=40,in=100,loop] to node [very near start, anchor=center] (m2) {}  (s1);
    \draw[->] (s2) [out=40,in=100,loop] to node [very near start, anchor=center] (m2) {}  (s2);
    \draw[->] (s3) [out=40,in=100,loop] to node [very near start, anchor=center] (m2) {}  (s3);
    \draw[->] (s4) [out=40,in=100,loop] to node [very near start, anchor=center] (m2) {}  (s4);

\end{tikzpicture}
\label{fig:ring}
\caption{5-state ring MDP as in Kearns}
\end{figure}

We use the 5-state ring MDP in \citet{kearns2000bias}. In each state there's a transition probability of .95 to the next state and .05 of staying in the current state. Two adjacent states in the environment have a +1 and -1 reward respectively. Example in Figure~\ref{fig:ring}.

We use value iteration to find the optimal value function. We run an open-source version of value iteration (VI) based on \href{https://github.com/Breakend/ValuePolicyIterationVariations}{github.com/Breakend/ValuePolicyIterationVariations} from the algorithm details in \citet{bellman1957markovian,sutton1998introduction}. 
After learning the optimal value through VI, we use TD and TD($\Delta$) to learn estimated values by sampling.
We terminate each learning run at 5000 steps. We run 200 experiments per setting with different random seeds affecting the transition dynamics (random seeds $0-199$ used). 

\subsubsection{Results}

\begin{figure}[!htbp]
    \centering
        \includegraphics[width=.45\textwidth]{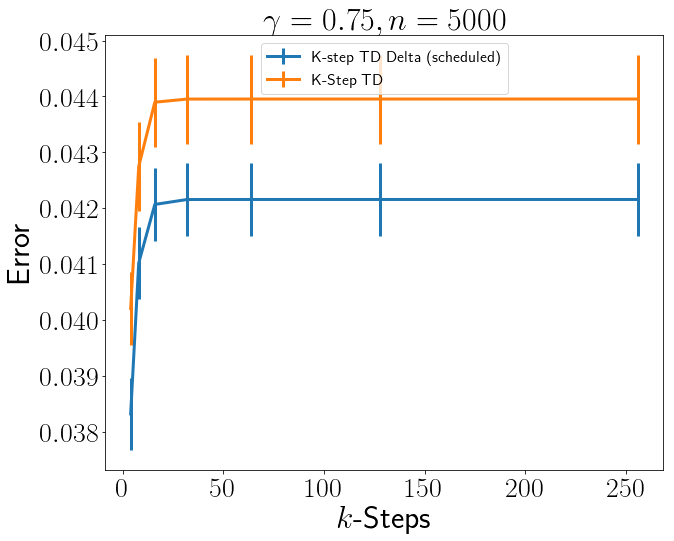}
        \includegraphics[width=.45\textwidth]{plots_stderr_tabular_ksteps_new_g0_93750aggregaten5000.png}
        \includegraphics[width=.45\textwidth]{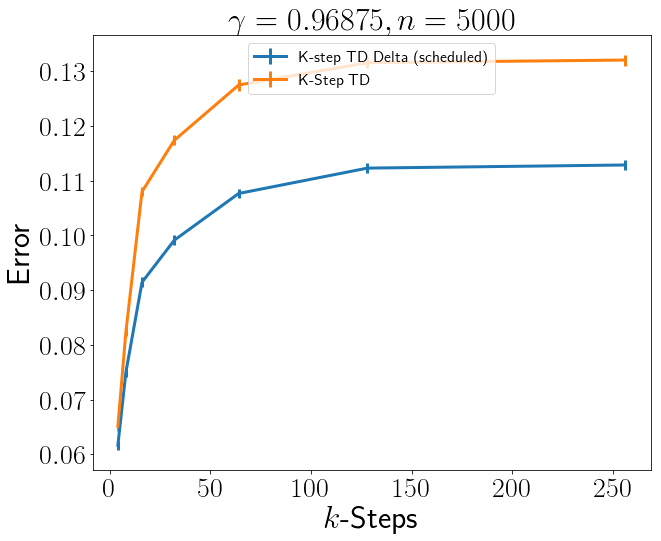}
        \includegraphics[width=.45\textwidth]{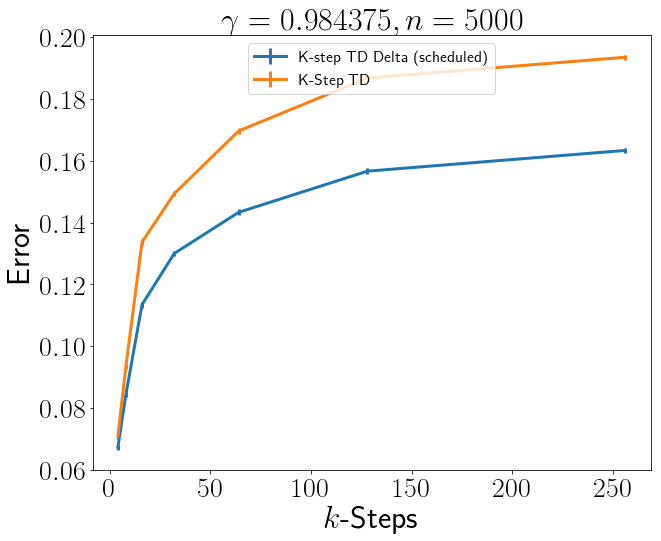}
        \includegraphics[width=.45\textwidth]{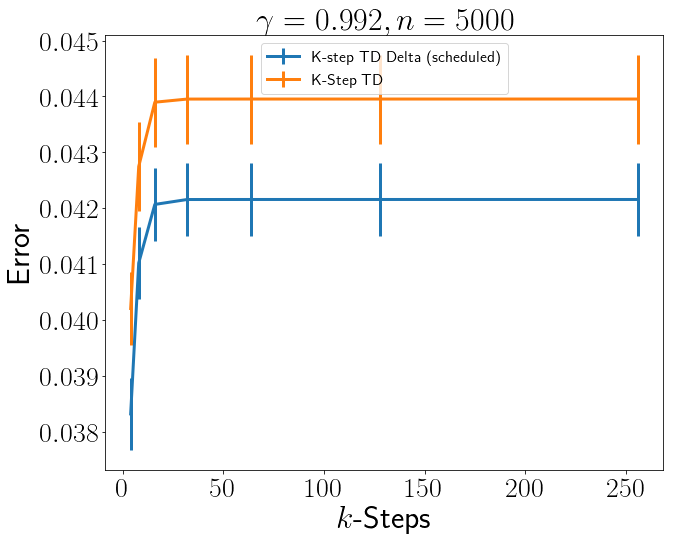}
        \includegraphics[width=.45\textwidth]{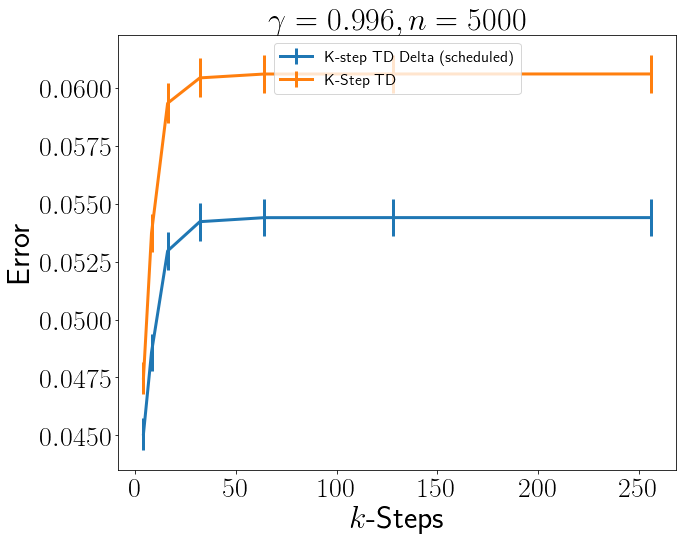}
    \caption{A comparison of the best performing learning rate at each $\gamma$ value, $n=5000$.}
    \label{fig:aggregates_tabular}
\end{figure}

\begin{figure}[!htbp]
    \centering
        \includegraphics[width=.45\textwidth]{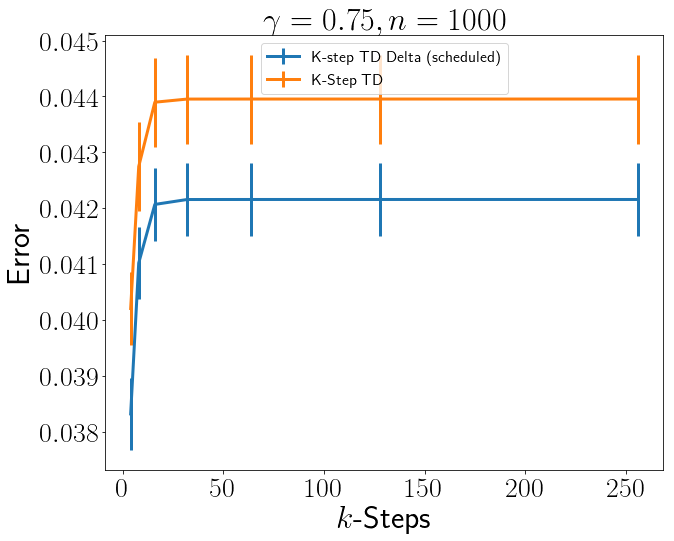}
        \includegraphics[width=.45\textwidth]{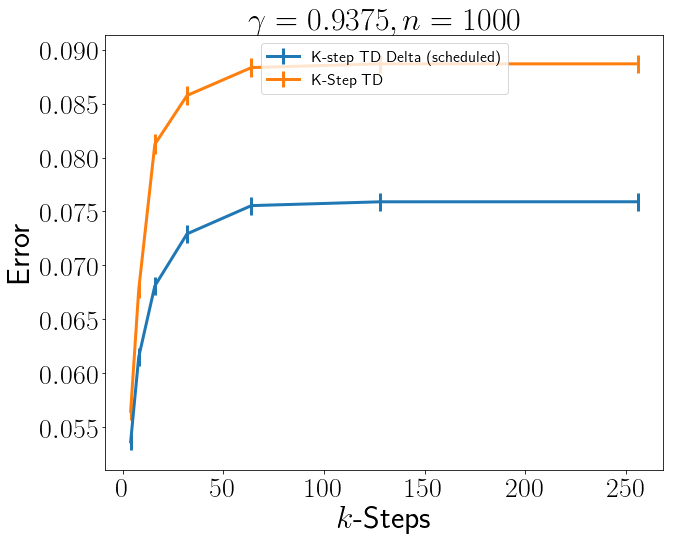}
        \includegraphics[width=.45\textwidth]{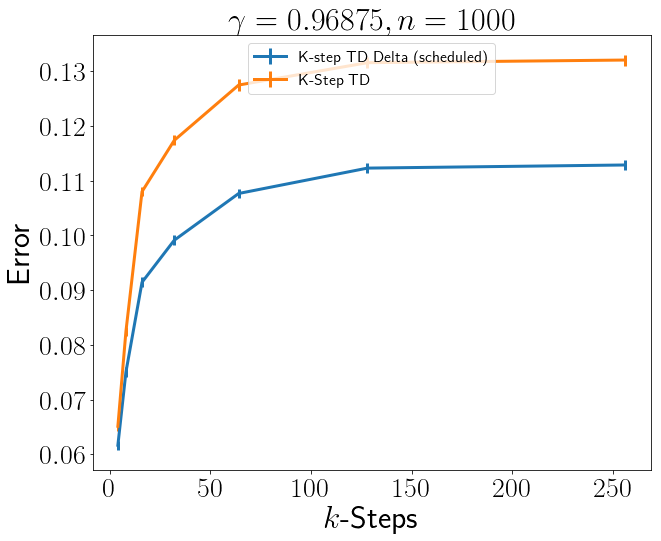}
        \includegraphics[width=.45\textwidth]{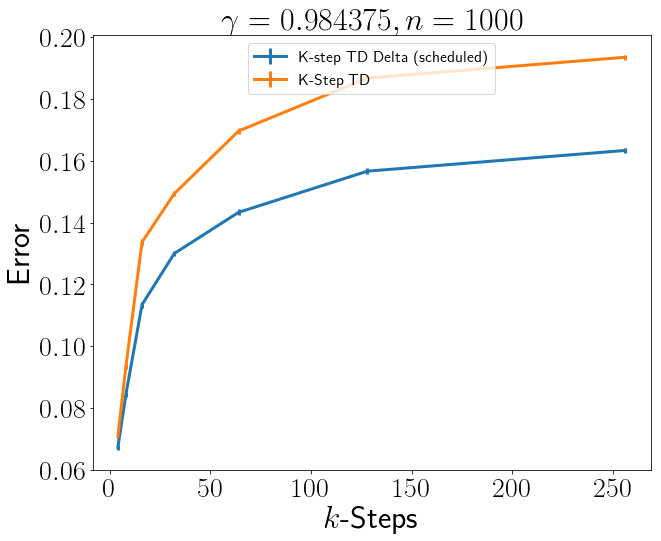}
        \includegraphics[width=.45\textwidth]{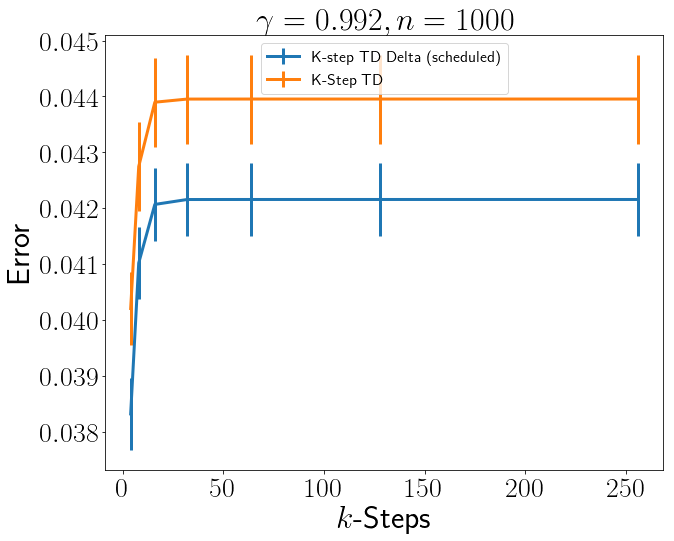}
        \includegraphics[width=.45\textwidth]{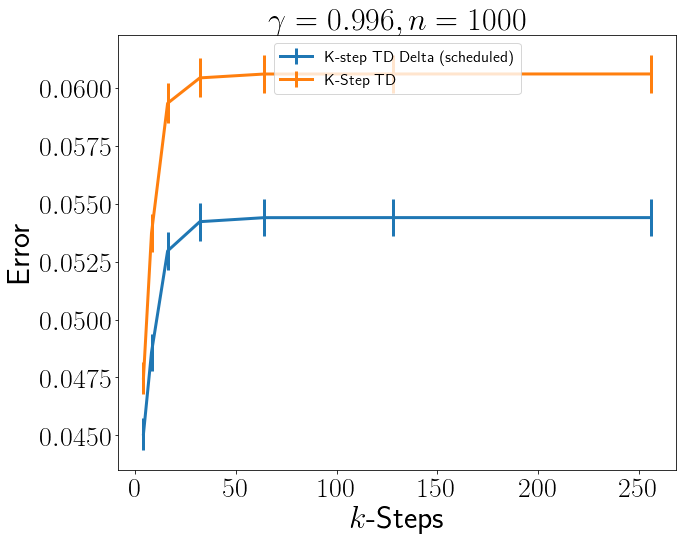}
    \caption{A comparison of the best performing learning rate at each $\gamma$ value, $n=1000$.}
    \label{fig:aggregates_tabular1000}
\end{figure}

\begin{figure}[!htbp]
    \centering
        \includegraphics[width=.45\textwidth]{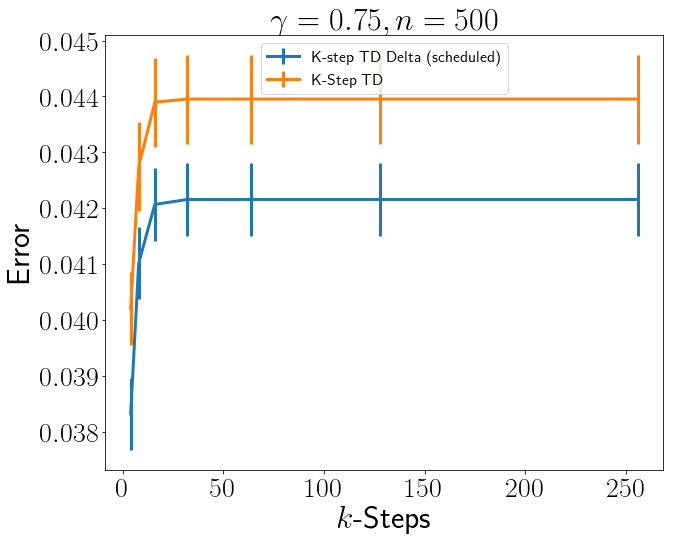}
        \includegraphics[width=.45\textwidth]{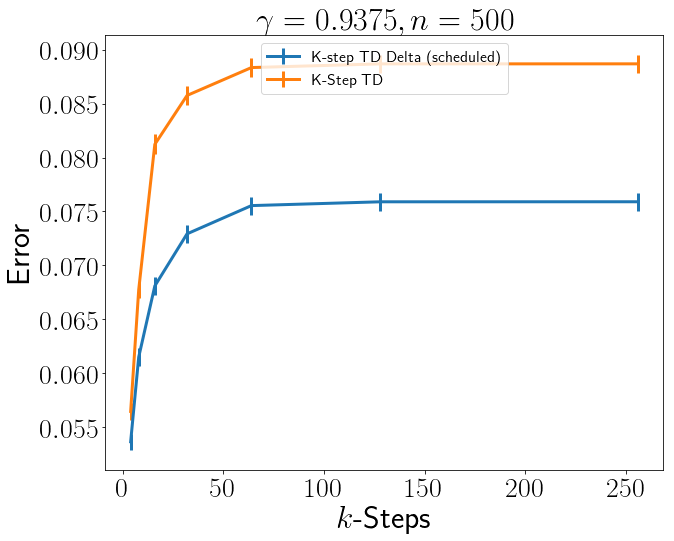}
        \includegraphics[width=.45\textwidth]{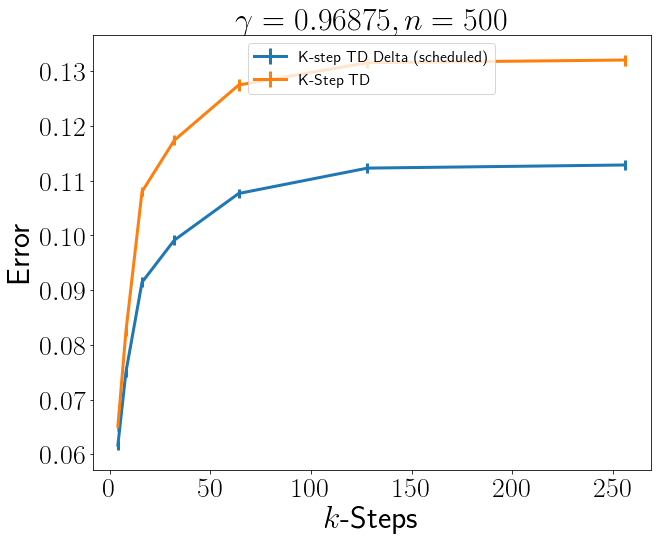}
        \includegraphics[width=.45\textwidth]{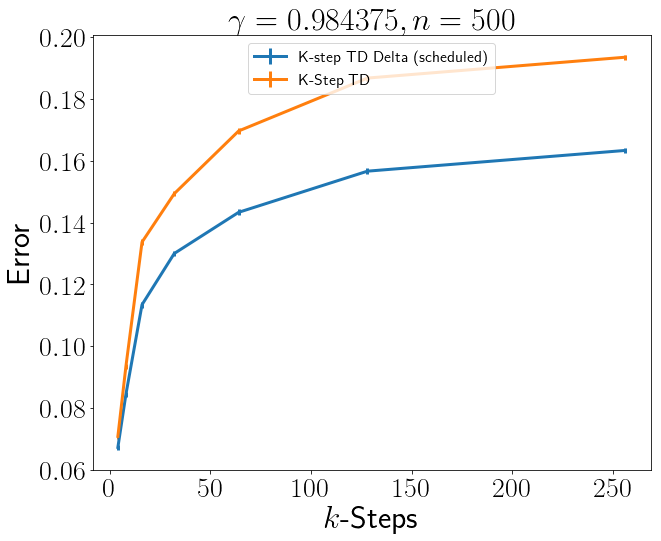}
        \includegraphics[width=.45\textwidth]{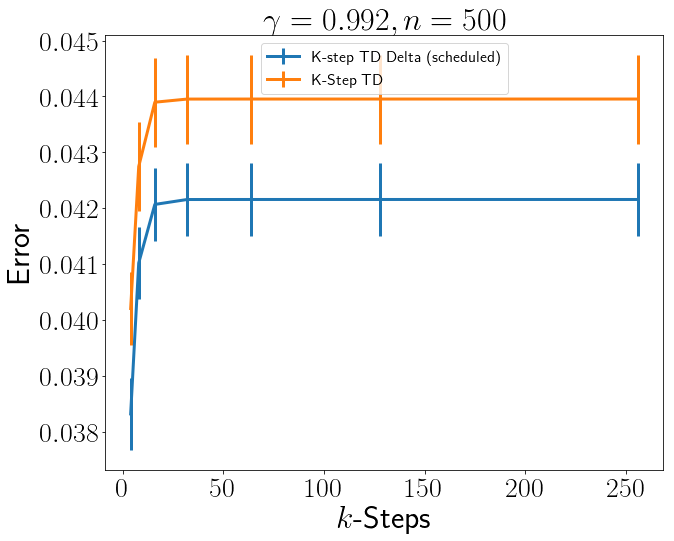}
        \includegraphics[width=.45\textwidth]{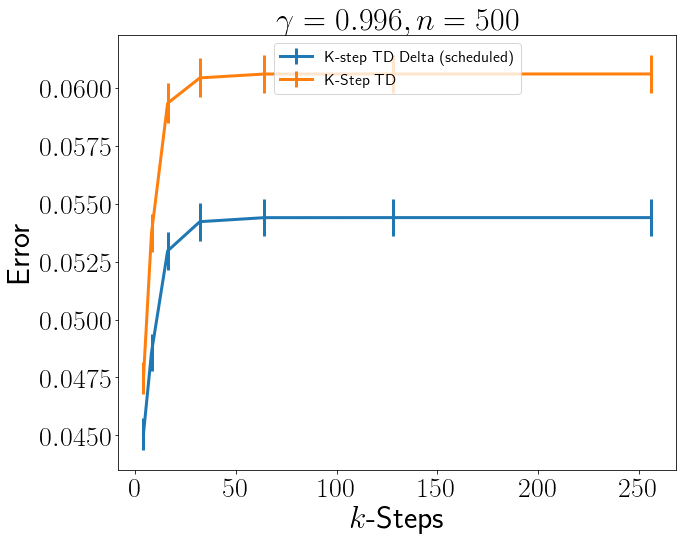}
    \caption{A comparison of the best performing learning rate at each $\gamma$ value, $n=500$.}
    \label{fig:aggregates_tabular500}
\end{figure}

\begin{figure}[!htbp]
    \centering
        \includegraphics[width=.45\textwidth]{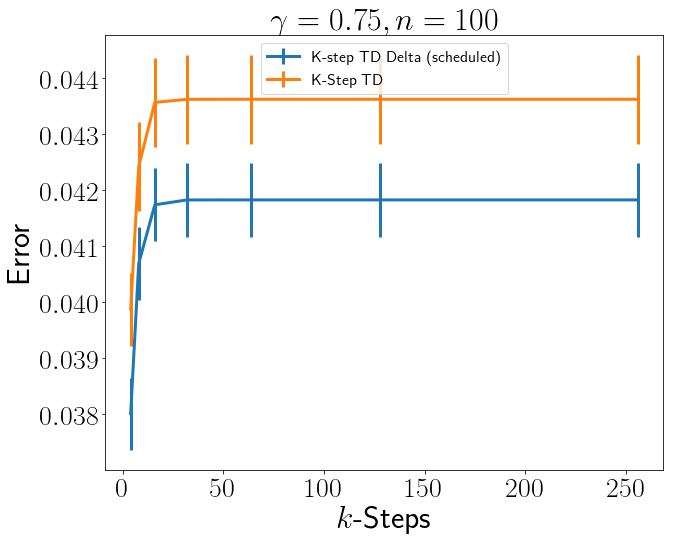}
        \includegraphics[width=.45\textwidth]{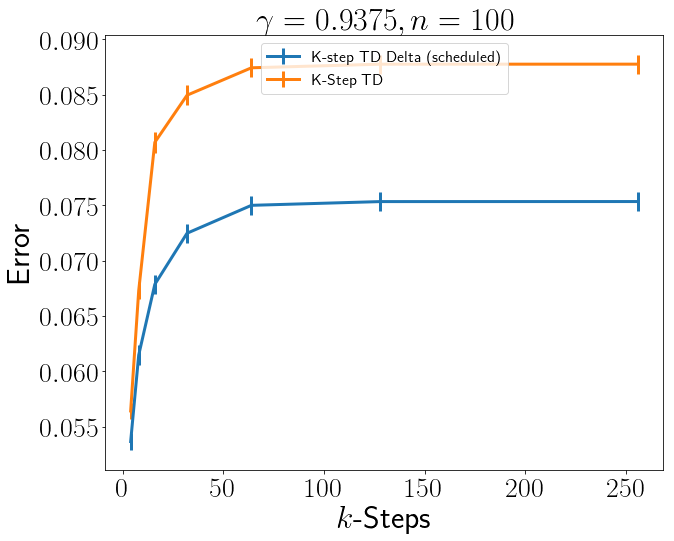}
        \includegraphics[width=.45\textwidth]{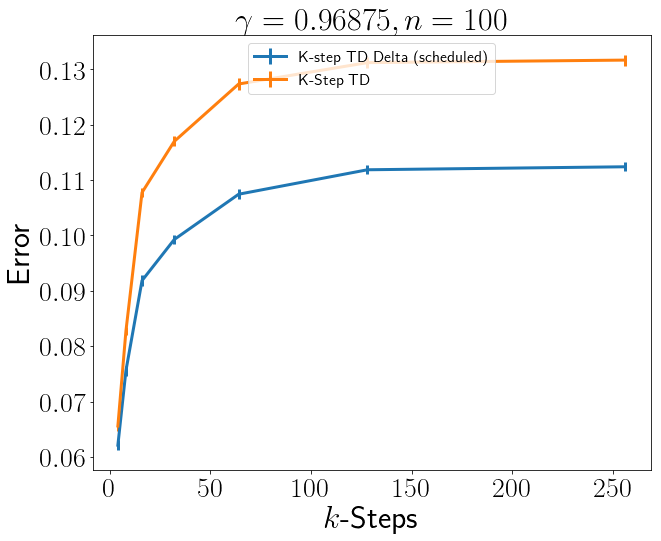}
        \includegraphics[width=.45\textwidth]{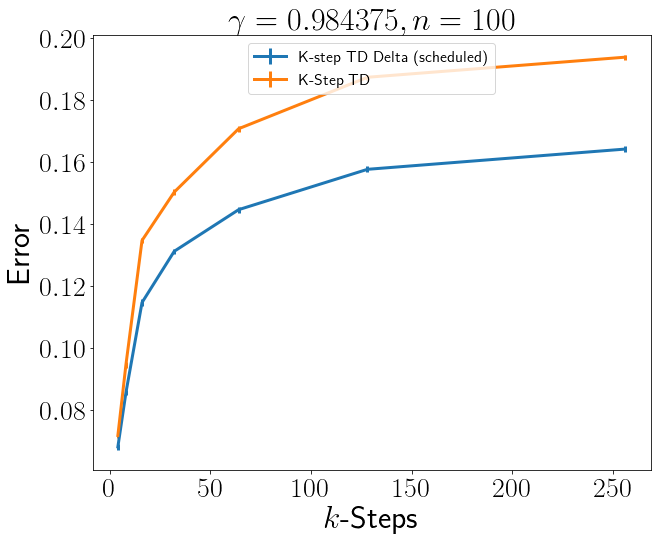}
        \includegraphics[width=.45\textwidth]{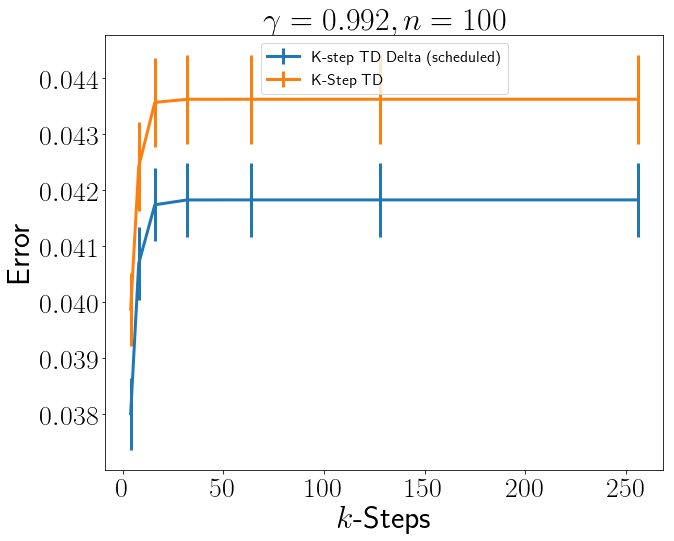}
        \includegraphics[width=.45\textwidth]{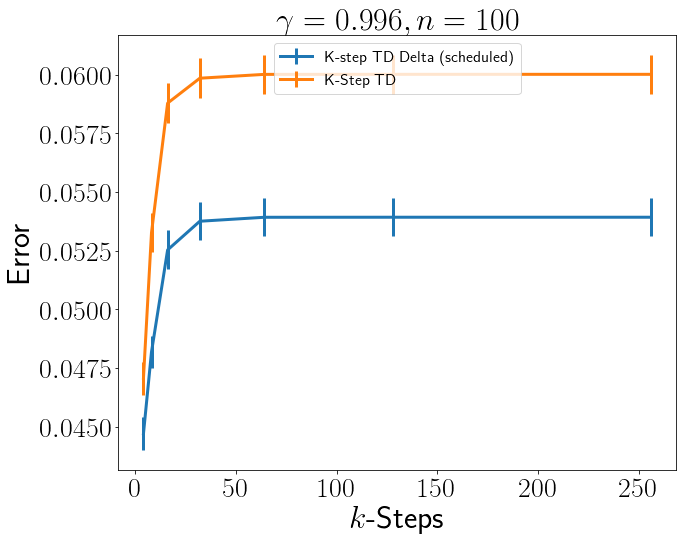}
    \caption{A comparison of the best performing learning rate at each $\gamma$ value, $n=100$.}
    \label{fig:aggregates_tabular100}
\end{figure}

For tabular results please see Figures below for the average return stopped at different timesteps $n$. Figures~\ref{fig:aggregates_tabular}, \ref{fig:aggregates_tabular1000}, \ref{fig:aggregates_tabular500}, \ref{fig:aggregates_tabular100} distill these experiments such that the best performing learning rate is compared across each discount factor and $n$. All other figures demonstrate the distribution of average return across the learning trajectory for all learning rate, $k$, $n$ combinations evaluated.

As can be seen in Figure~\ref{fig:aggregates_tabular}, TD($\Delta$) learns a better than or equal to value function estimate (in terms of error) in the time frame on average across all timesteps in all scenarios. 
We see a trend that performance gains increase with increasing $\gamma$ and $k$ values. This is intuitive from the theoretical bias-variance trade-off induced by TD($\Delta$) and demonstrates the benefit of using TD($\Delta$) in long horizon settings. 
We also note that in this particular MDP, error increases with $k$ due to the bias-variance trade-off of $k$-step returns consistent with \citet{kearns2000bias}.

\begin{figure}[!htbp]
    \centering
    \includegraphics[width=.3\textwidth]{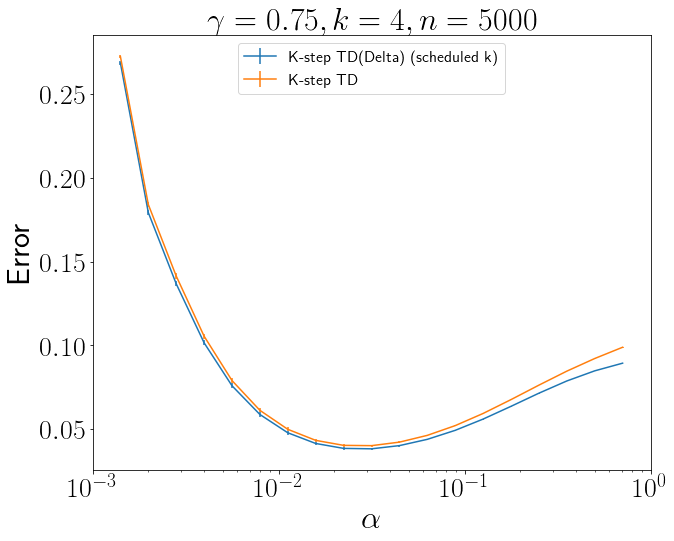}
    \includegraphics[width=.3\textwidth]{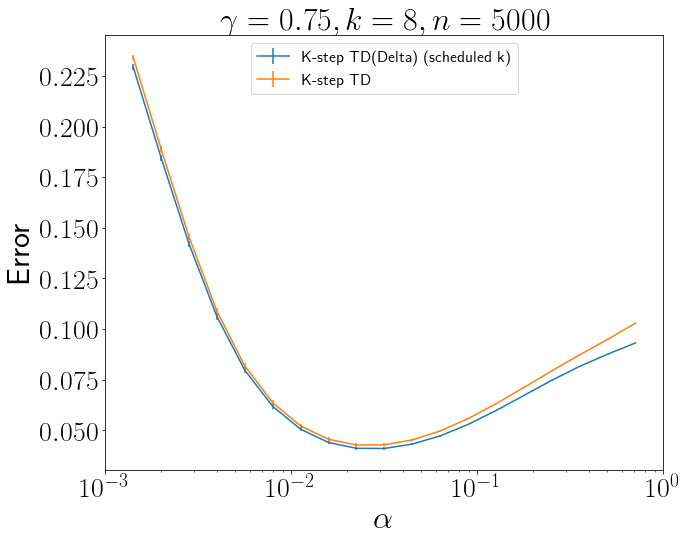}
    \includegraphics[width=.3\textwidth]{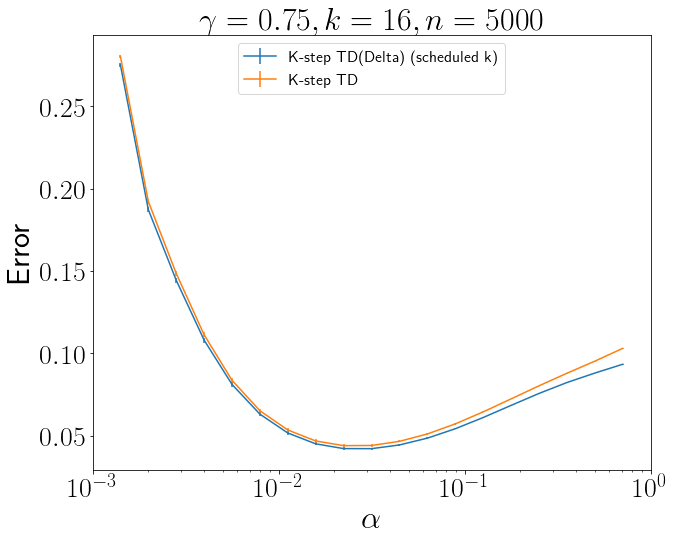}
    \includegraphics[width=.3\textwidth]{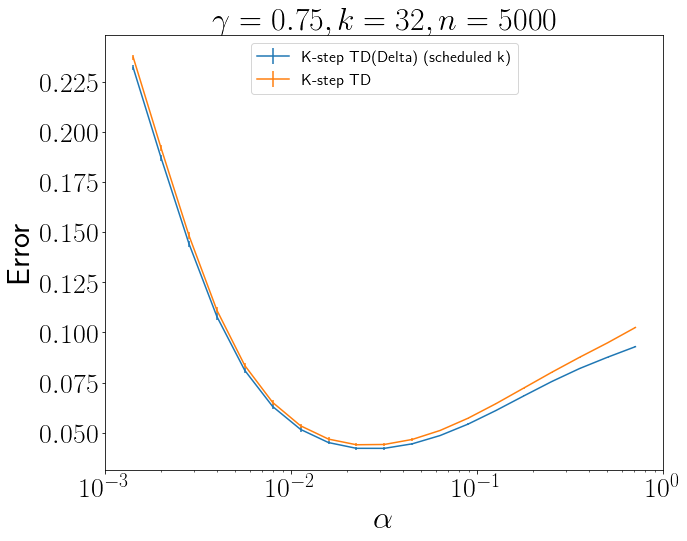}
    \includegraphics[width=.3\textwidth]{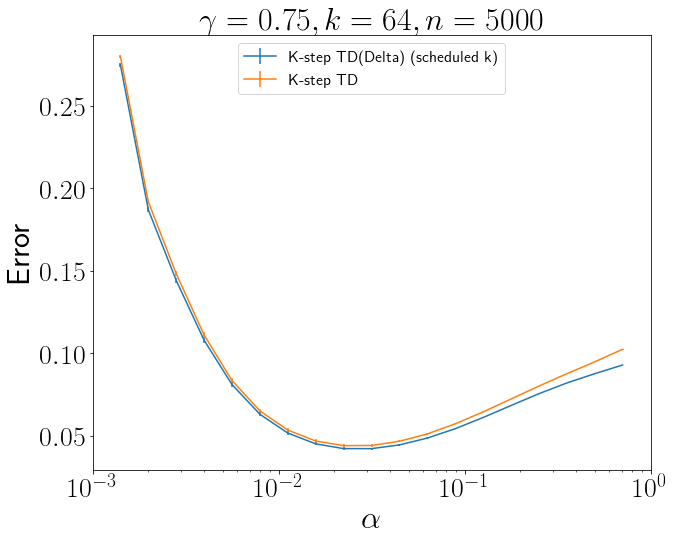}
    \includegraphics[width=.3\textwidth]{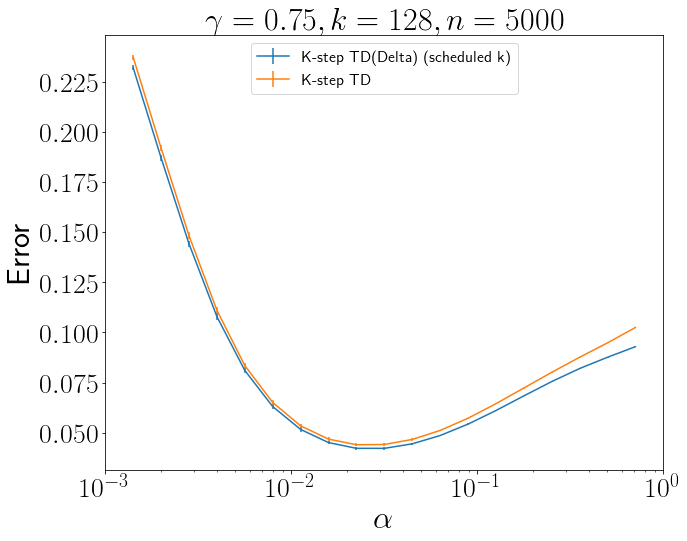}
    \includegraphics[width=.3\textwidth]{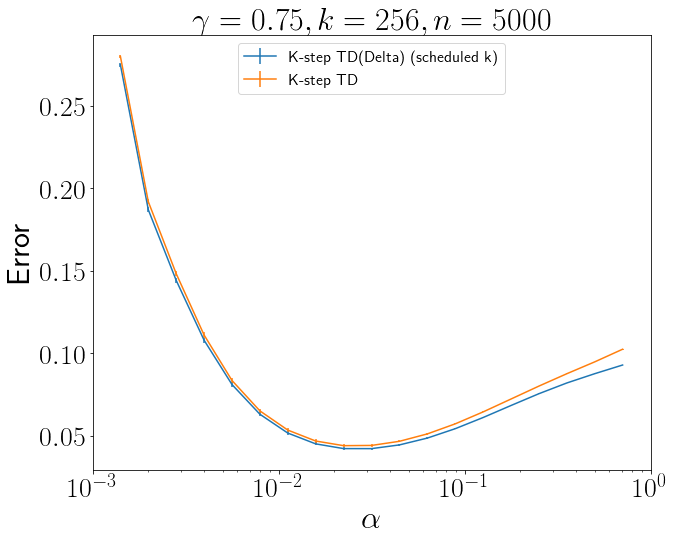}
    \caption{$\gamma=.75$ and different k values at different learning rates, where the number of timesteps $n=5000$.}
    \label{fig:g_75}
\end{figure}

\begin{figure}[!htbp]
    \centering
    \includegraphics[width=.3\textwidth]{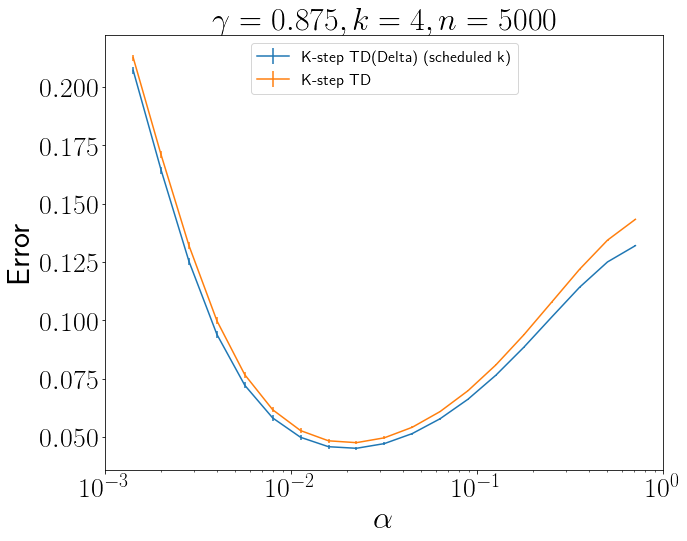}
    \includegraphics[width=.3\textwidth]{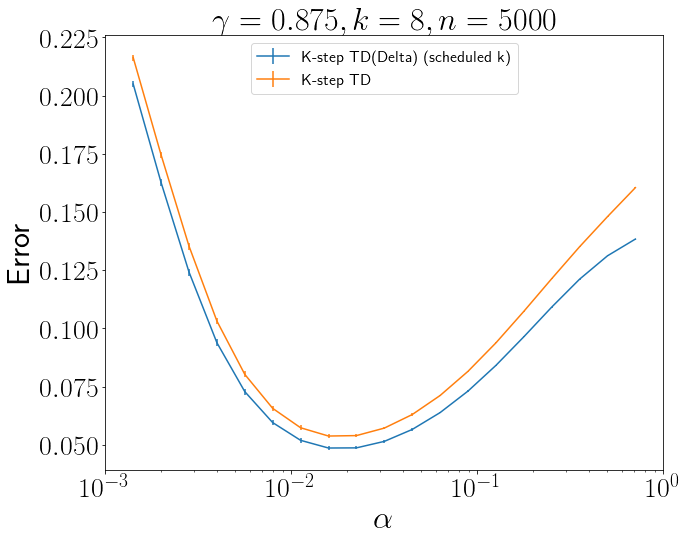}
    \includegraphics[width=.3\textwidth]{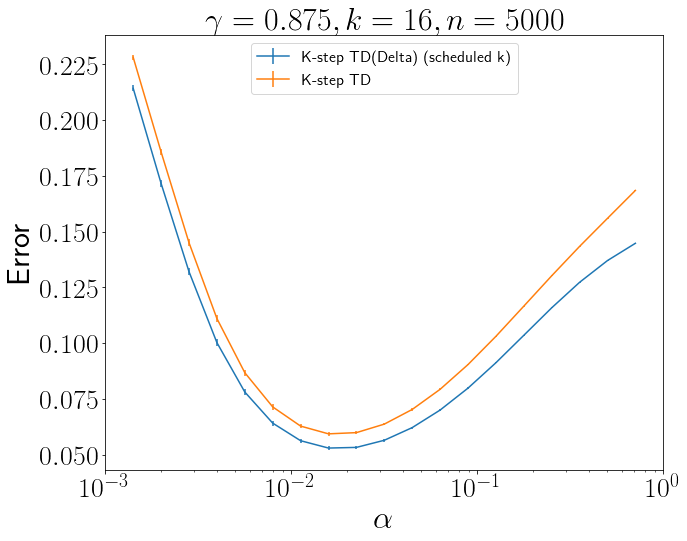}
    \includegraphics[width=.3\textwidth]{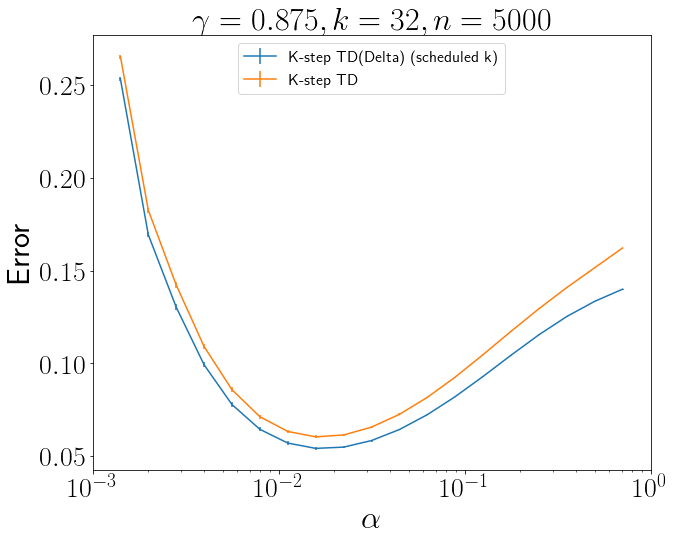}
    \includegraphics[width=.3\textwidth]{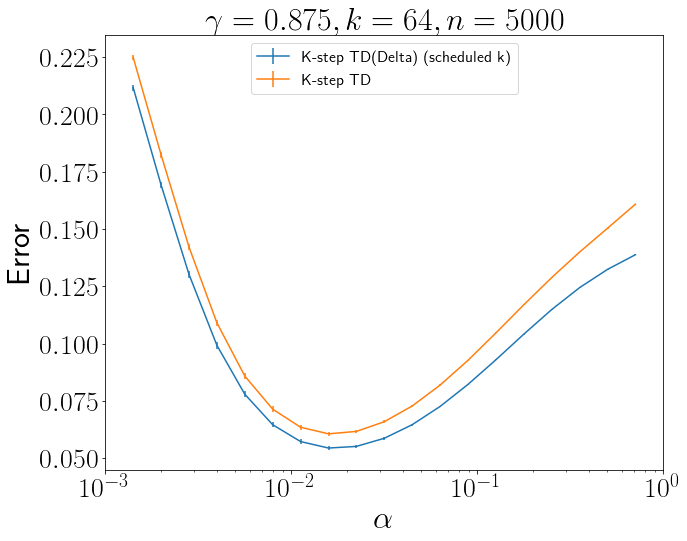}
    \includegraphics[width=.3\textwidth]{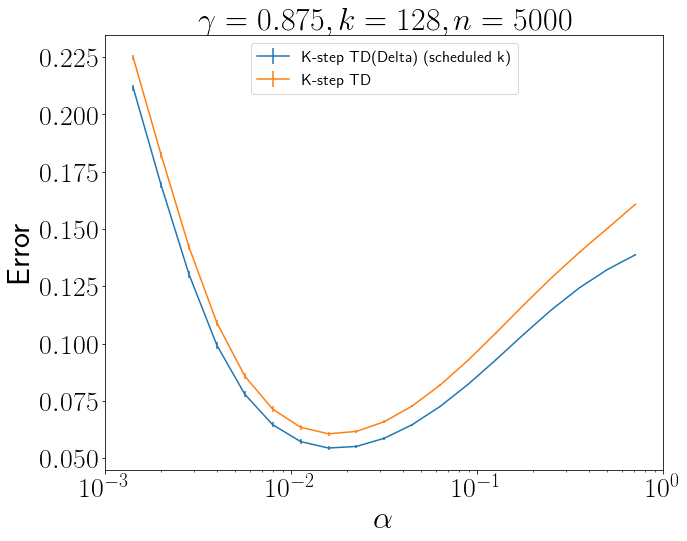}
    \includegraphics[width=.3\textwidth]{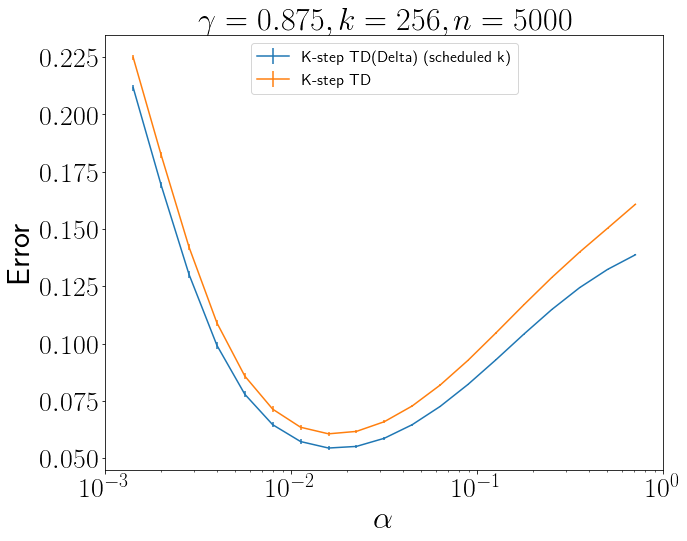}
    \caption{$\gamma=.875$ and different k values at different learning rates, where the number of timesteps $n=5000$.}
    \label{fig:g875}
\end{figure}

\begin{figure}[!htbp]
    \centering
    \includegraphics[width=.3\textwidth]{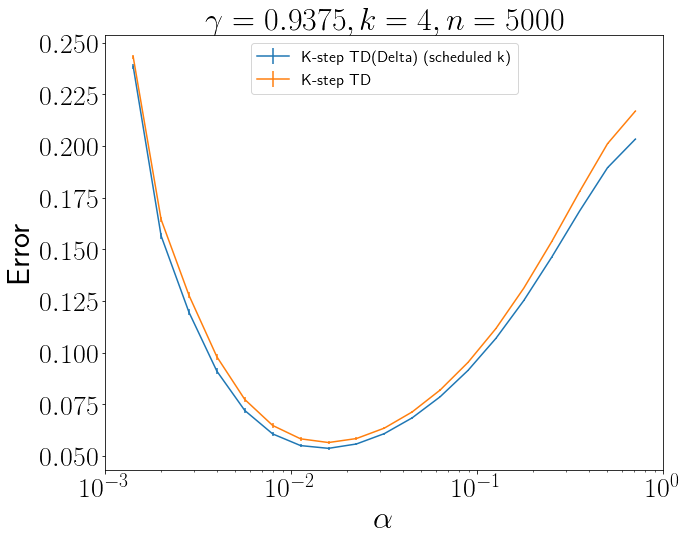}
    \includegraphics[width=.3\textwidth]{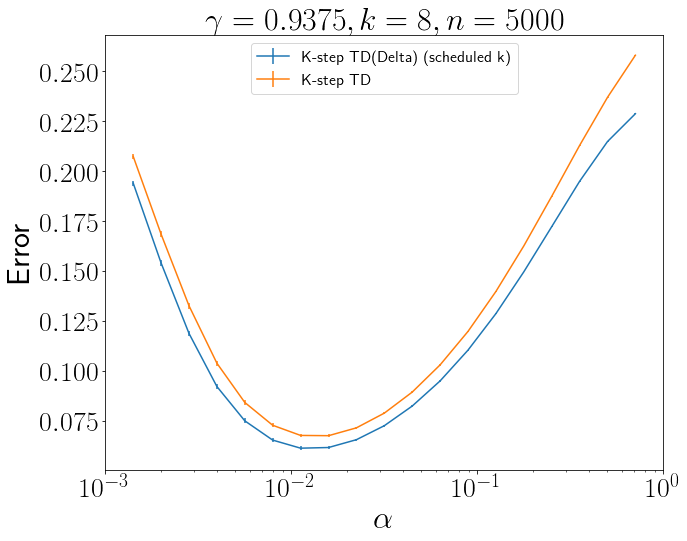}
    \includegraphics[width=.3\textwidth]{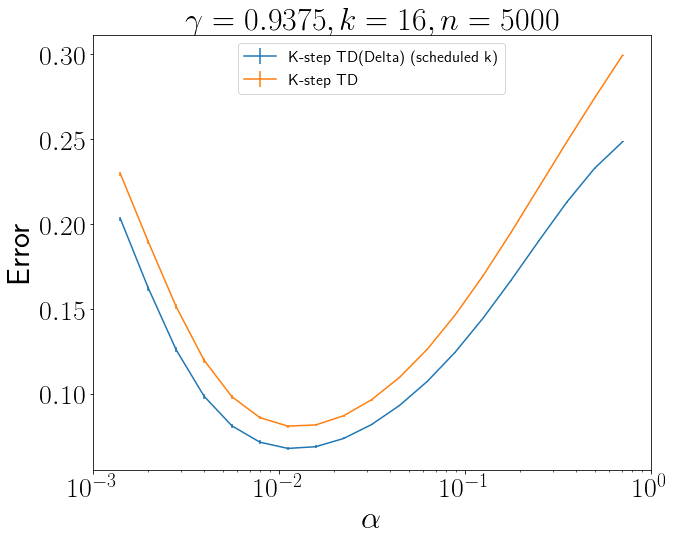}
    \includegraphics[width=.3\textwidth]{plots_stderr_tabular_ksteps_new_g0937500runs200k32iter5000.png}
    \includegraphics[width=.3\textwidth]{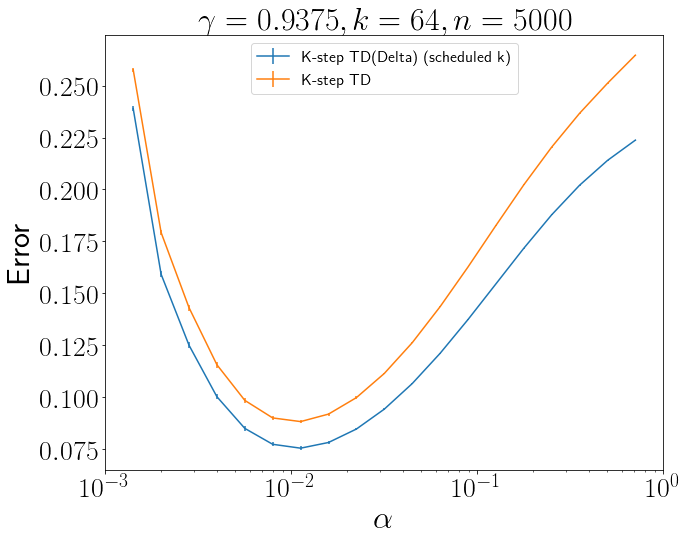}
    \includegraphics[width=.3\textwidth]{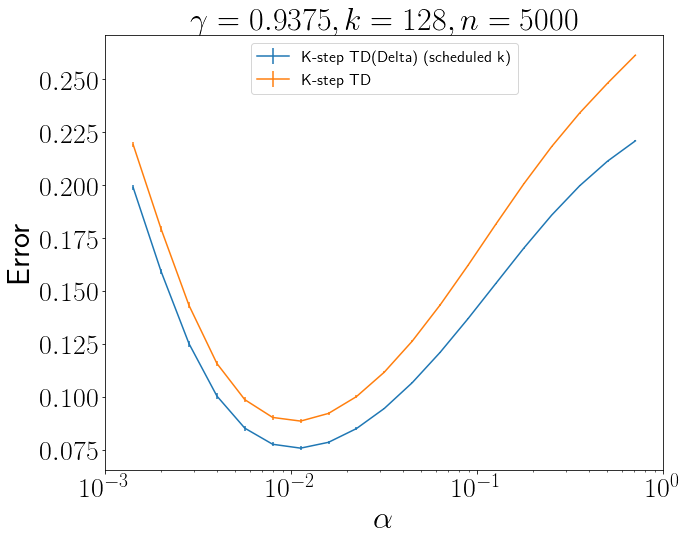}
    \includegraphics[width=.3\textwidth]{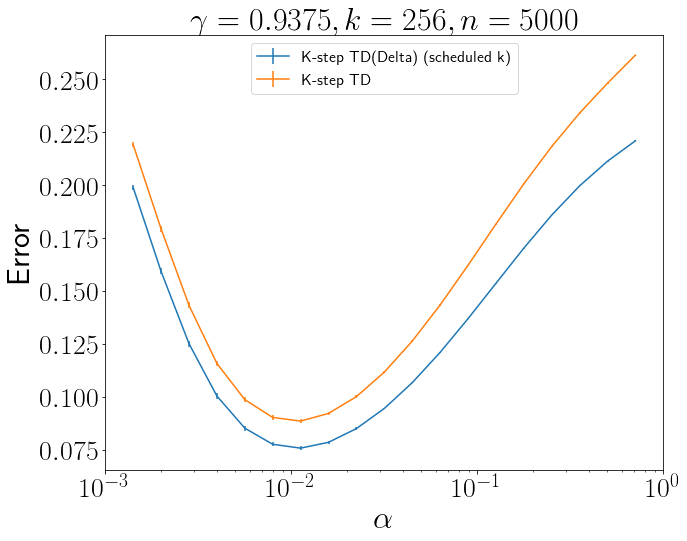}
    \caption{$\gamma=.93750$ and different k values at different learning rates, where the number of timesteps $n=5000$.}
    \label{fig:g9375}
\end{figure}

\begin{figure}[!htbp]
    \centering
        \includegraphics[width=.3\textwidth]{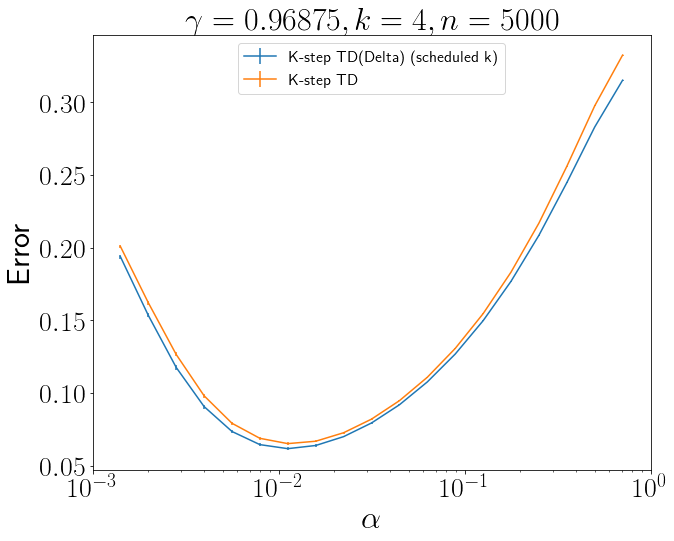}
    \includegraphics[width=.3\textwidth]{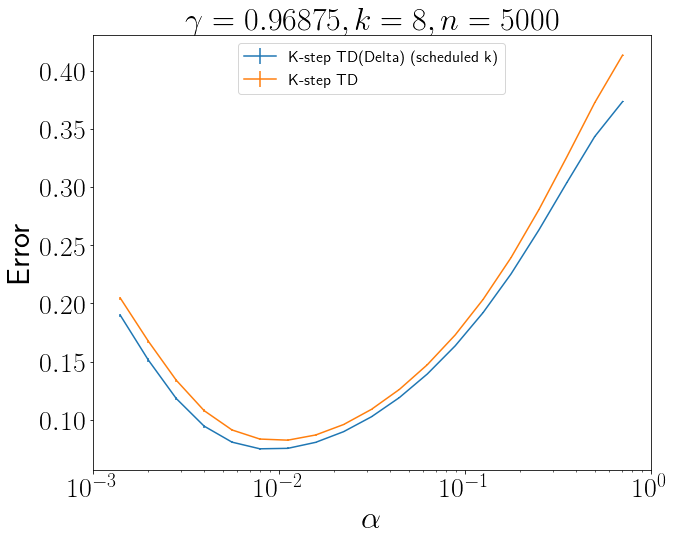}
    \includegraphics[width=.3\textwidth]{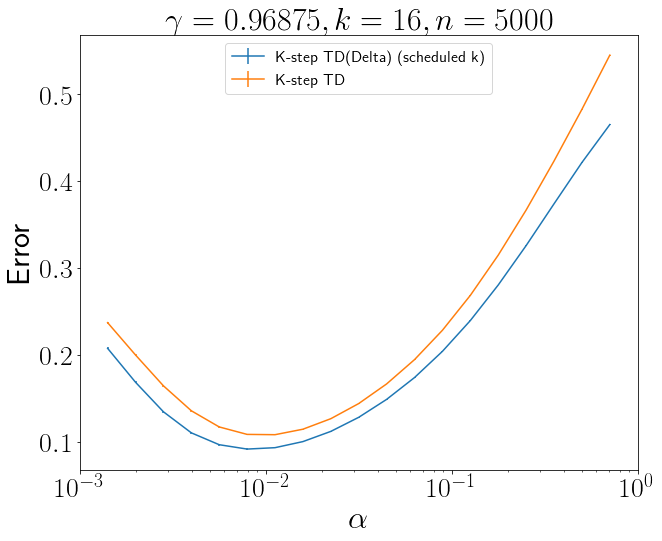}
    \includegraphics[width=.3\textwidth]{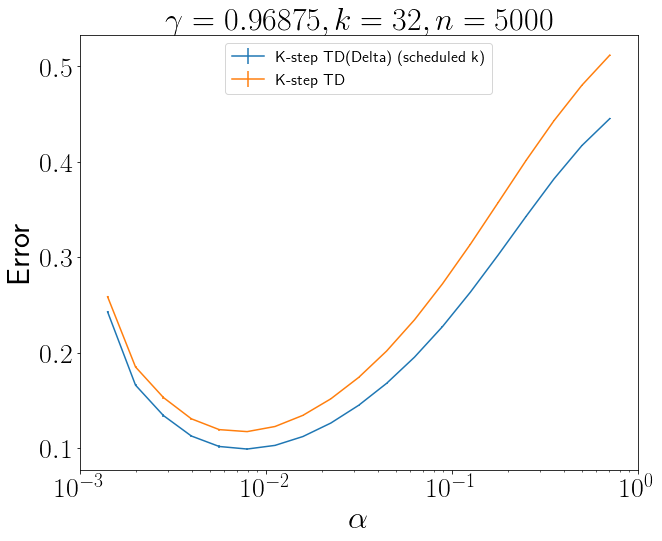}
    \includegraphics[width=.3\textwidth]{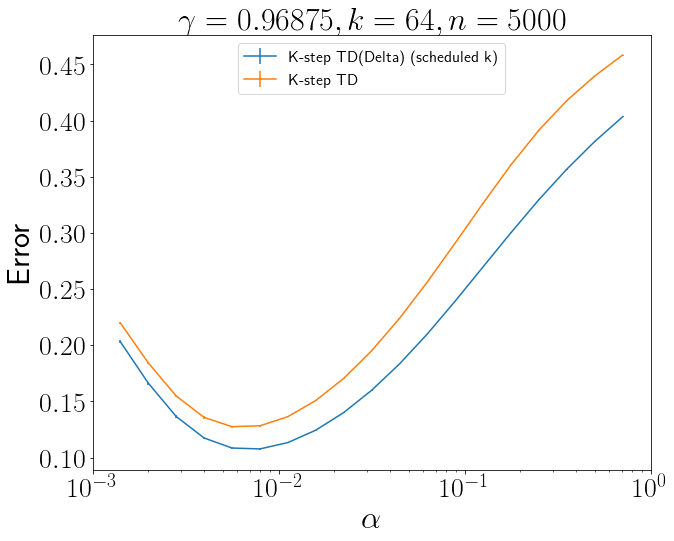}
    \includegraphics[width=.3\textwidth]{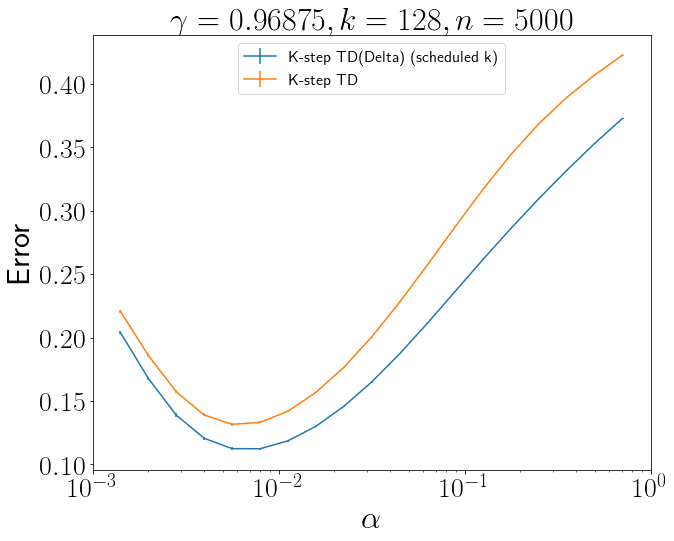}
    \includegraphics[width=.3\textwidth]{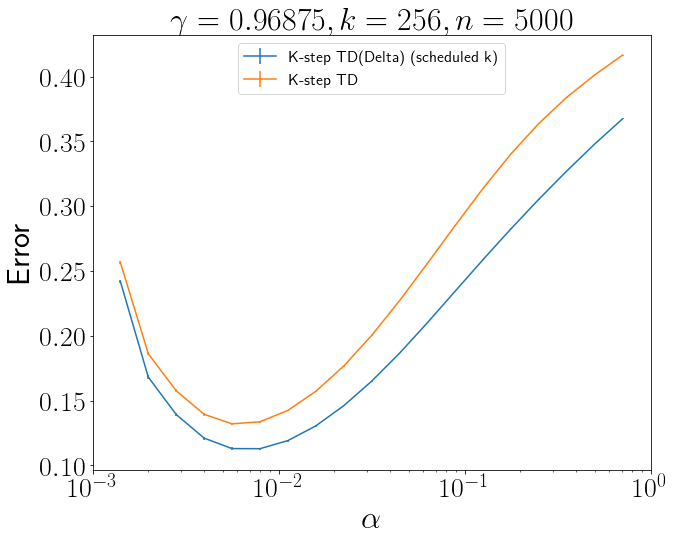}
    \caption{$\gamma=.96875$ and different k values at different learning rates, where the number of timesteps $n=5000$.}
    \label{fig:g96875}
\end{figure}

\begin{figure}[!htbp]
    \centering
        \includegraphics[width=.3\textwidth]{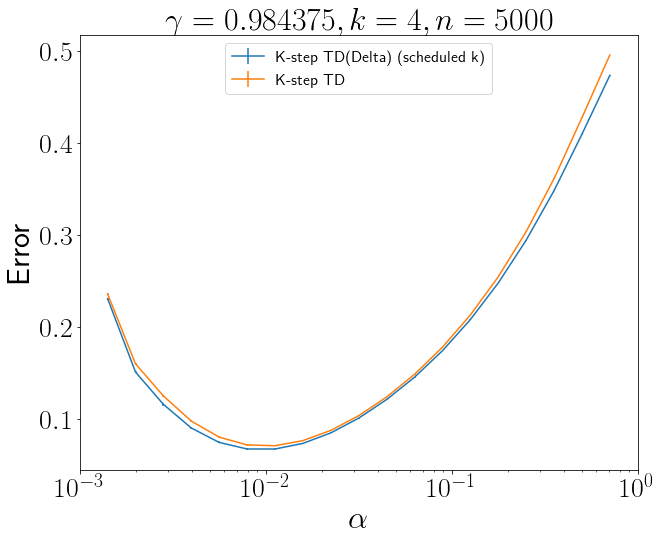}
    \includegraphics[width=.3\textwidth]{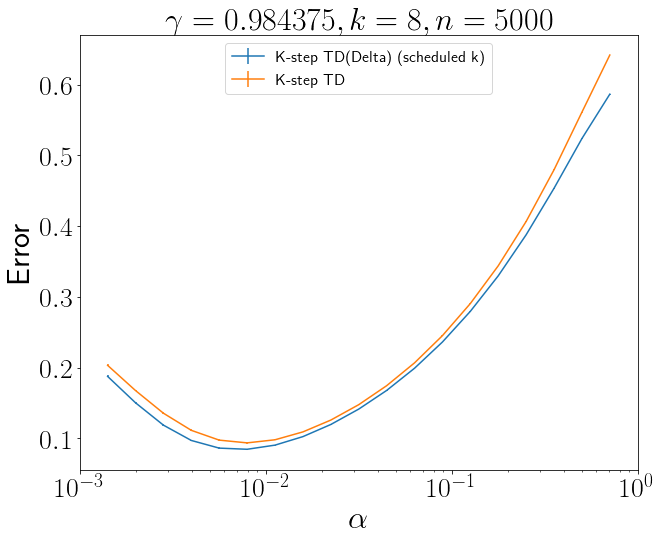}
    \includegraphics[width=.3\textwidth]{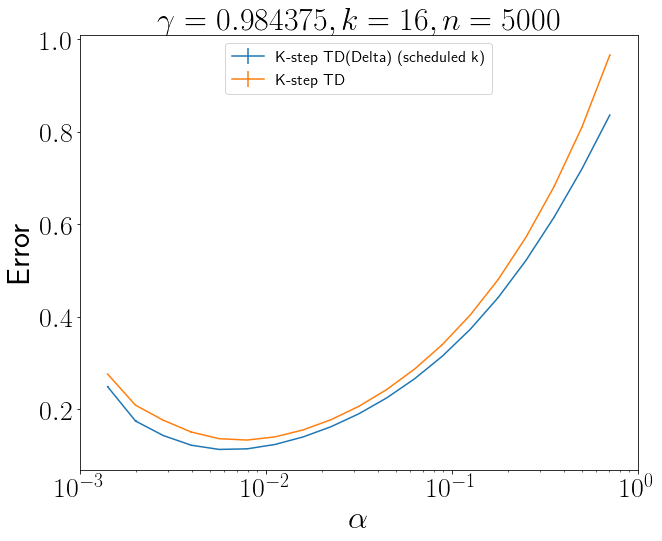}
    \includegraphics[width=.3\textwidth]{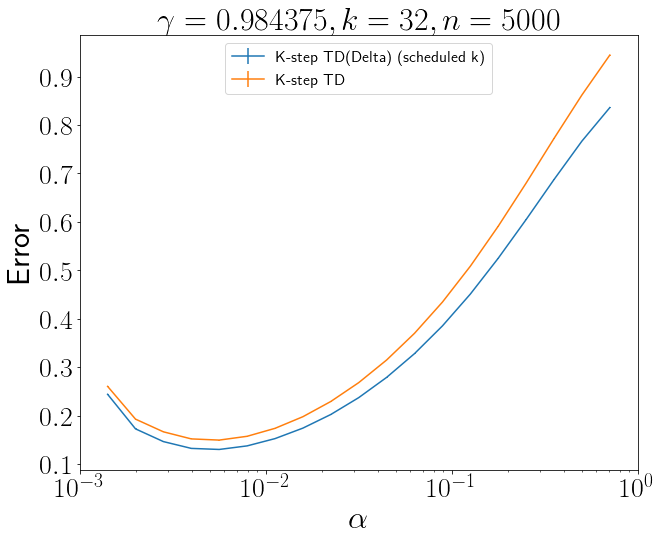}
    \includegraphics[width=.3\textwidth]{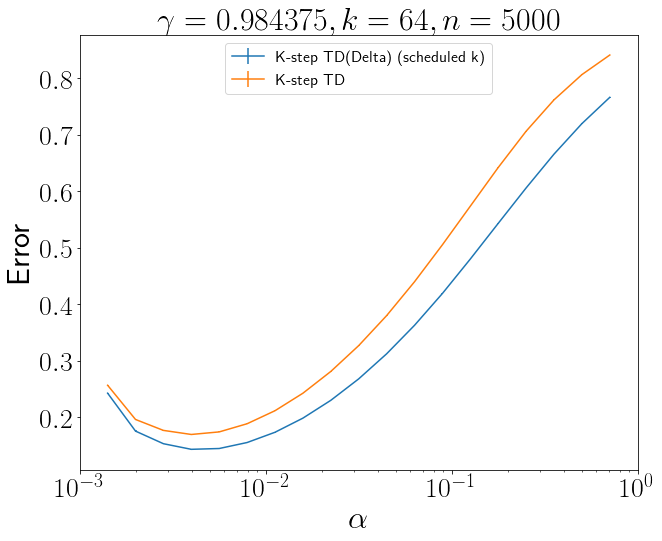}
    \includegraphics[width=.3\textwidth]{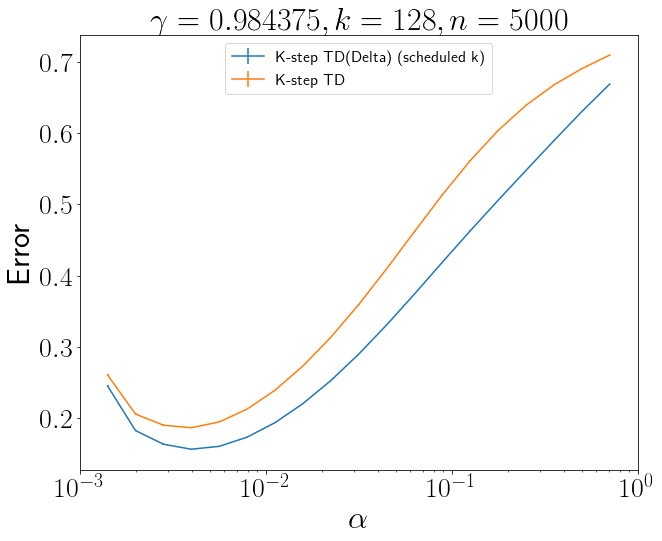}
    \includegraphics[width=.3\textwidth]{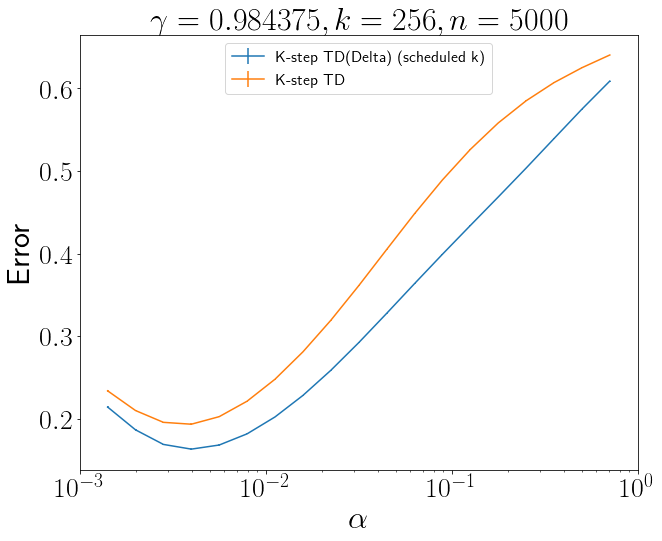}
    \caption{$\gamma=.98475$ and different k values at different learning rates, where the number of timesteps $n=5000$.}
    \label{fig:g984375}
\end{figure}

\begin{figure}[!htbp]
    \centering
        \includegraphics[width=.3\textwidth]{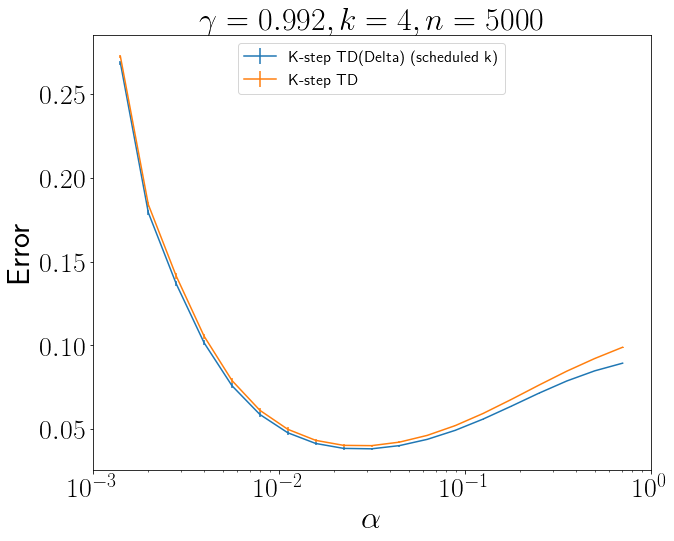}
    \includegraphics[width=.3\textwidth]{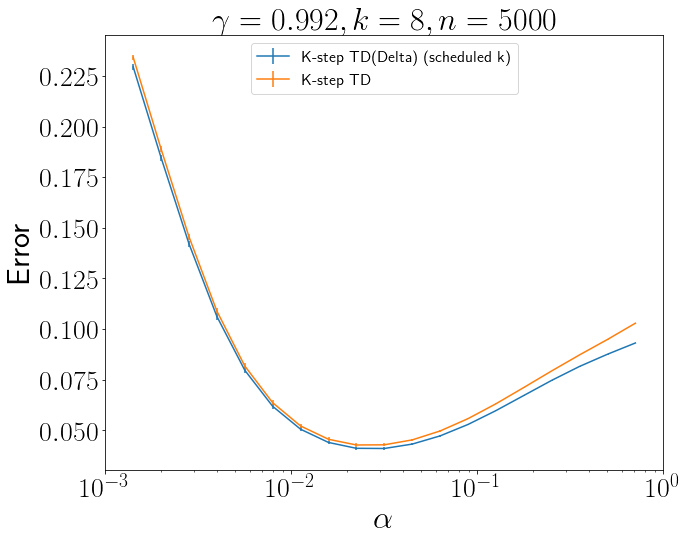}
    \includegraphics[width=.3\textwidth]{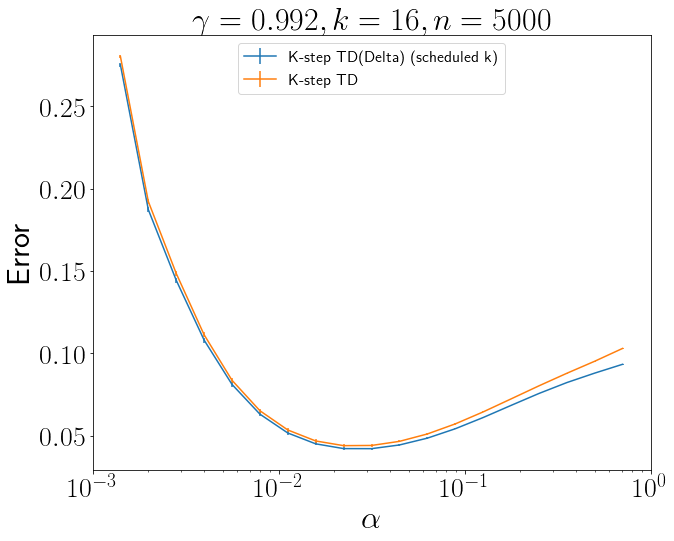}
    \includegraphics[width=.3\textwidth]{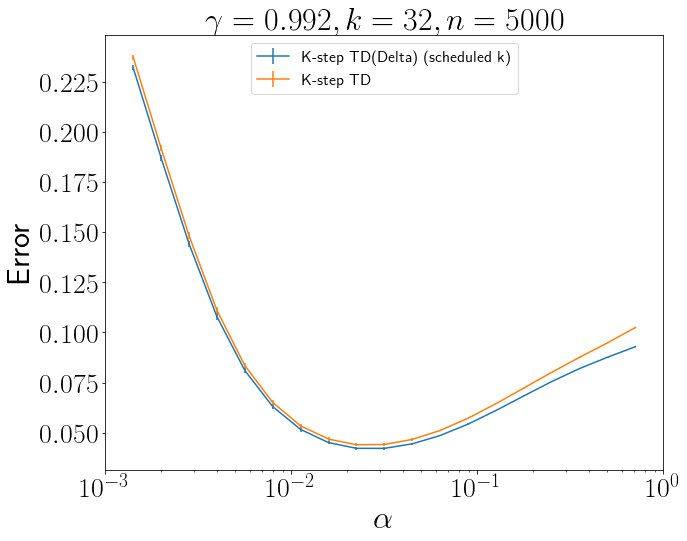}
    \includegraphics[width=.3\textwidth]{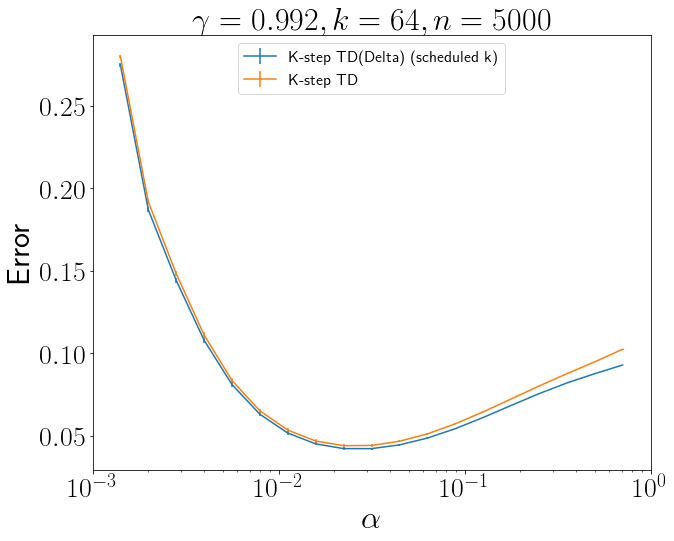}
    \includegraphics[width=.3\textwidth]{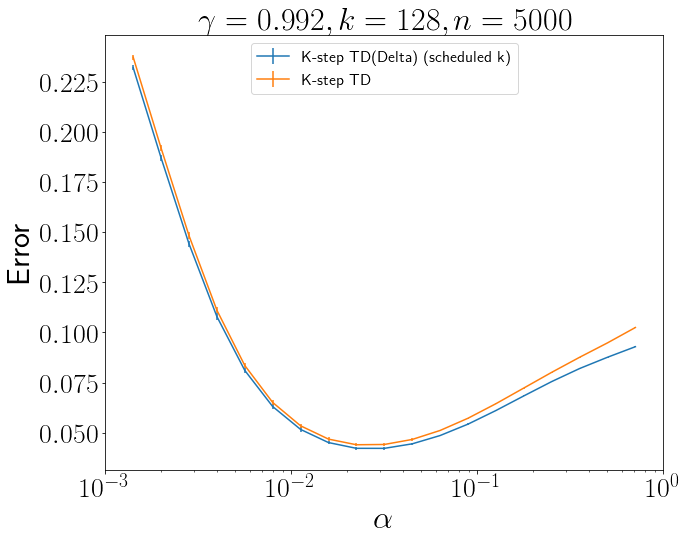}
    \includegraphics[width=.3\textwidth]{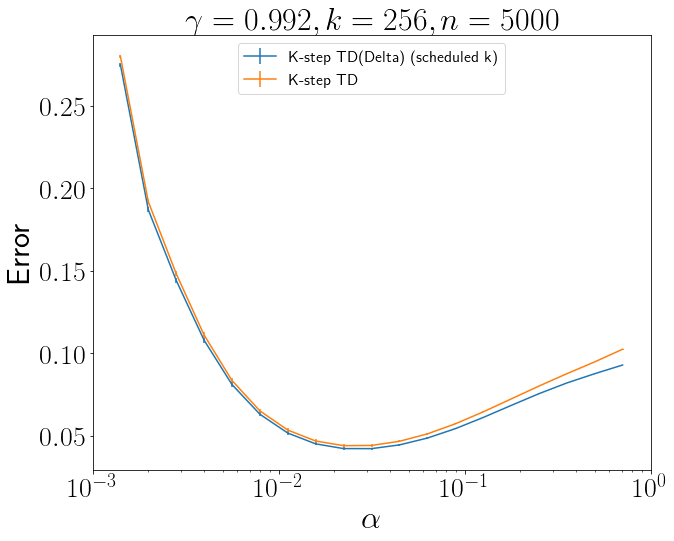}
    \caption{$\gamma=.992$ and different k values at different learning rates, where the number of timesteps $n=5000$.}
    \label{fig:g992}
\end{figure}

\begin{figure}[!htbp]
    \centering
        \includegraphics[width=.3\textwidth]{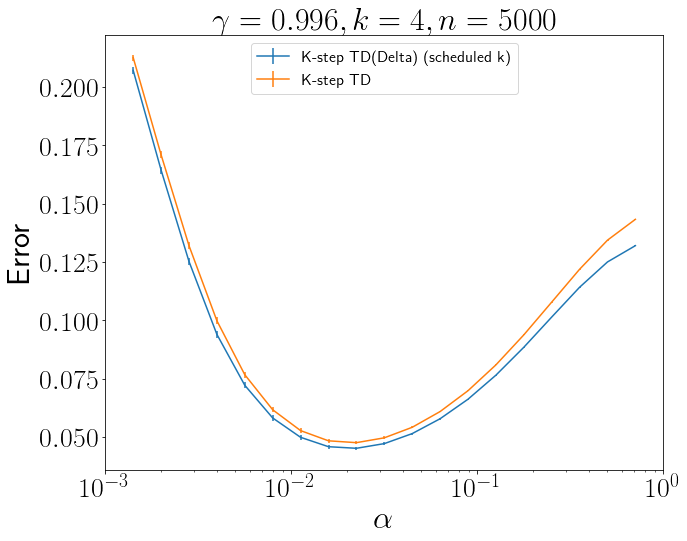}
    \includegraphics[width=.3\textwidth]{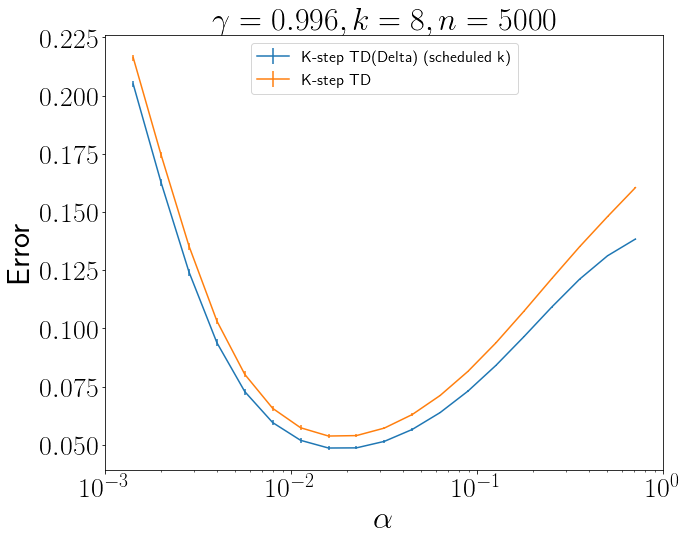}
    \includegraphics[width=.3\textwidth]{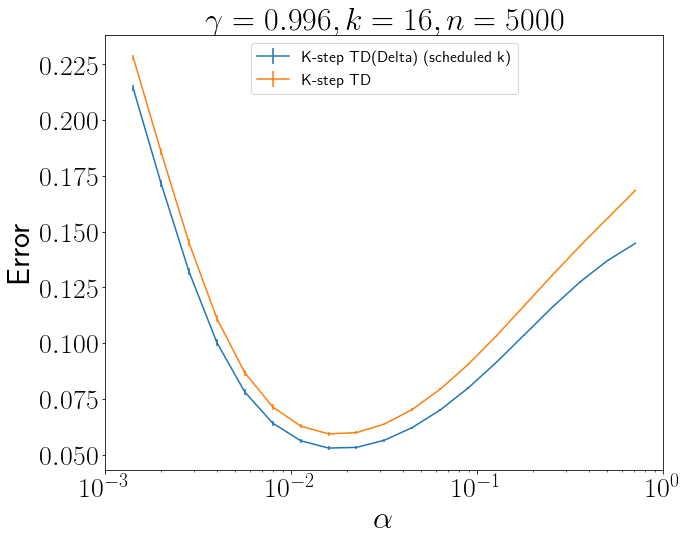}
    \includegraphics[width=.3\textwidth]{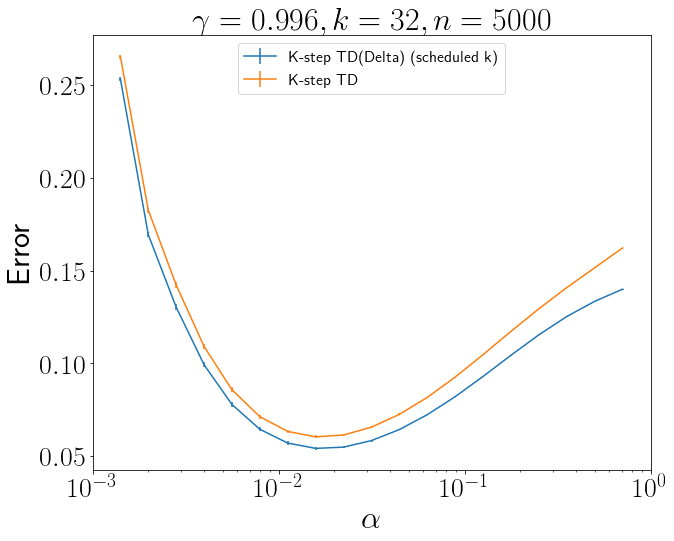}
    \includegraphics[width=.3\textwidth]{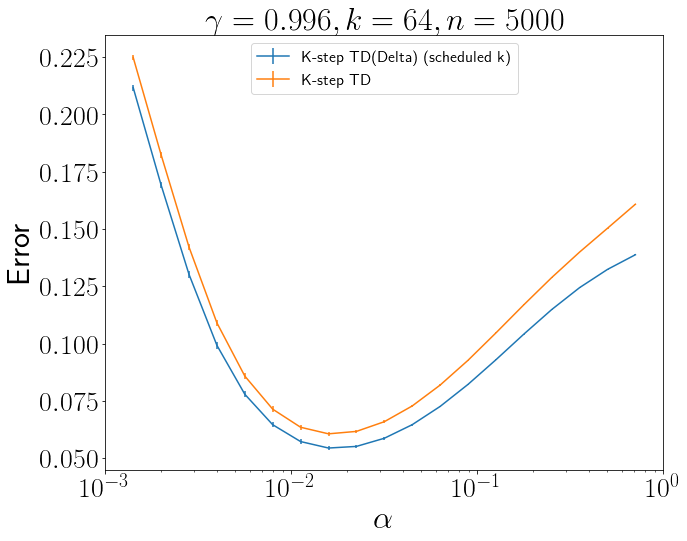}
    \includegraphics[width=.3\textwidth]{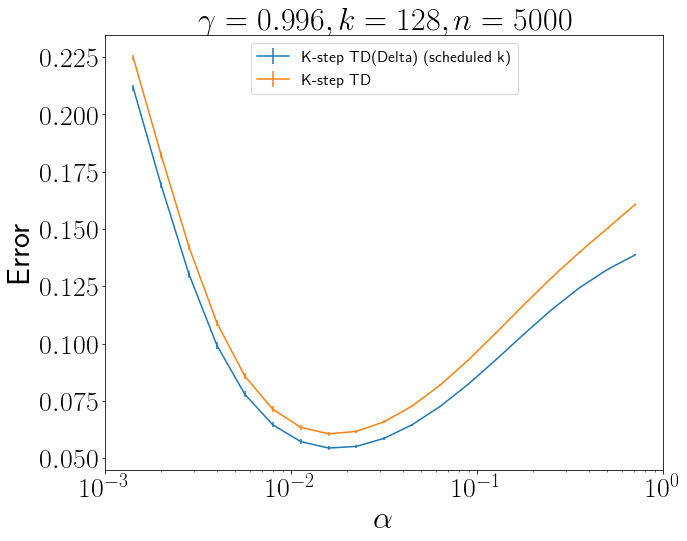}
    \includegraphics[width=.3\textwidth]{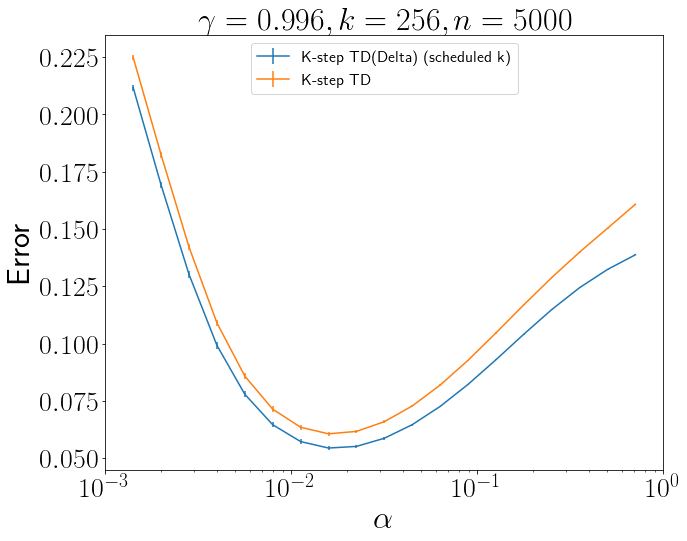}
    \caption{$\gamma=.996$ and different k values at different learning rates, where the number of timesteps $n=5000$.}
    \label{fig:g996}
\end{figure}

\begin{figure}[!htbp]
    \centering
    \includegraphics[width=.3\textwidth]{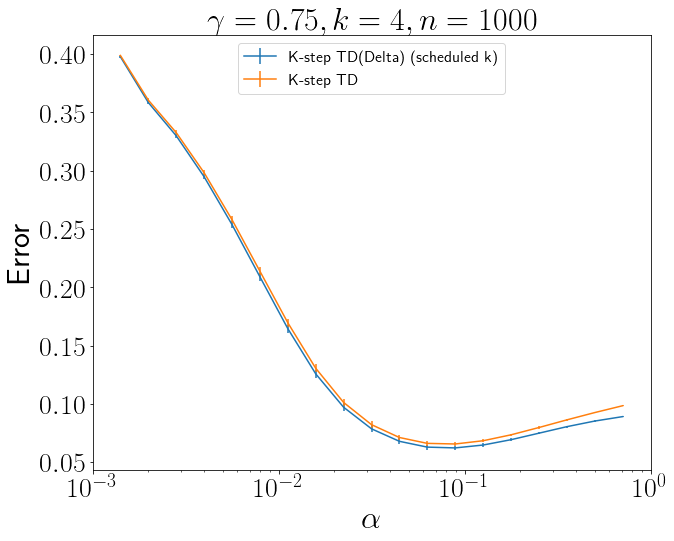}
    \includegraphics[width=.3\textwidth]{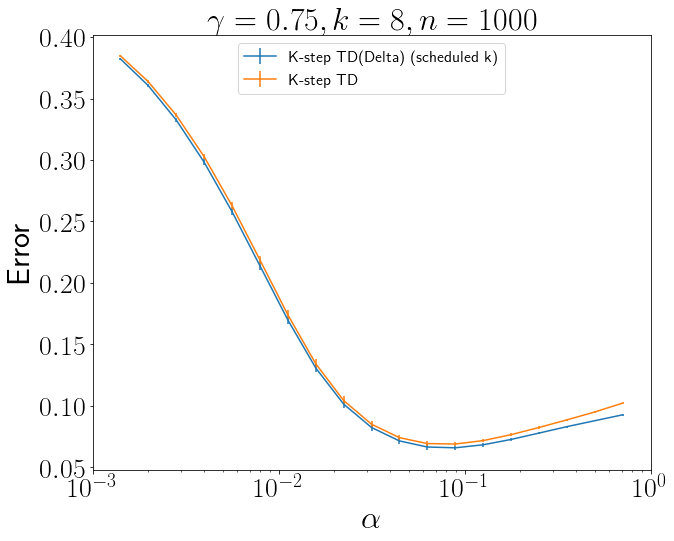}
    \includegraphics[width=.3\textwidth]{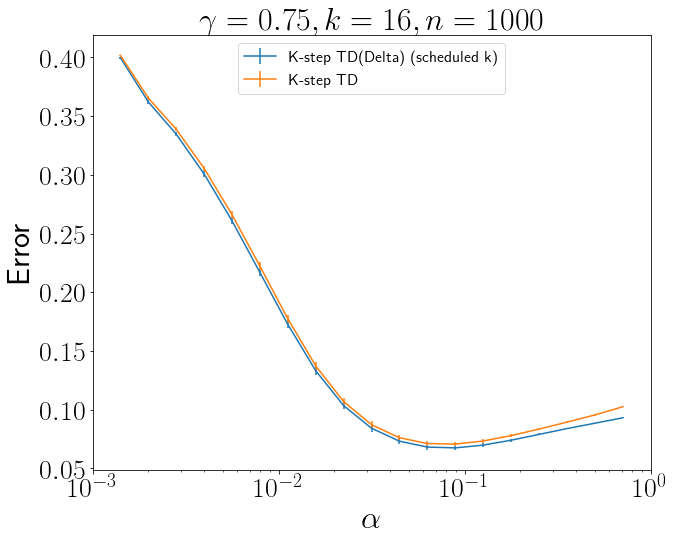}
    \includegraphics[width=.3\textwidth]{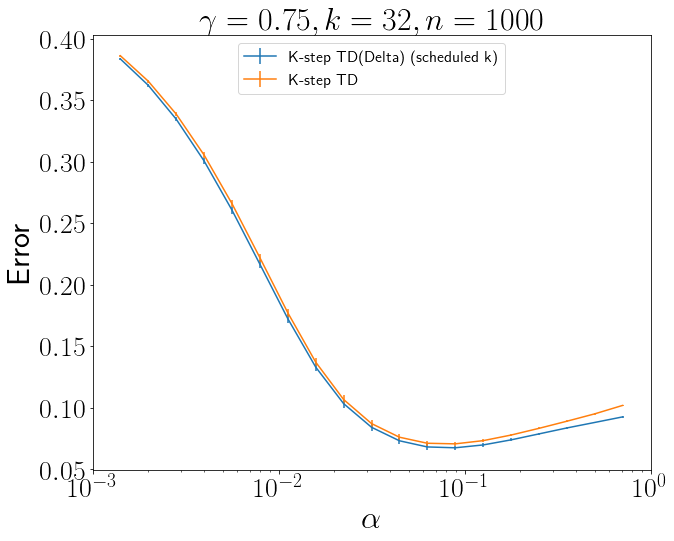}
    \includegraphics[width=.3\textwidth]{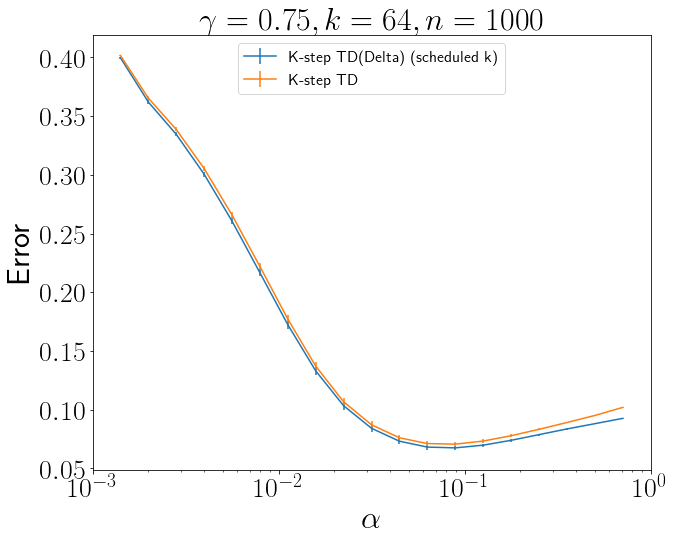}
    \includegraphics[width=.3\textwidth]{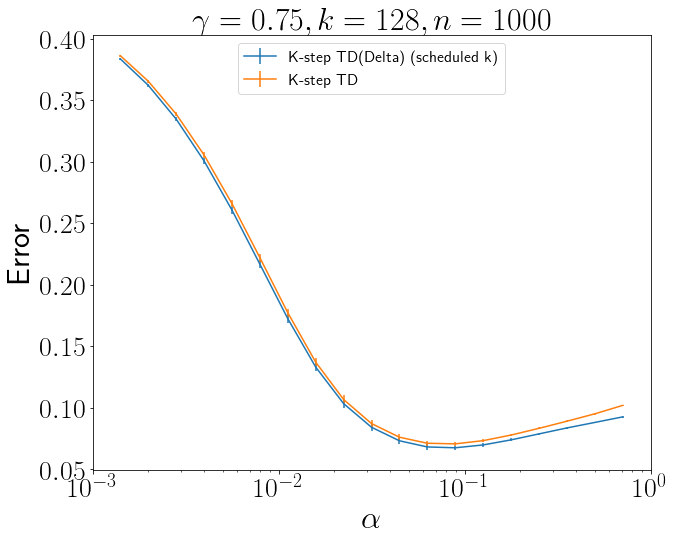}
    \includegraphics[width=.3\textwidth]{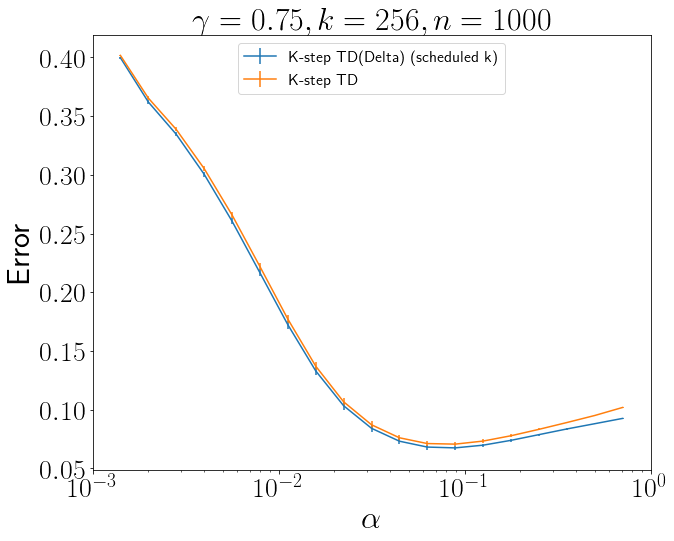}
    \caption{$\gamma=.75$ and different k values at different learning rates, where the number of timesteps $n=1000$.}
    \label{fig:g_75}
\end{figure}

\begin{figure}[!htbp]
    \centering
    \includegraphics[width=.3\textwidth]{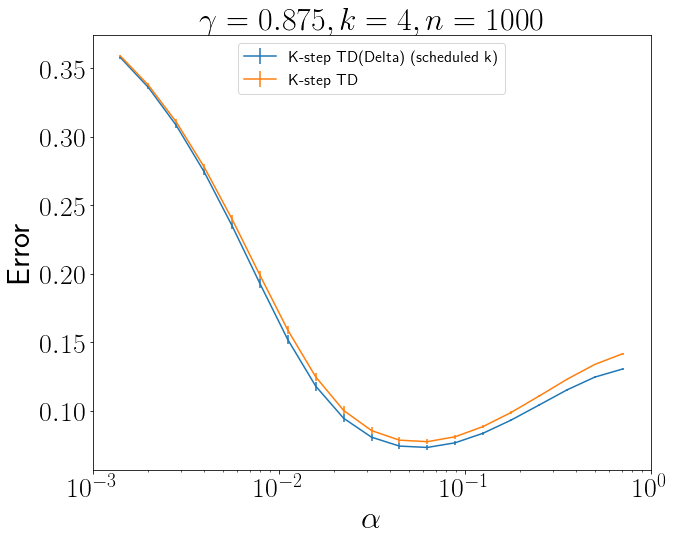}
    \includegraphics[width=.3\textwidth]{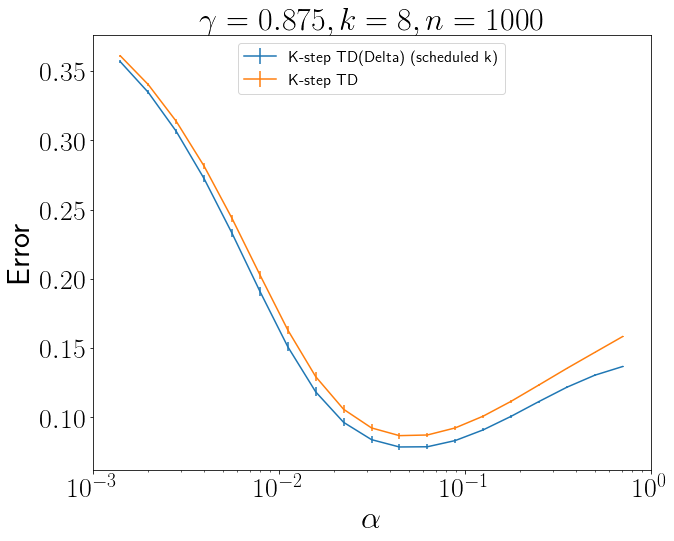}
    \includegraphics[width=.3\textwidth]{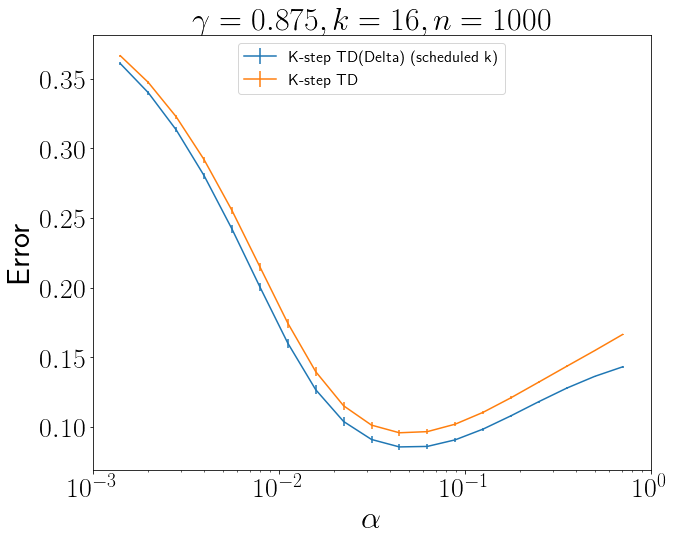}
    \includegraphics[width=.3\textwidth]{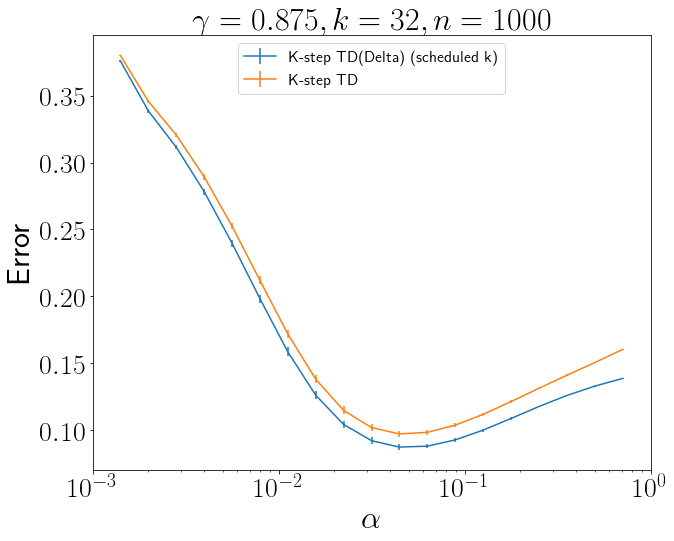}
    \includegraphics[width=.3\textwidth]{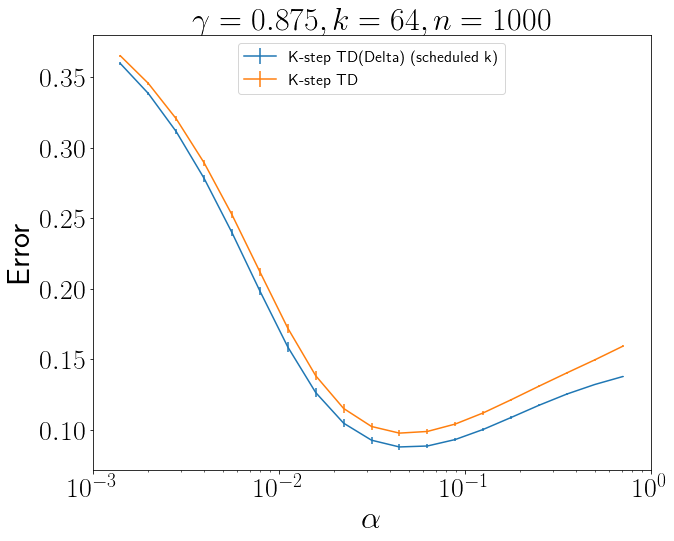}
    \includegraphics[width=.3\textwidth]{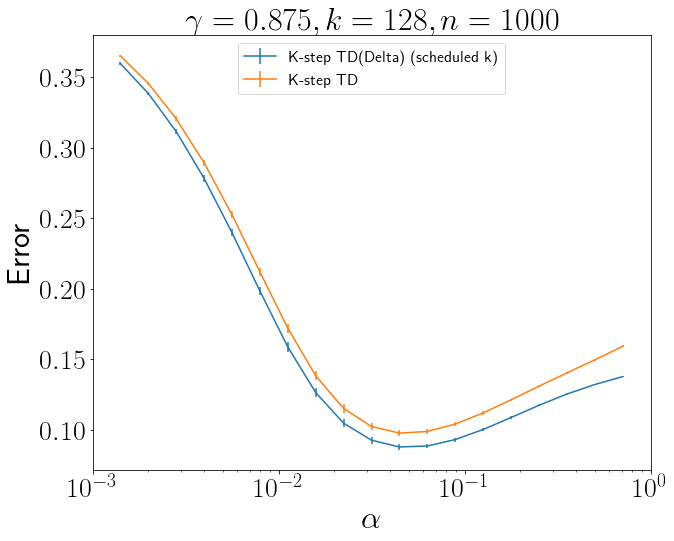}
    \includegraphics[width=.3\textwidth]{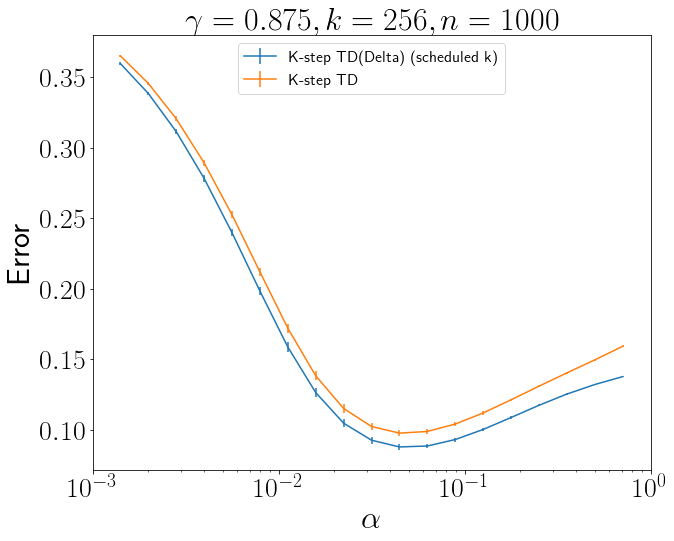}
    \caption{$\gamma=.875$ and different k values at different learning rates, where the number of timesteps $n=1000$.}
    \label{fig:g875}
\end{figure}

\begin{figure}[!htbp]
    \centering
    \includegraphics[width=.3\textwidth]{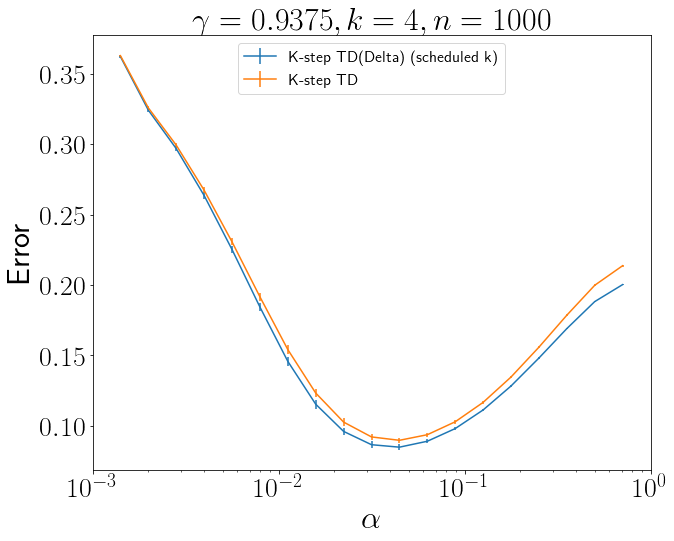}
    \includegraphics[width=.3\textwidth]{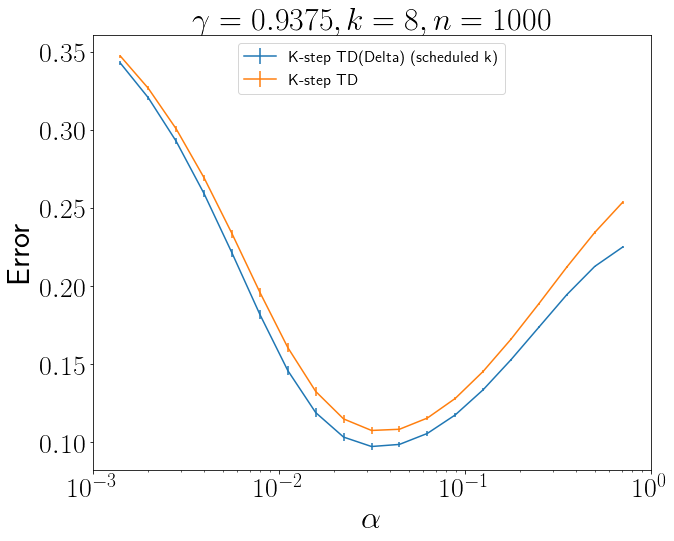}
    \includegraphics[width=.3\textwidth]{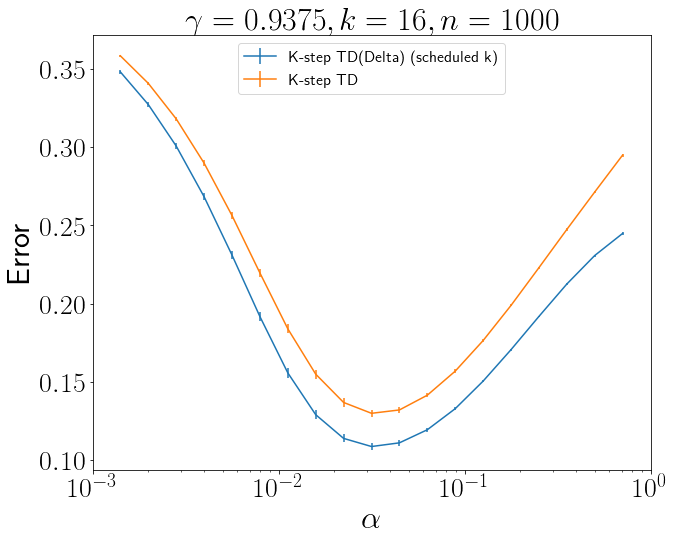}
    \includegraphics[width=.3\textwidth]{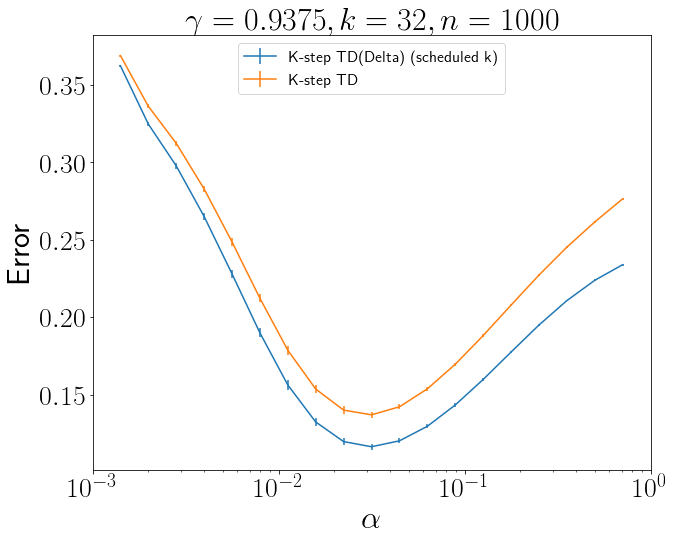}
    \includegraphics[width=.3\textwidth]{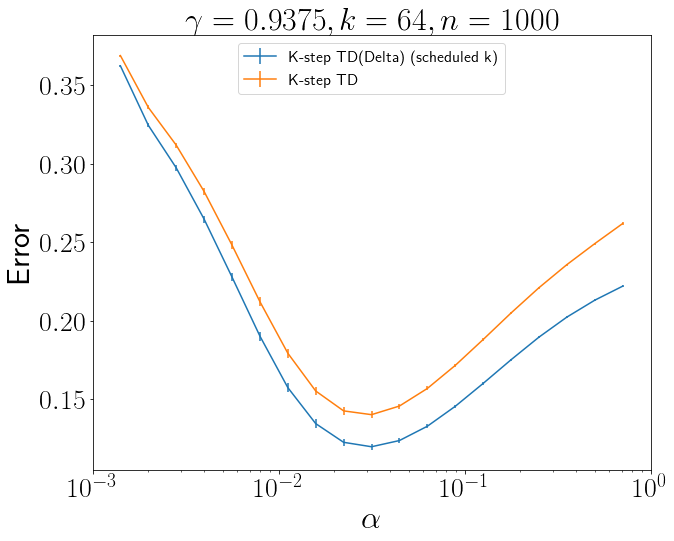}
    \includegraphics[width=.3\textwidth]{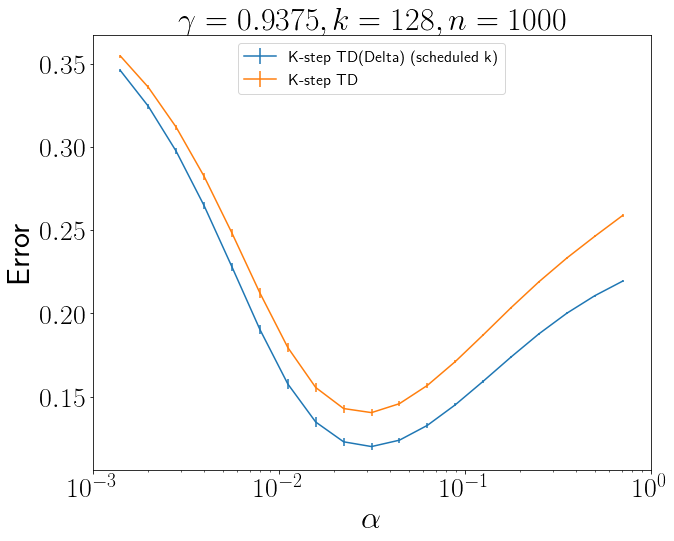}
    \includegraphics[width=.3\textwidth]{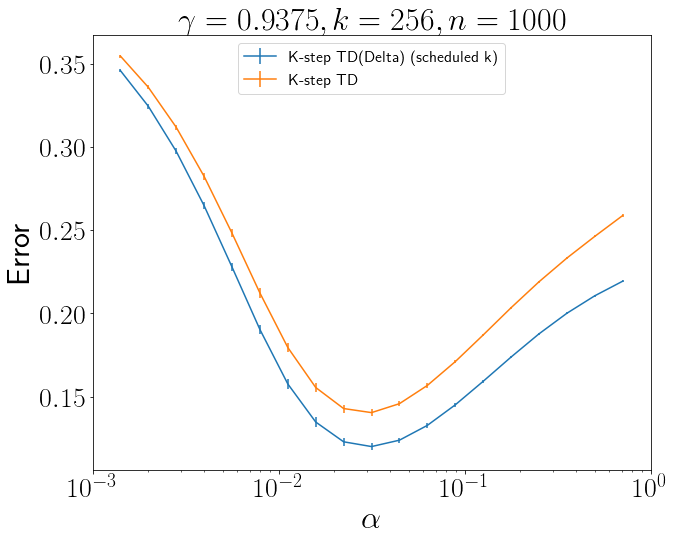}
    \caption{$\gamma=.93750$ and different k values at different learning rates, where the number of timesteps $n=1000$.}
    \label{fig:g9375}
\end{figure}

\begin{figure}[!htbp]
    \centering
        \includegraphics[width=.3\textwidth]{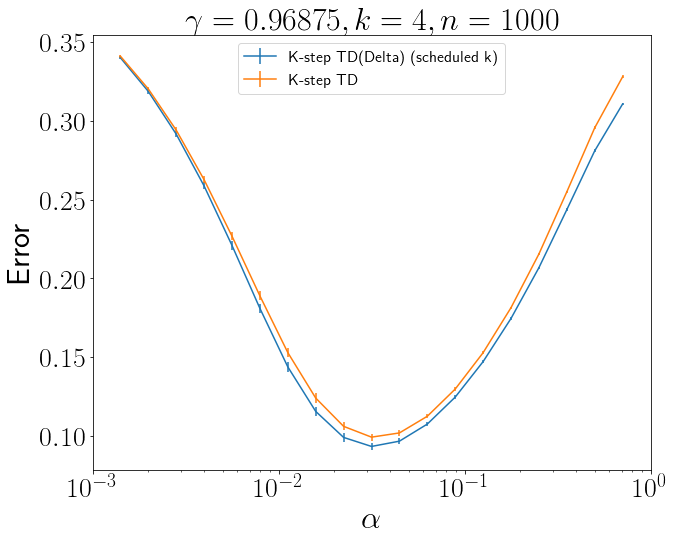}
    \includegraphics[width=.3\textwidth]{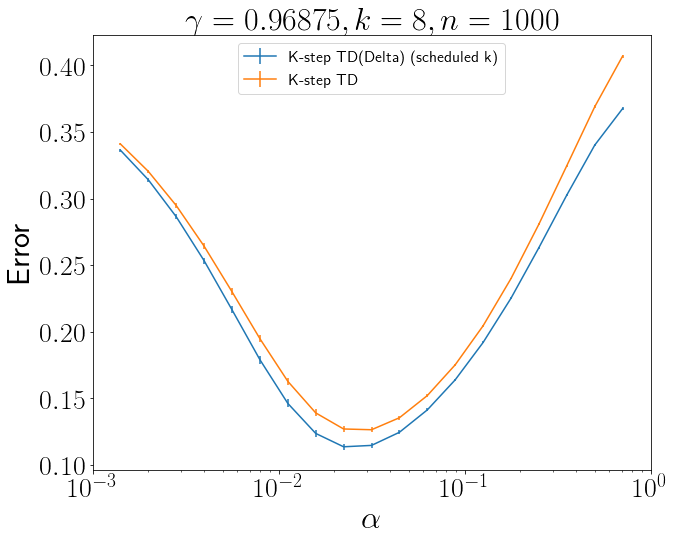}
    \includegraphics[width=.3\textwidth]{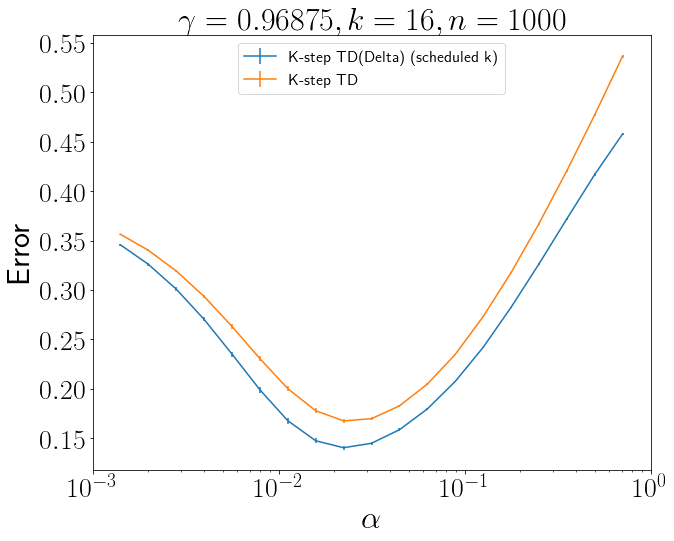}
    \includegraphics[width=.3\textwidth]{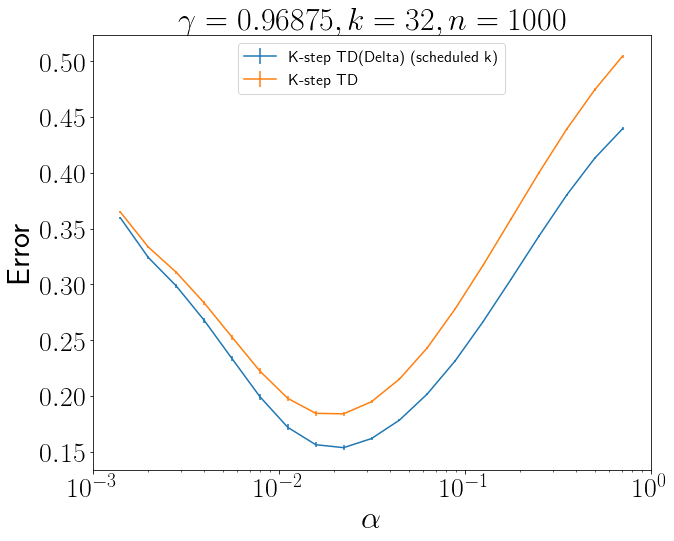}
    \includegraphics[width=.3\textwidth]{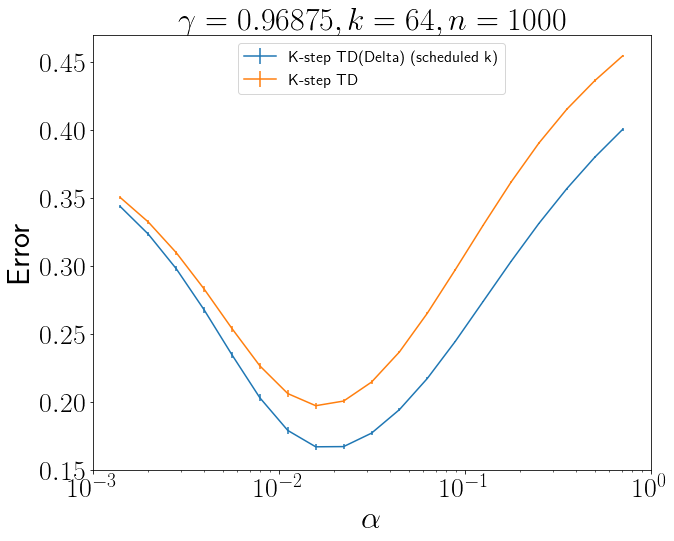}
    \includegraphics[width=.3\textwidth]{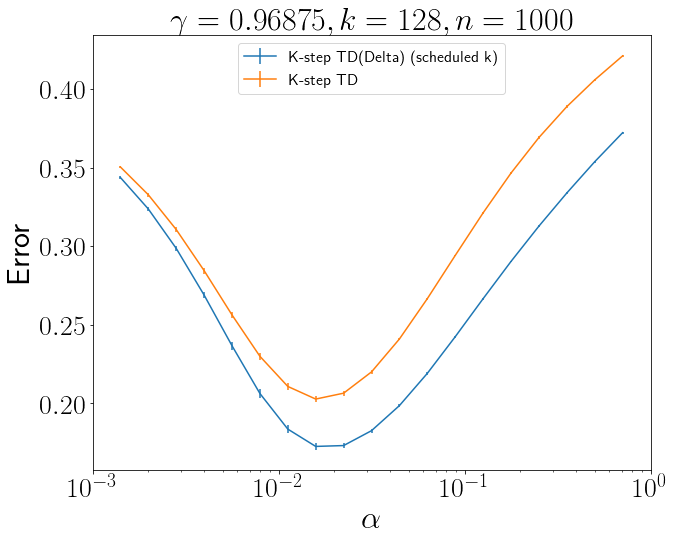}
    \includegraphics[width=.3\textwidth]{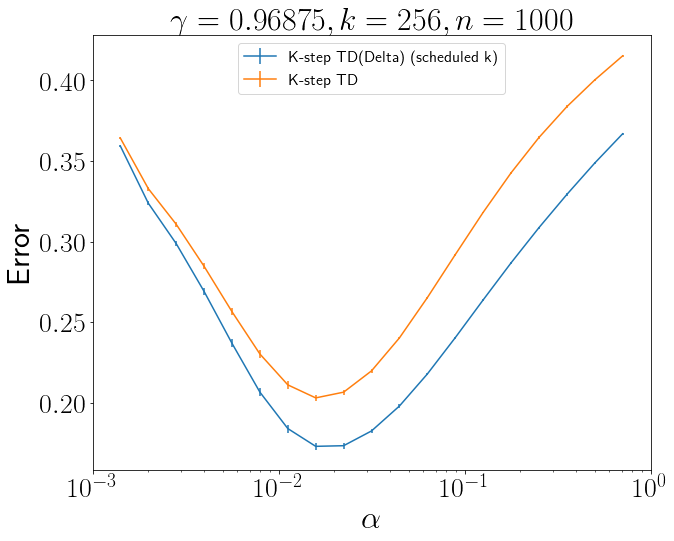}
    \caption{$\gamma=.96875$ and different k values at different learning rates, where the number of timesteps $n=1000$.}
    \label{fig:g96875}
\end{figure}

\begin{figure}[!htbp]
    \centering
        \includegraphics[width=.3\textwidth]{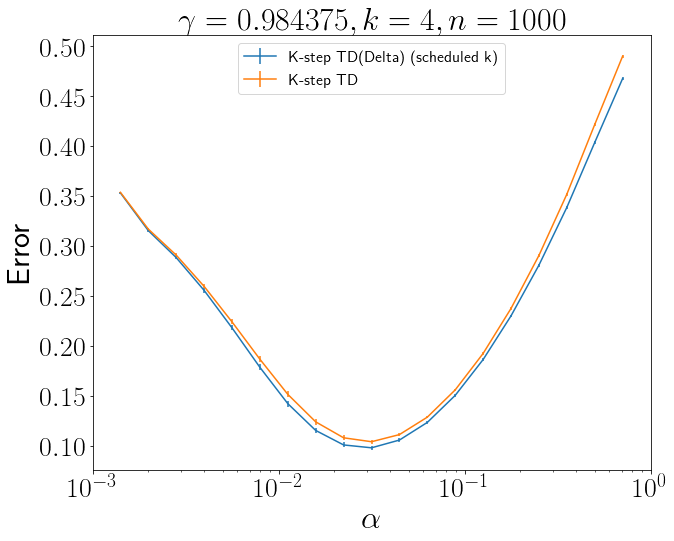}
    \includegraphics[width=.3\textwidth]{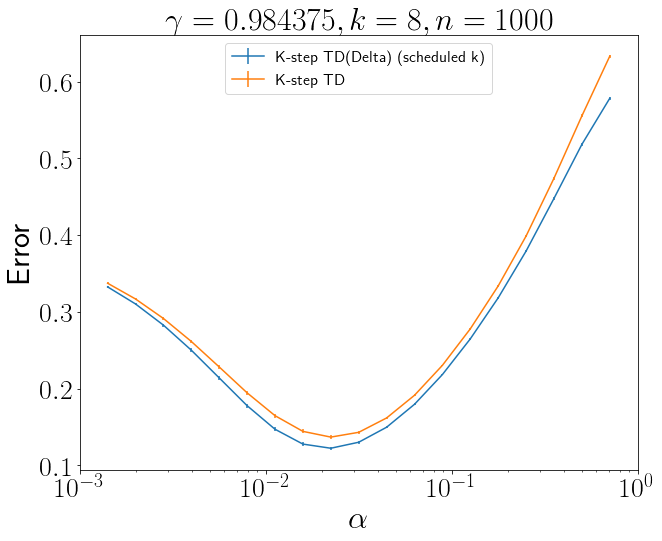}
    \includegraphics[width=.3\textwidth]{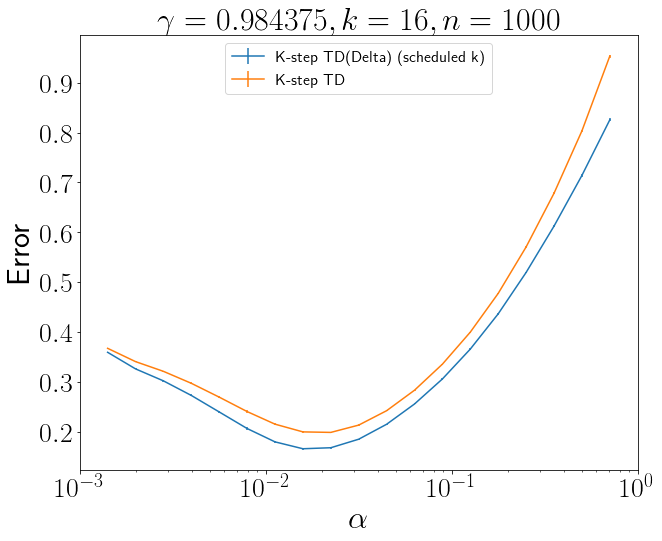}
    \includegraphics[width=.3\textwidth]{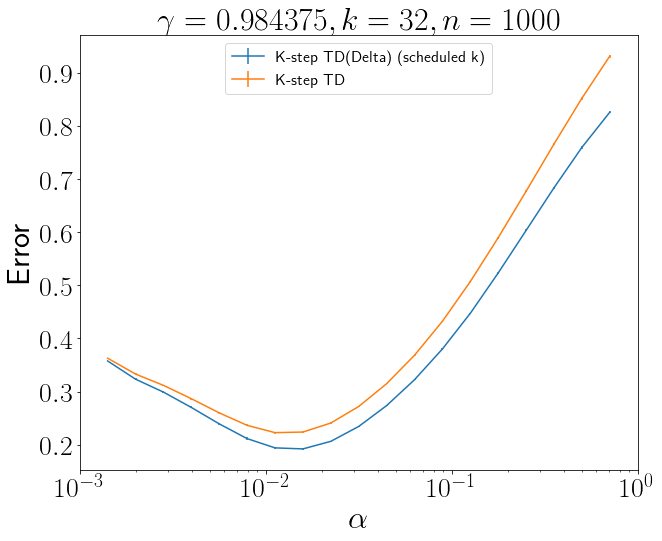}
    \includegraphics[width=.3\textwidth]{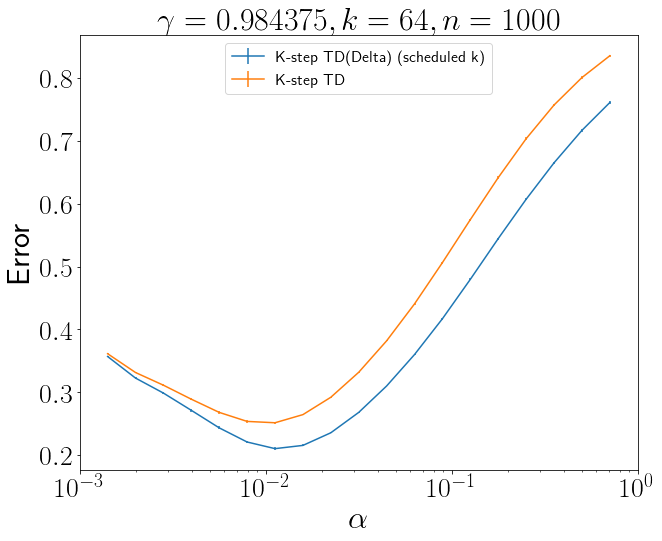}
    \includegraphics[width=.3\textwidth]{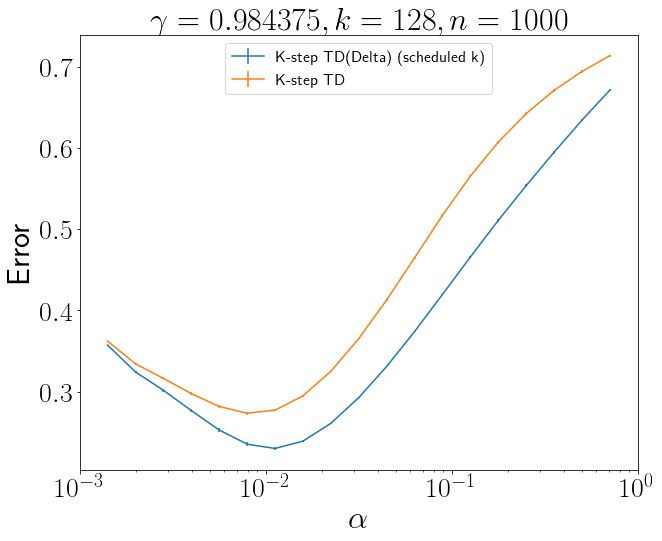}
    \includegraphics[width=.3\textwidth]{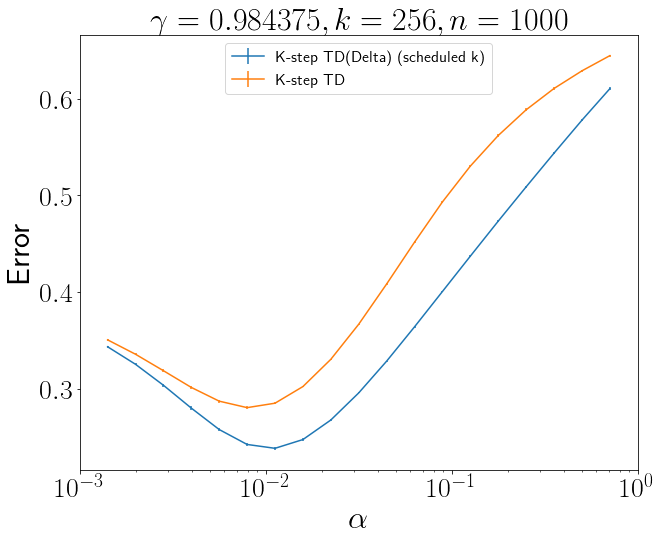}
    \caption{$\gamma=.98475$ and different k values at different learning rates, where the number of timesteps $n=1000$.}
    \label{fig:g984375}
\end{figure}

\begin{figure}[!htbp]
    \centering
        \includegraphics[width=.3\textwidth]{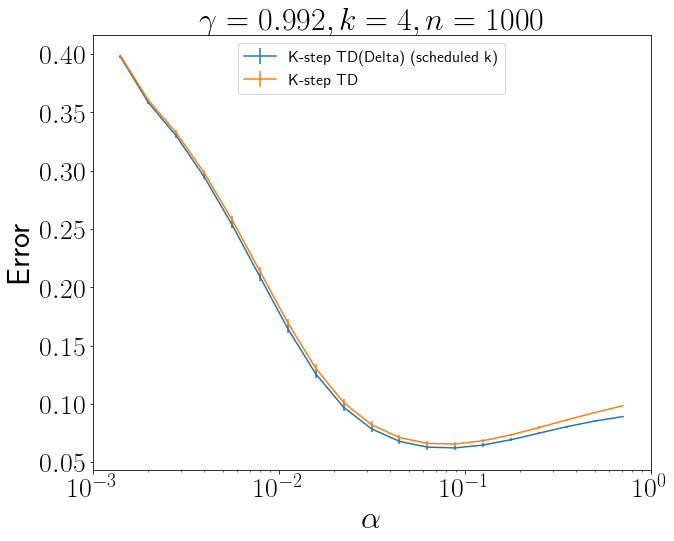}
    \includegraphics[width=.3\textwidth]{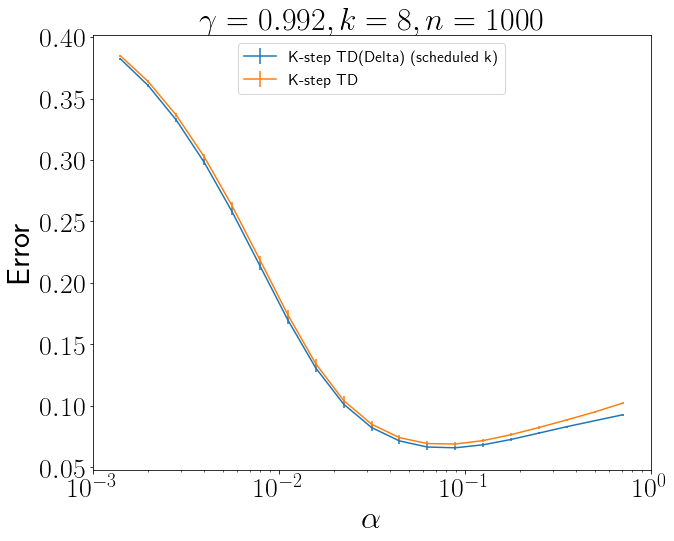}
    \includegraphics[width=.3\textwidth]{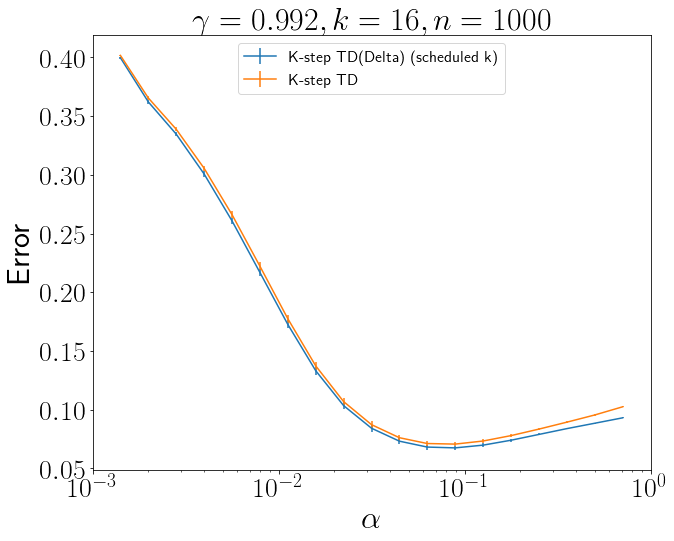}
    \includegraphics[width=.3\textwidth]{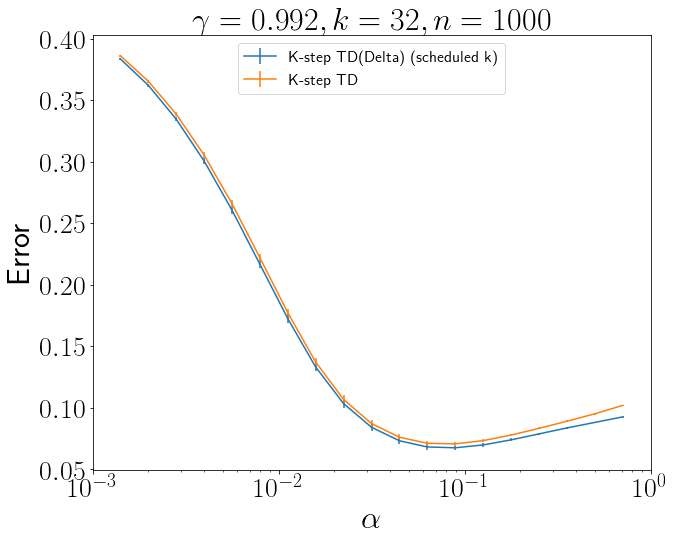}
    \includegraphics[width=.3\textwidth]{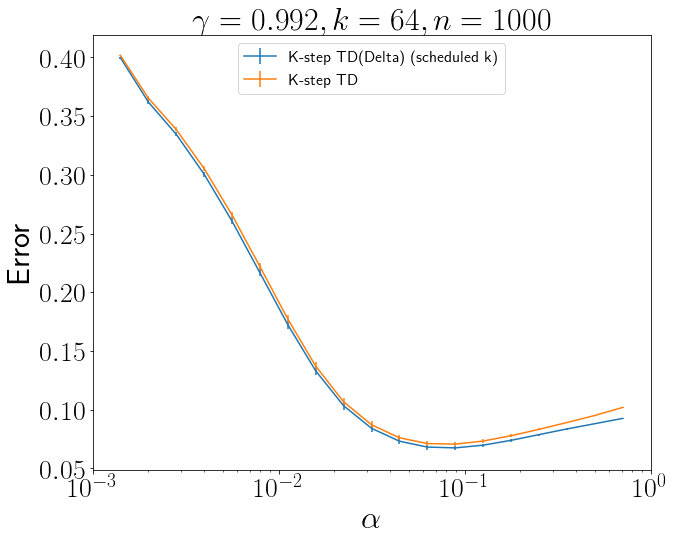}
    \includegraphics[width=.3\textwidth]{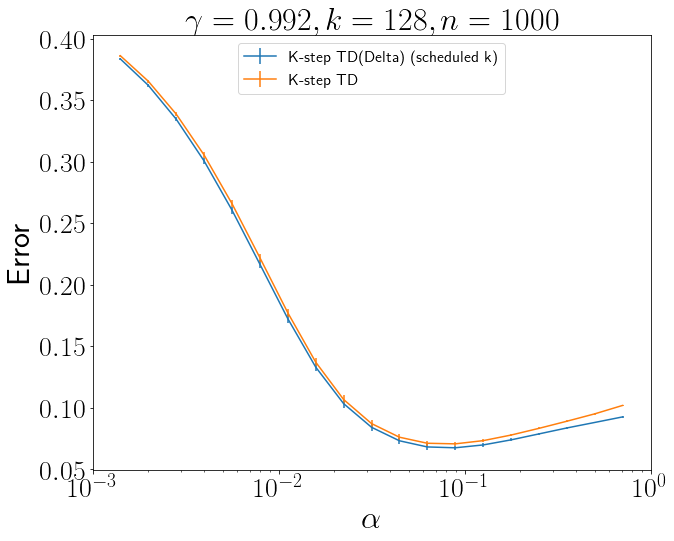}
    \includegraphics[width=.3\textwidth]{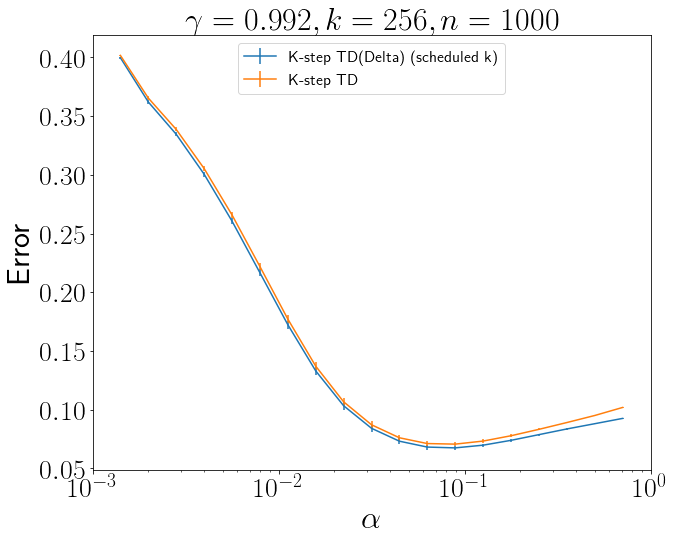}
    \caption{$\gamma=.992$ and different k values at different learning rates, where the number of timesteps $n=1000$.}
    \label{fig:g992}
\end{figure}

\begin{figure}[!htbp]
    \centering
        \includegraphics[width=.3\textwidth]{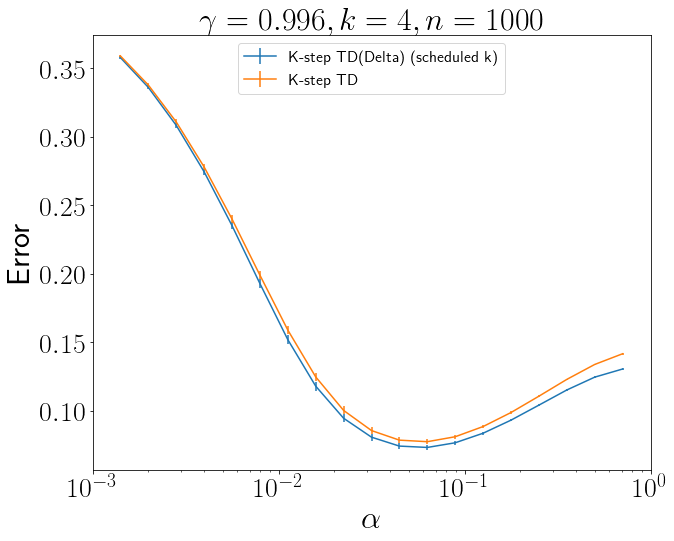}
    \includegraphics[width=.3\textwidth]{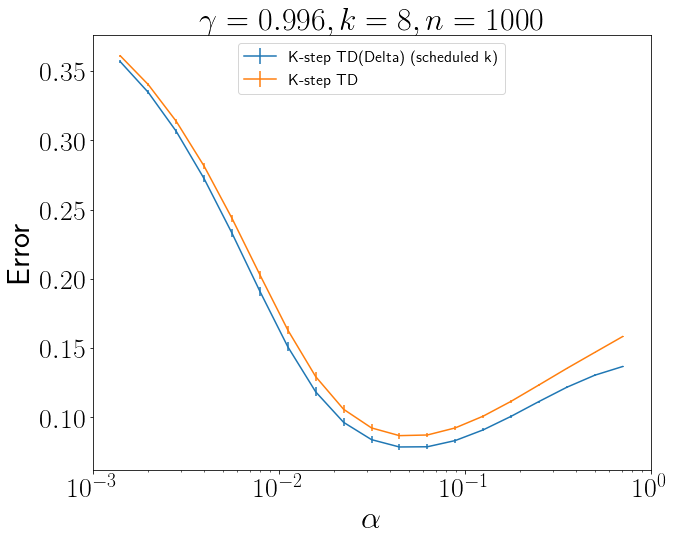}
    \includegraphics[width=.3\textwidth]{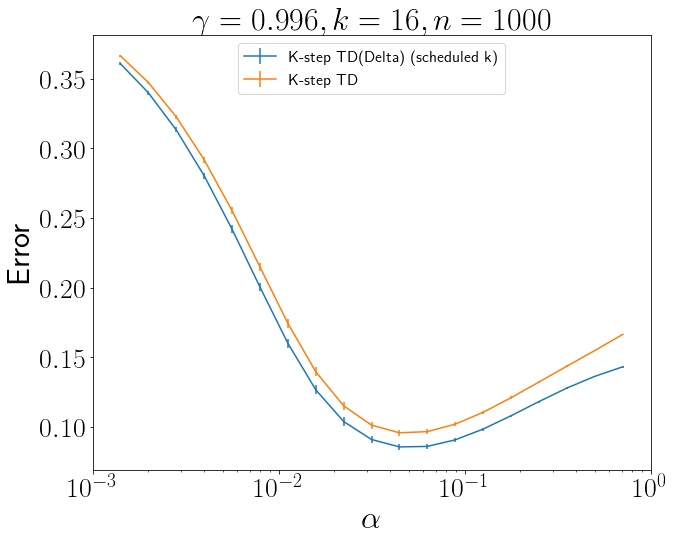}
    \includegraphics[width=.3\textwidth]{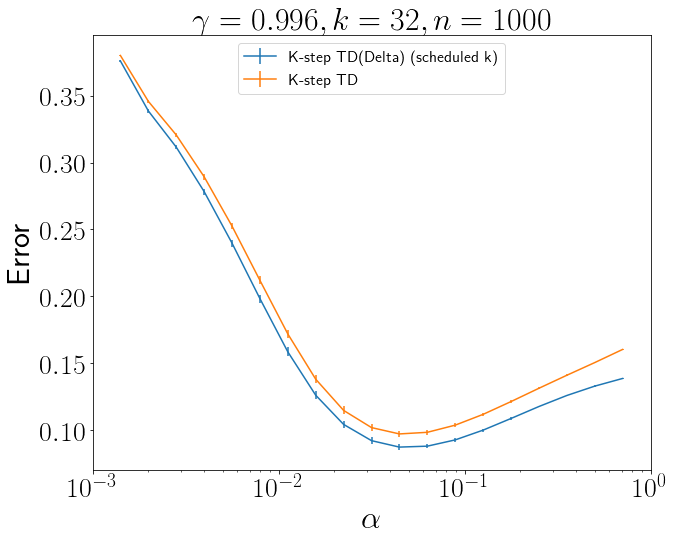}
    \includegraphics[width=.3\textwidth]{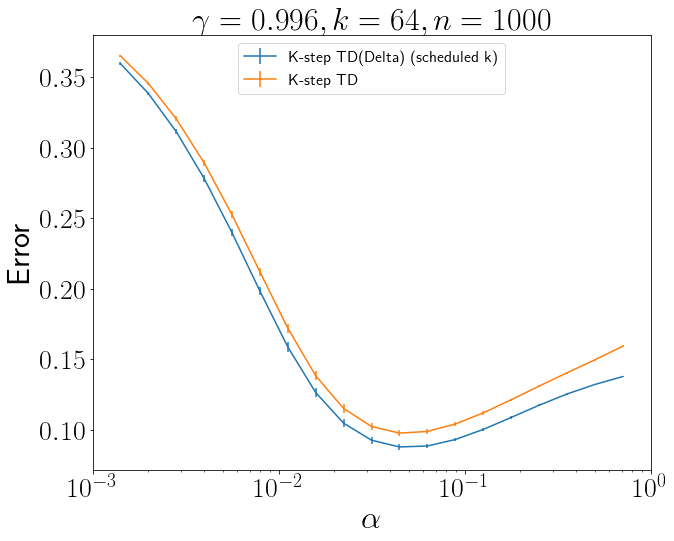}
    \includegraphics[width=.3\textwidth]{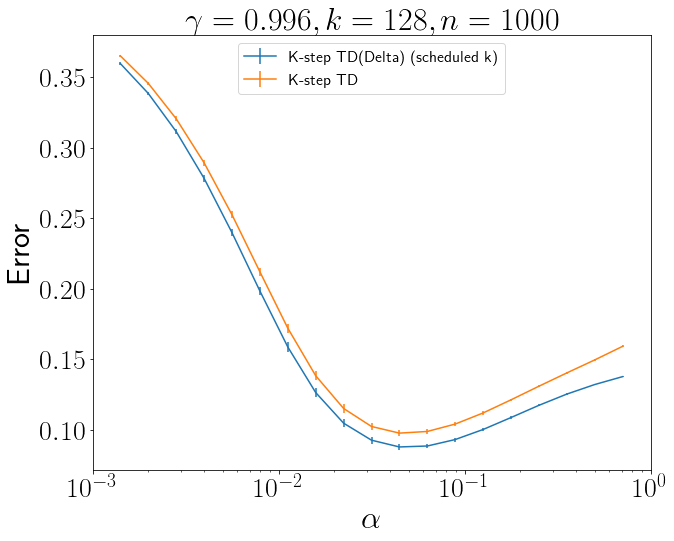}
    \includegraphics[width=.3\textwidth]{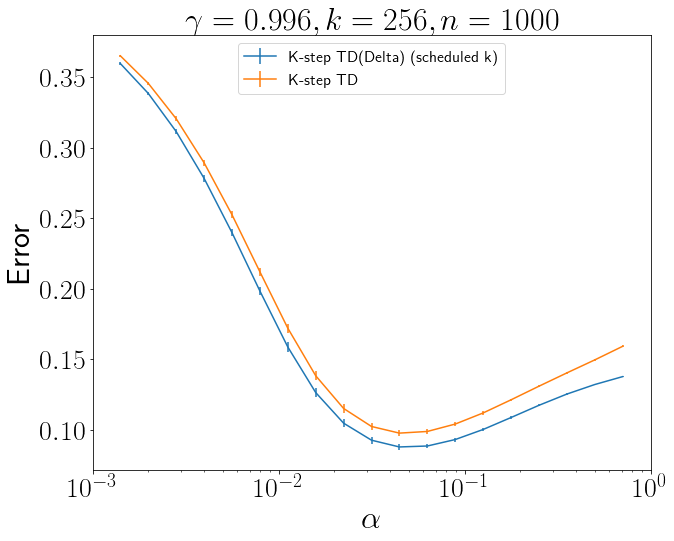}
    \caption{$\gamma=.996$ and different k values at different learning rates, where the number of timesteps $n=1000$.}
    \label{fig:g996}
\end{figure}

\begin{figure}[!htbp]
    \centering
    \includegraphics[width=.3\textwidth]{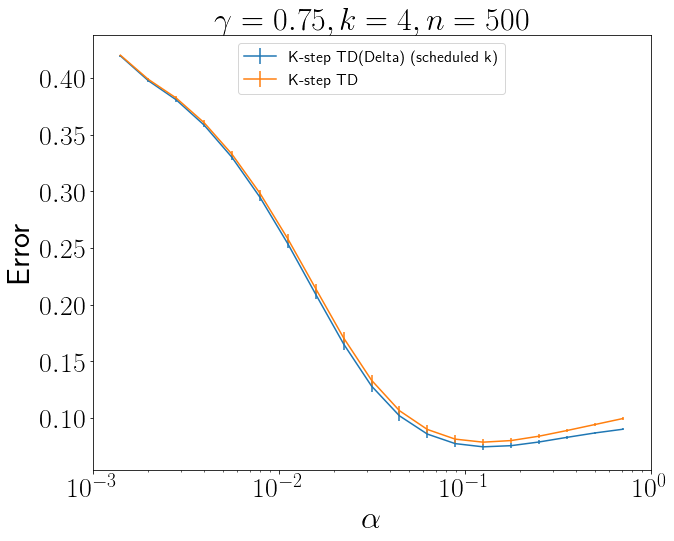}
    \includegraphics[width=.3\textwidth]{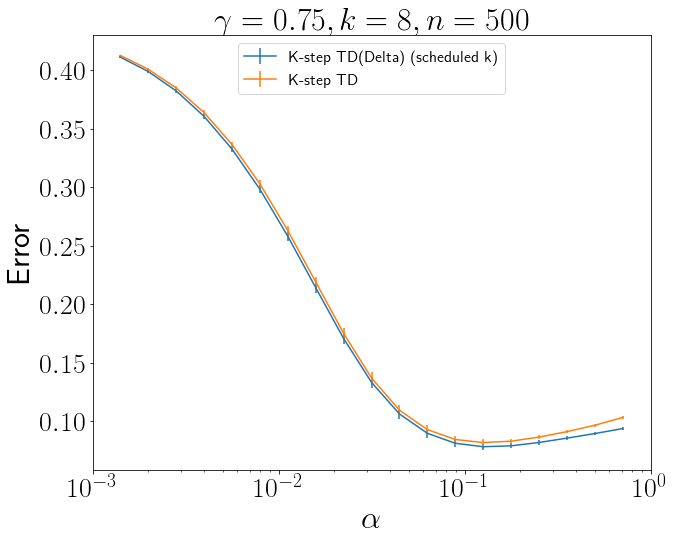}
    \includegraphics[width=.3\textwidth]{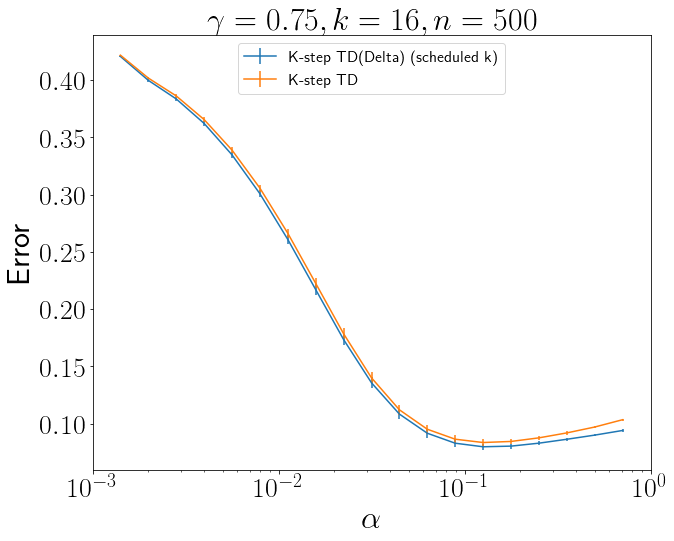}
    \includegraphics[width=.3\textwidth]{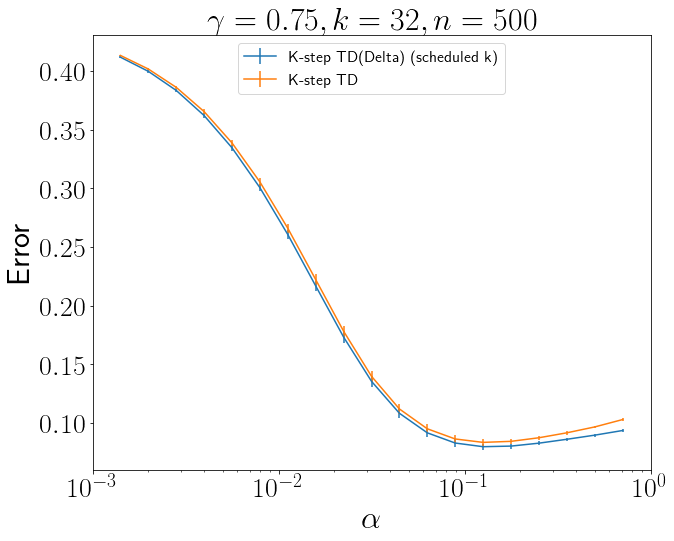}
    \includegraphics[width=.3\textwidth]{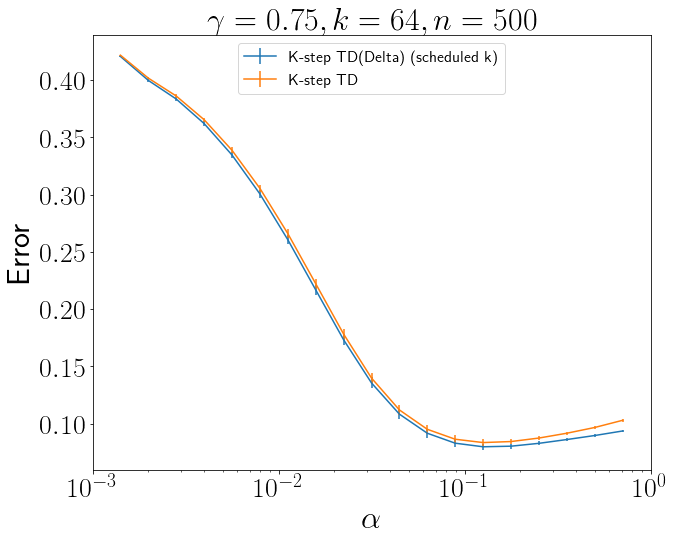}
    \includegraphics[width=.3\textwidth]{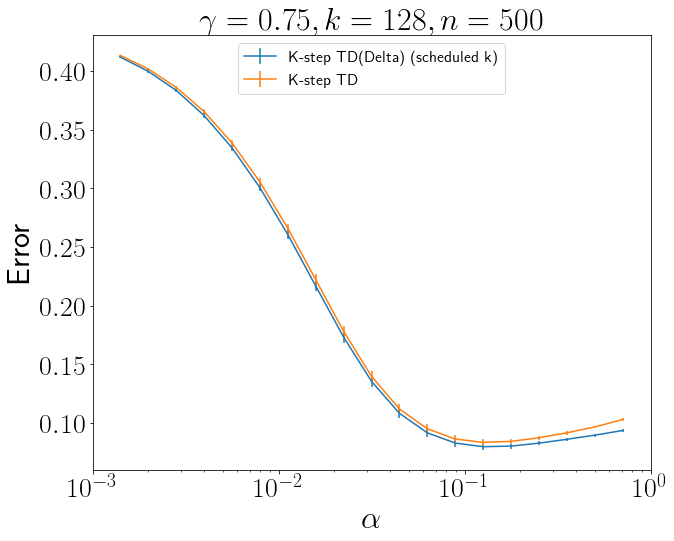}
    \includegraphics[width=.3\textwidth]{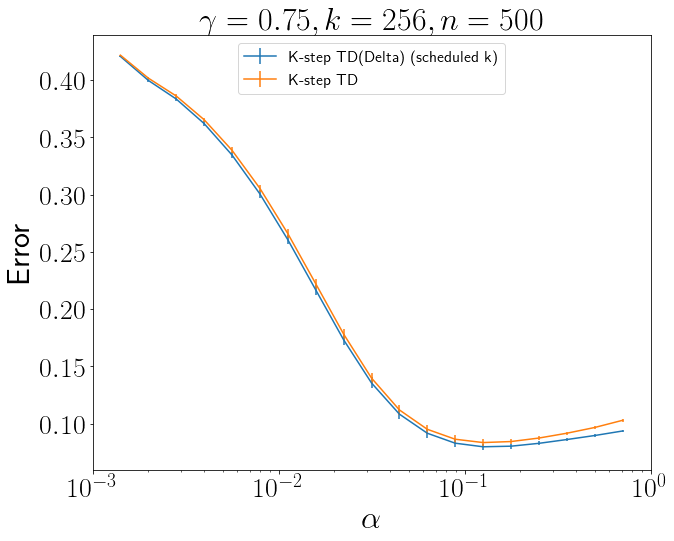}
    \caption{$\gamma=.75$ and different k values at different learning rates, where the number of timesteps $n=500$.}
    \label{fig:g_75}
\end{figure}

\begin{figure}[!htbp]
    \centering
    \includegraphics[width=.3\textwidth]{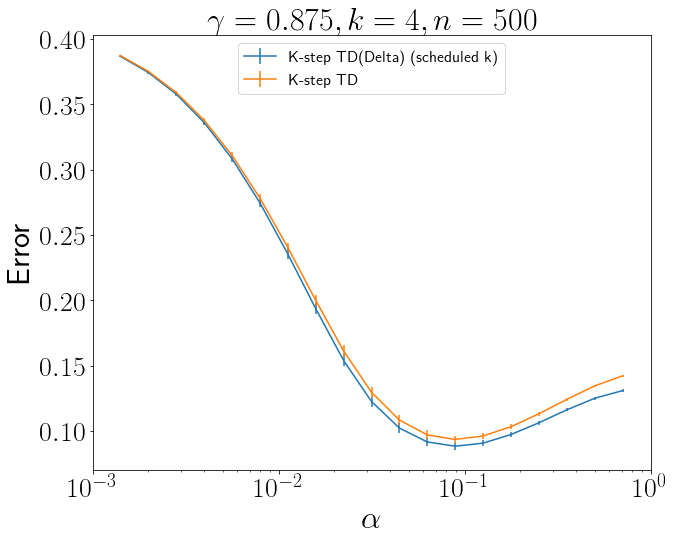}
    \includegraphics[width=.3\textwidth]{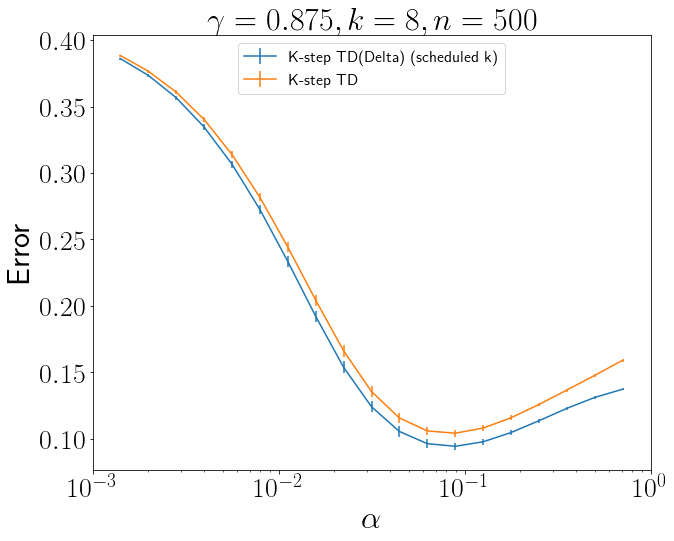}
    \includegraphics[width=.3\textwidth]{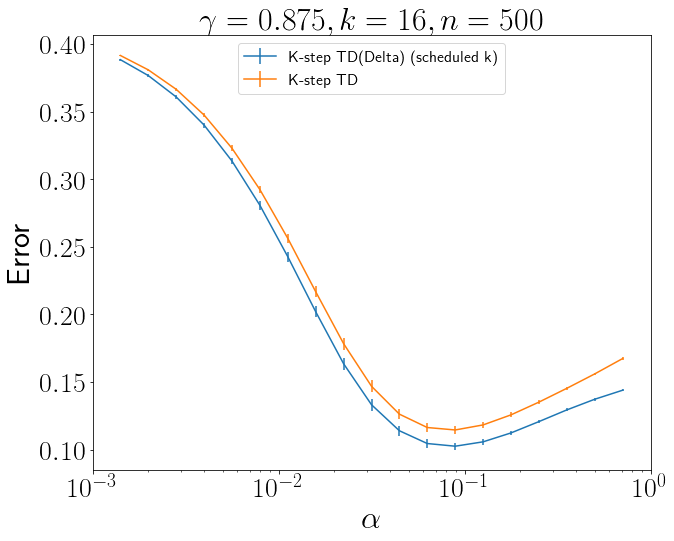}
    \includegraphics[width=.3\textwidth]{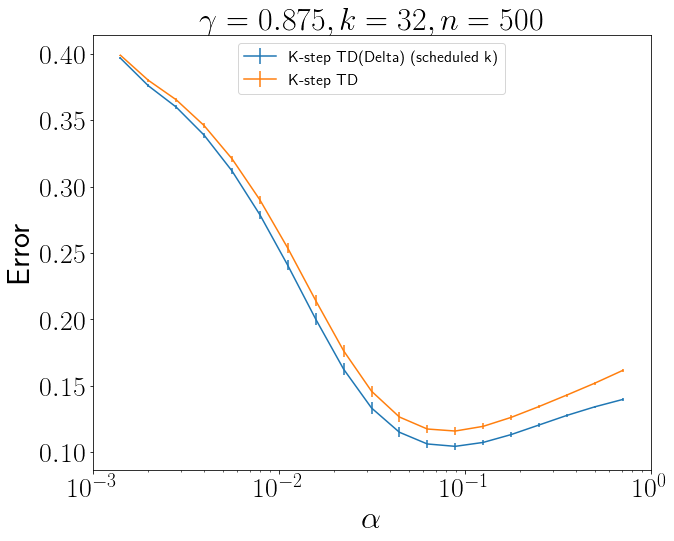}
    \includegraphics[width=.3\textwidth]{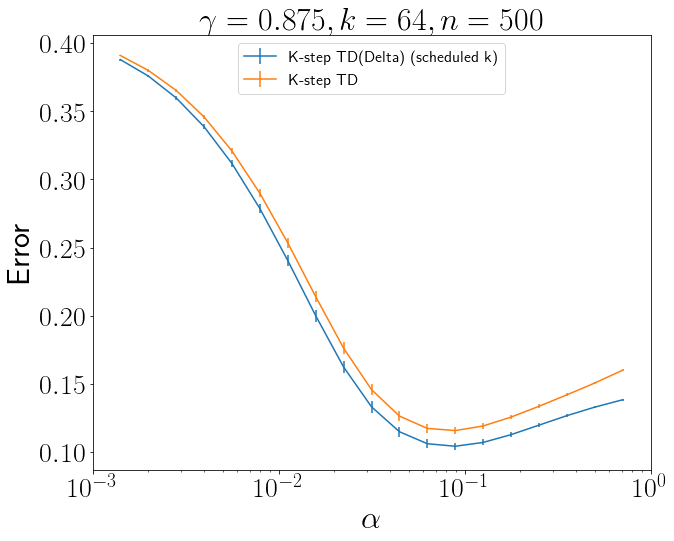}
    \includegraphics[width=.3\textwidth]{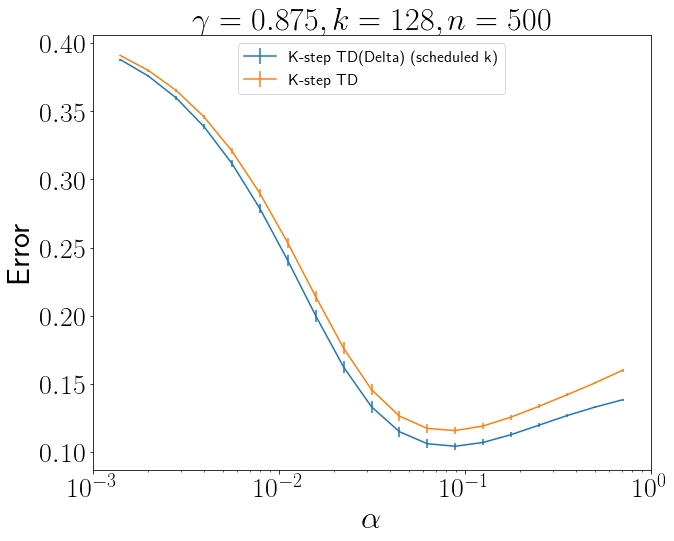}
    \includegraphics[width=.3\textwidth]{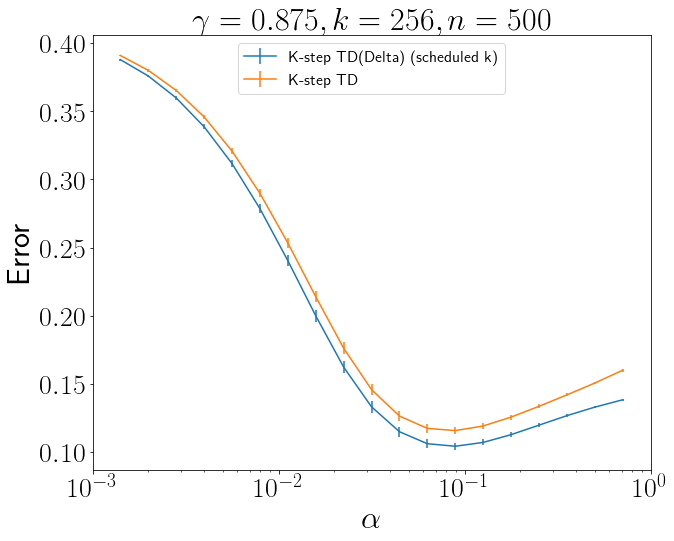}
    \caption{$\gamma=.875$ and different k values at different learning rates, where the number of timesteps $n=500$.}
    \label{fig:g875}
\end{figure}

\begin{figure}[!htbp]
    \centering
    \includegraphics[width=.3\textwidth]{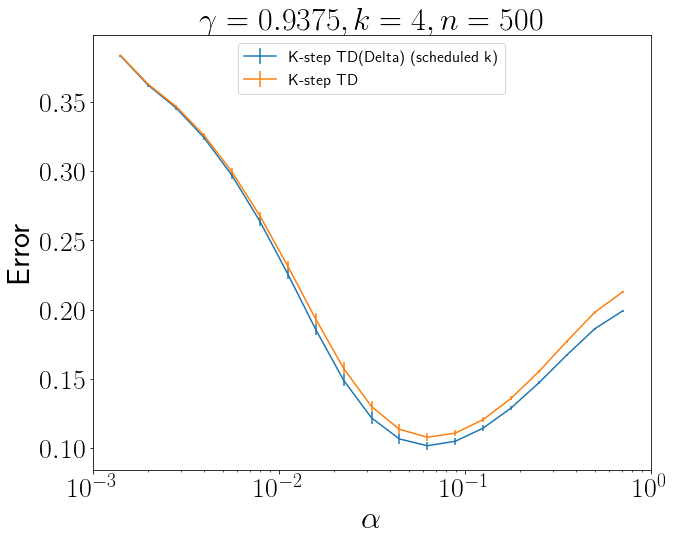}
    \includegraphics[width=.3\textwidth]{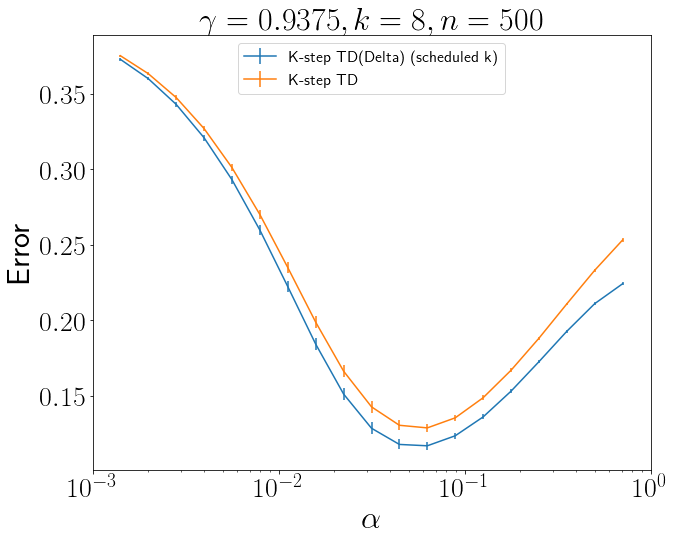}
    \includegraphics[width=.3\textwidth]{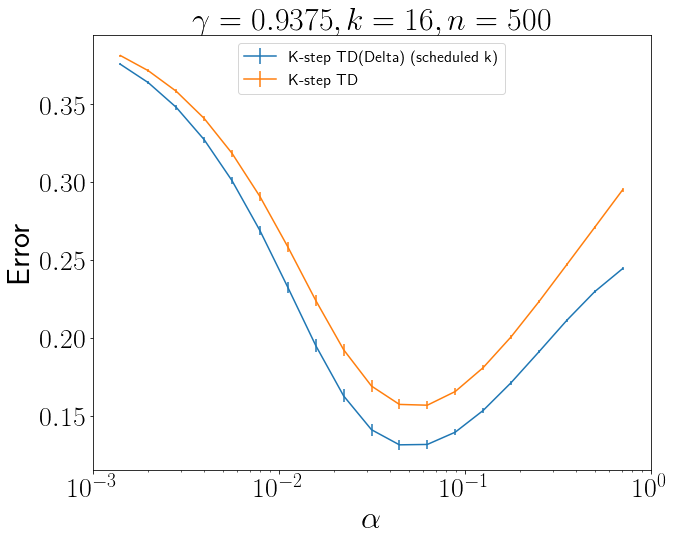}
    \includegraphics[width=.3\textwidth]{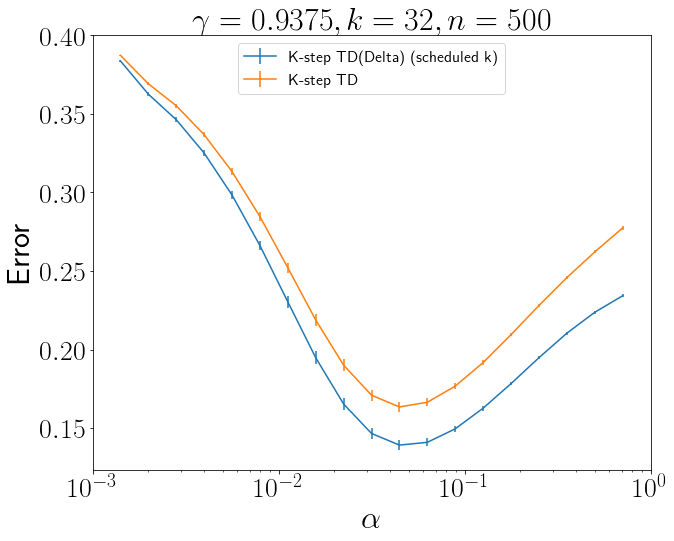}
    \includegraphics[width=.3\textwidth]{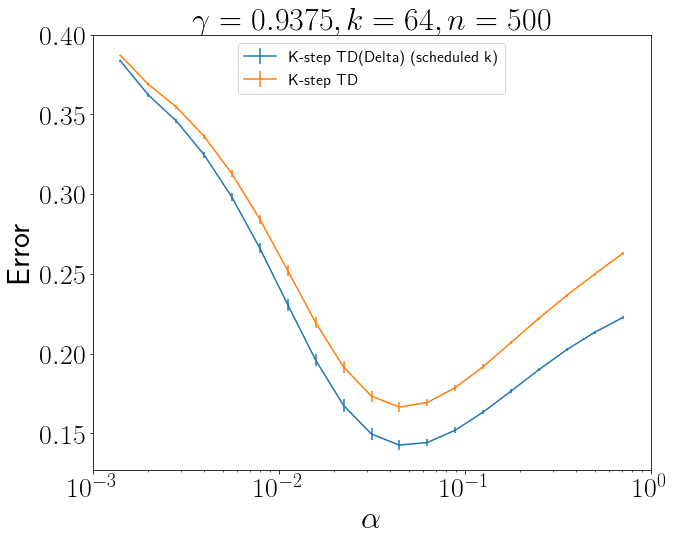}
    \includegraphics[width=.3\textwidth]{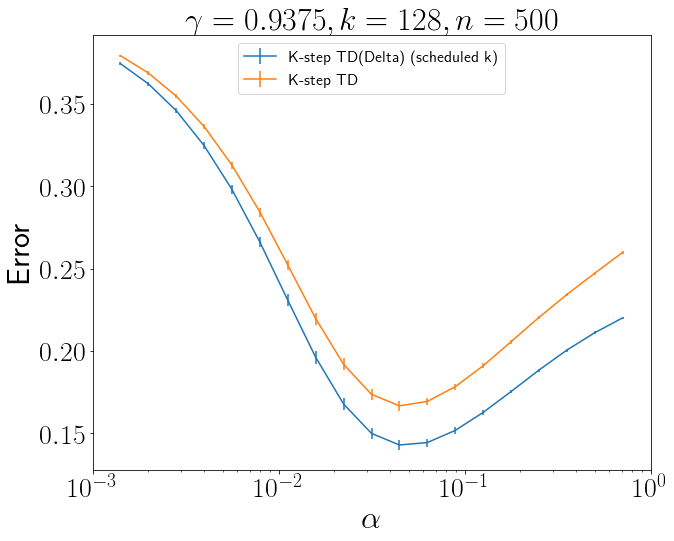}
    \includegraphics[width=.3\textwidth]{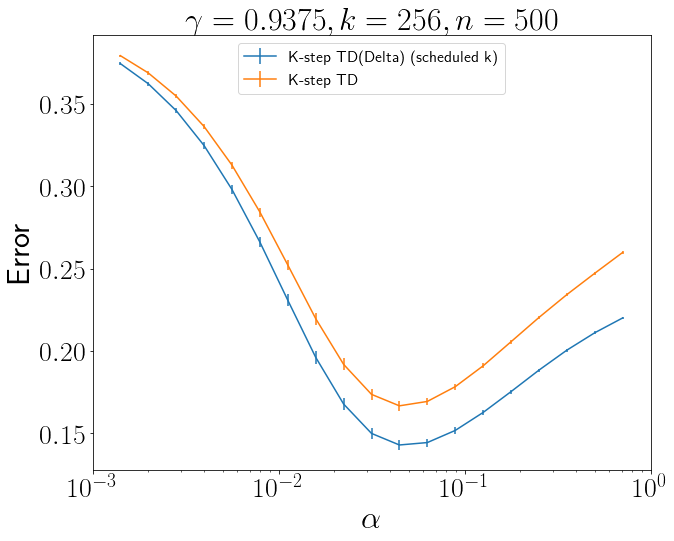}
    \caption{$\gamma=.93750$ and different k values at different learning rates, where the number of timesteps $n=500$.}
    \label{fig:g9375}
\end{figure}

\begin{figure}[!htbp]
    \centering
        \includegraphics[width=.3\textwidth]{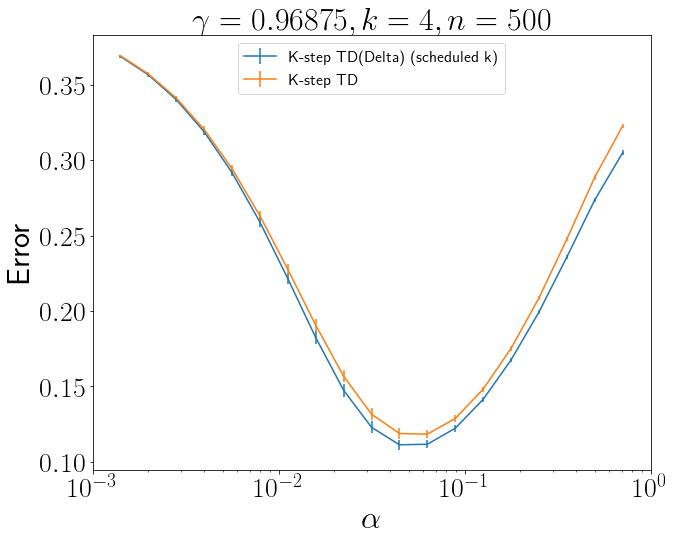}
    \includegraphics[width=.3\textwidth]{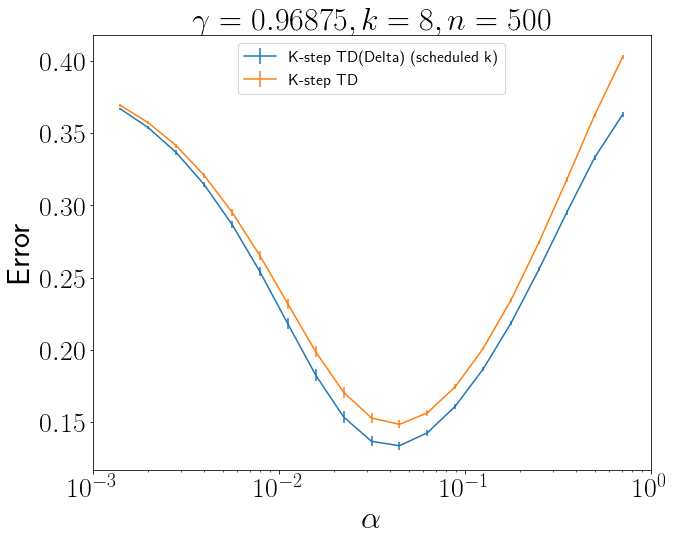}
    \includegraphics[width=.3\textwidth]{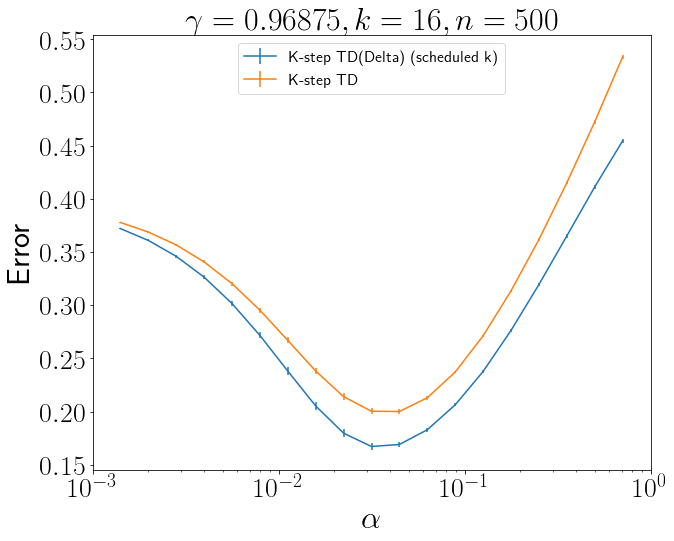}
    \includegraphics[width=.3\textwidth]{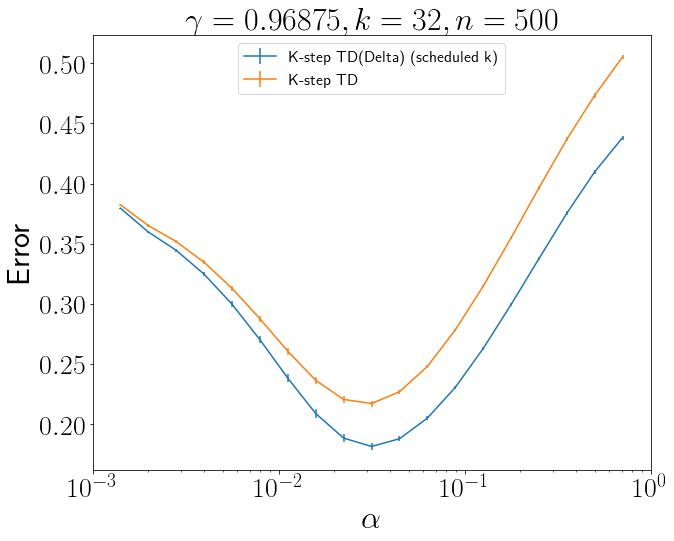}
    \includegraphics[width=.3\textwidth]{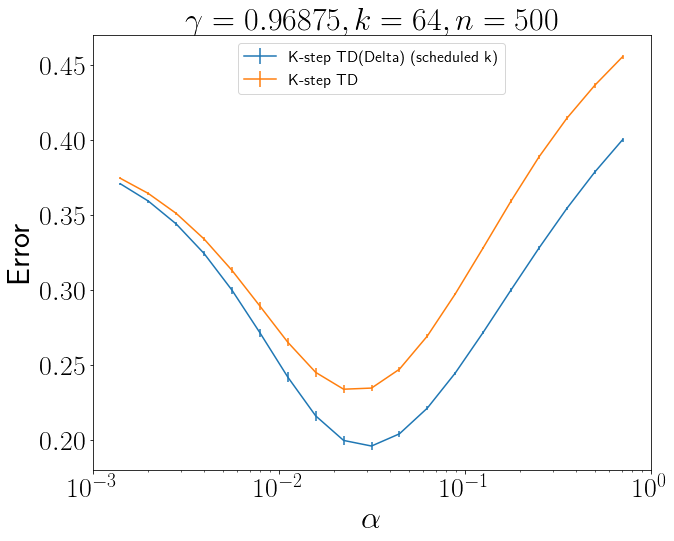}
    \includegraphics[width=.3\textwidth]{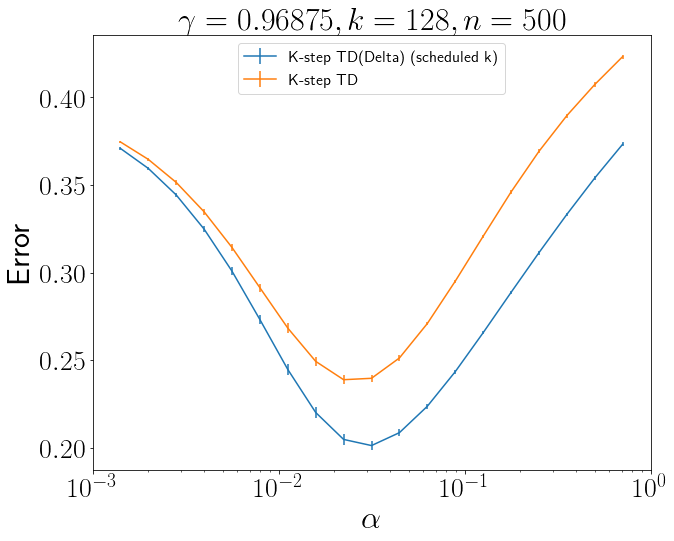}
    \includegraphics[width=.3\textwidth]{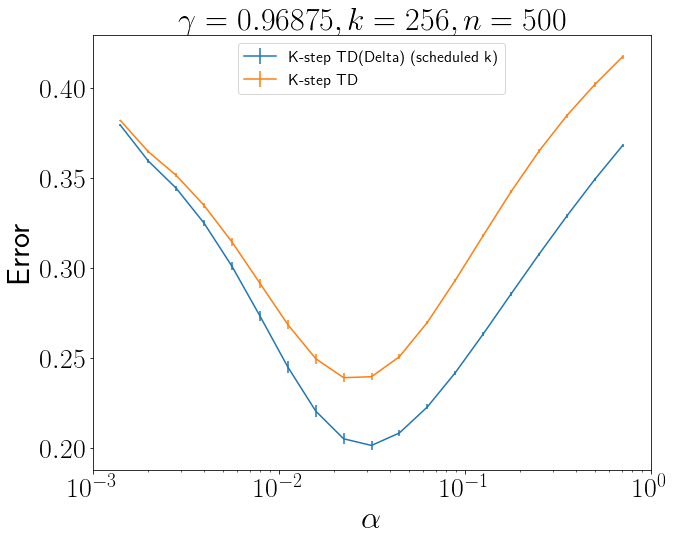}
    \caption{$\gamma=.96875$ and different k values at different learning rates, where the number of timesteps $n=500$.}
    \label{fig:g96875}
\end{figure}

\begin{figure}[!htbp]
    \centering
        \includegraphics[width=.3\textwidth]{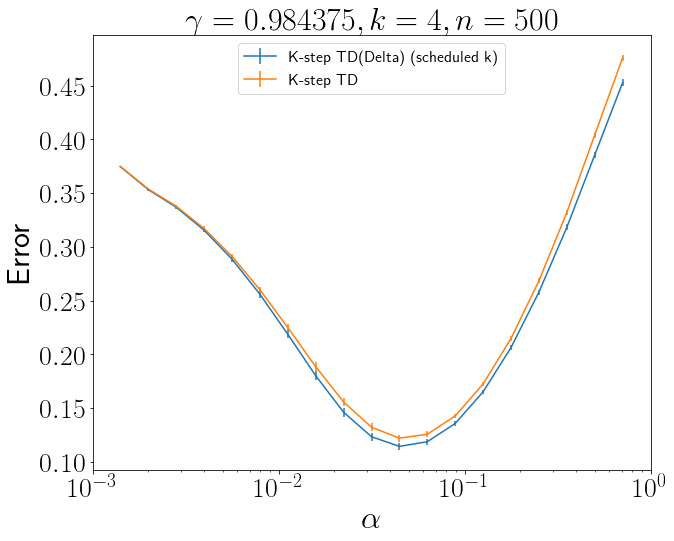}
    \includegraphics[width=.3\textwidth]{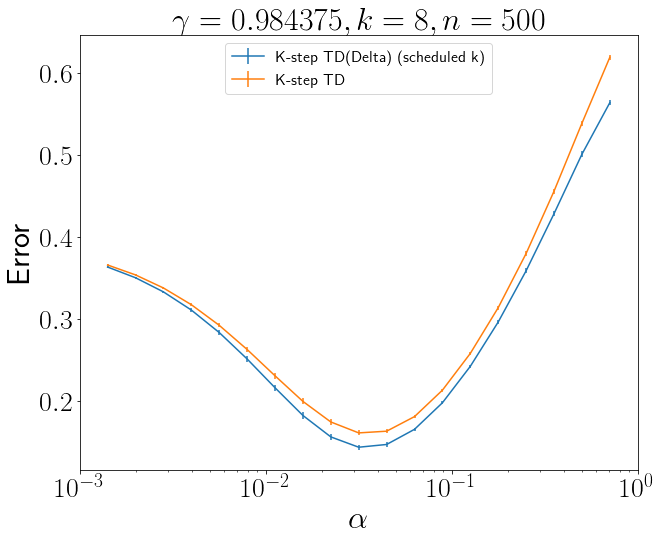}
    \includegraphics[width=.3\textwidth]{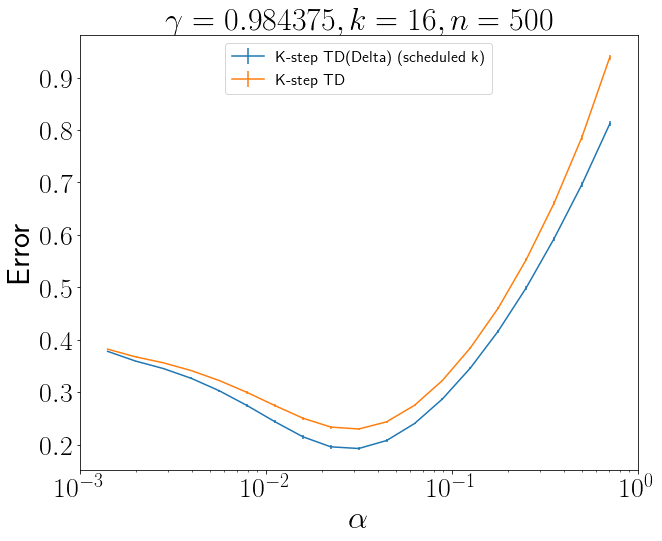}
    \includegraphics[width=.3\textwidth]{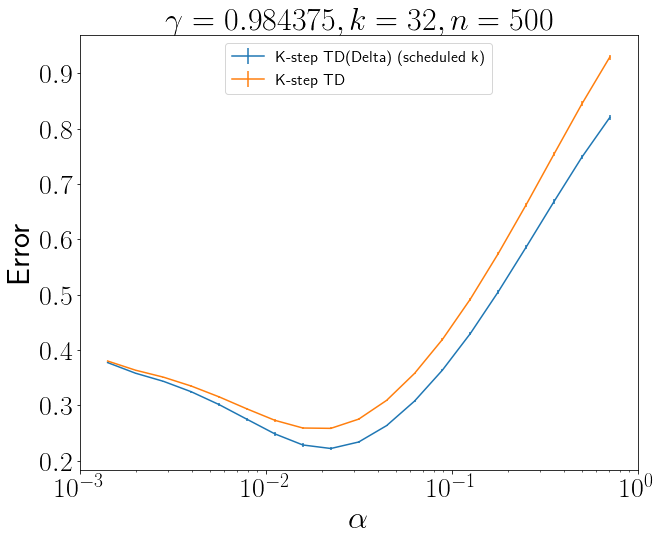}
    \includegraphics[width=.3\textwidth]{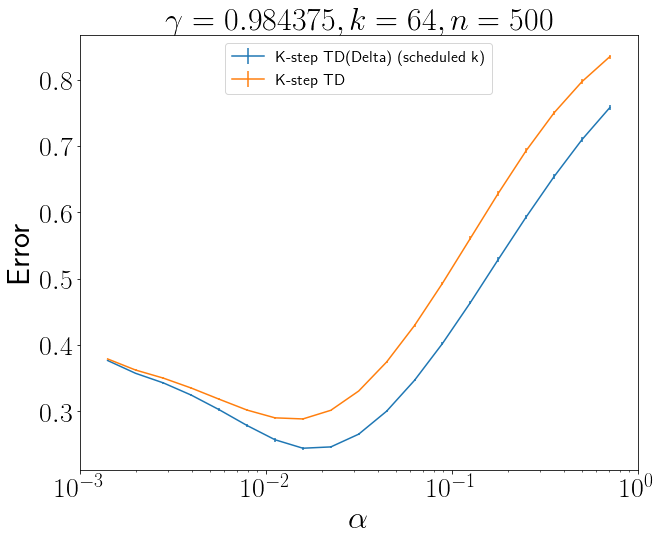}
    \includegraphics[width=.3\textwidth]{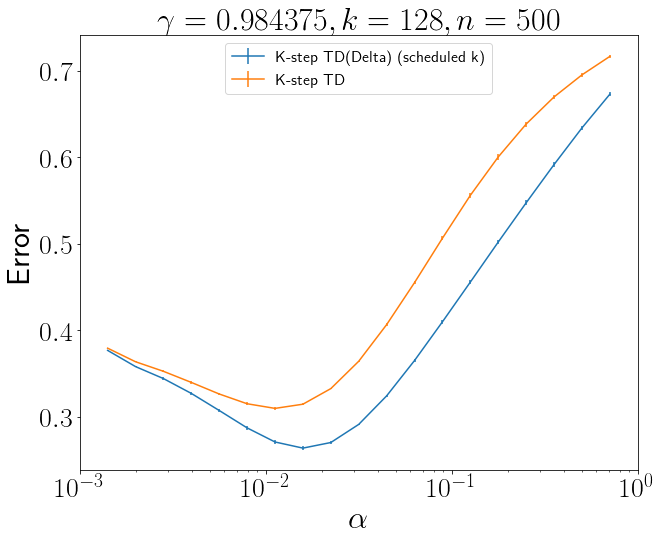}
    \includegraphics[width=.3\textwidth]{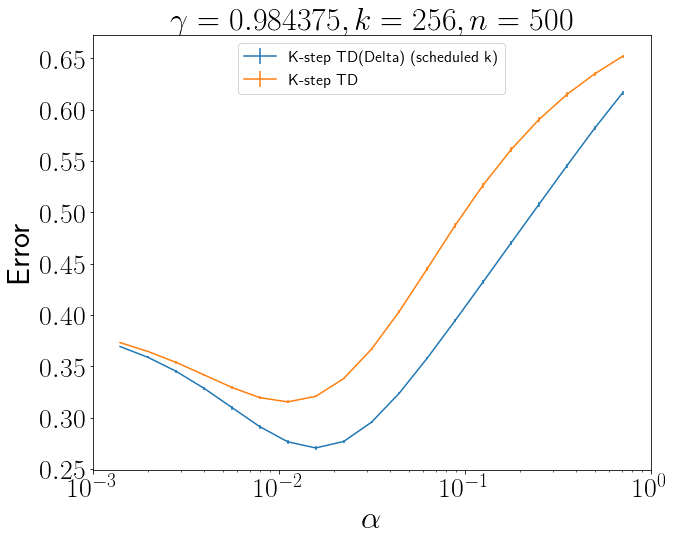}
    \caption{$\gamma=.98475$ and different k values at different learning rates, where the number of timesteps $n=500$.}
    \label{fig:g984375}
\end{figure}

\begin{figure}[!htbp]
    \centering
        \includegraphics[width=.3\textwidth]{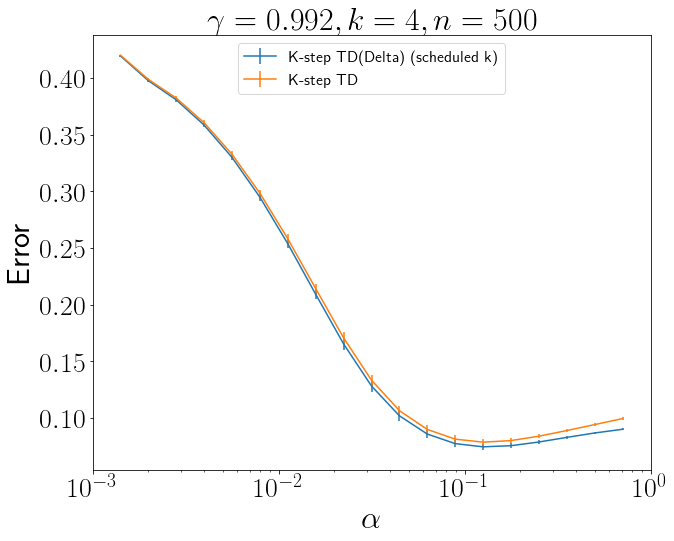}
    \includegraphics[width=.3\textwidth]{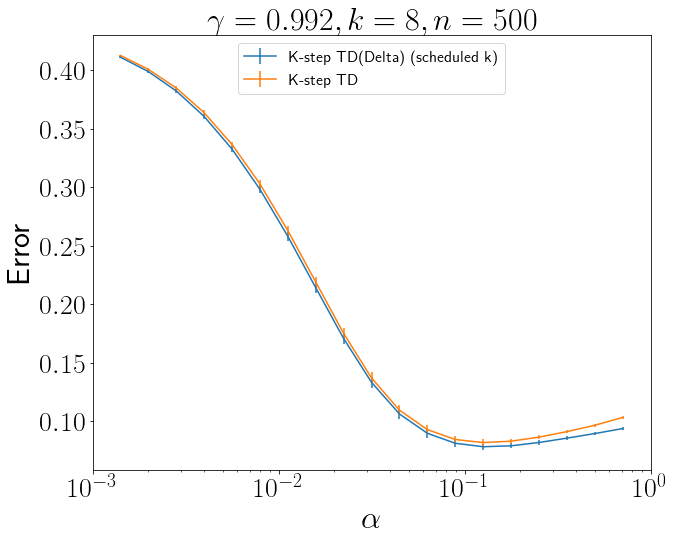}
    \includegraphics[width=.3\textwidth]{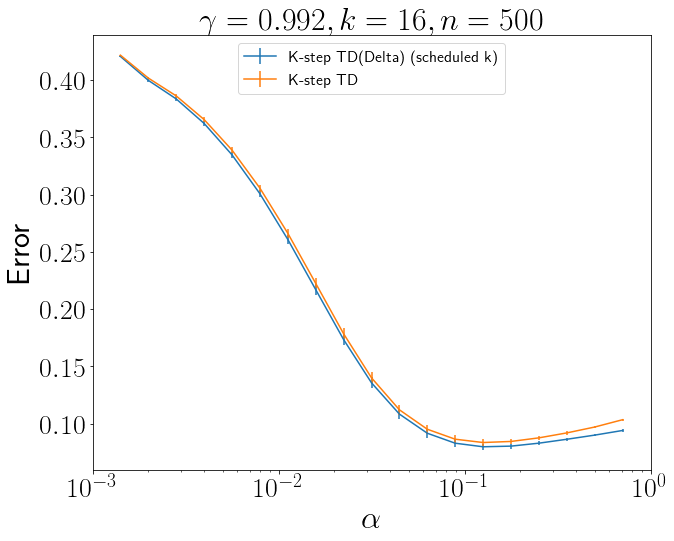}
    \includegraphics[width=.3\textwidth]{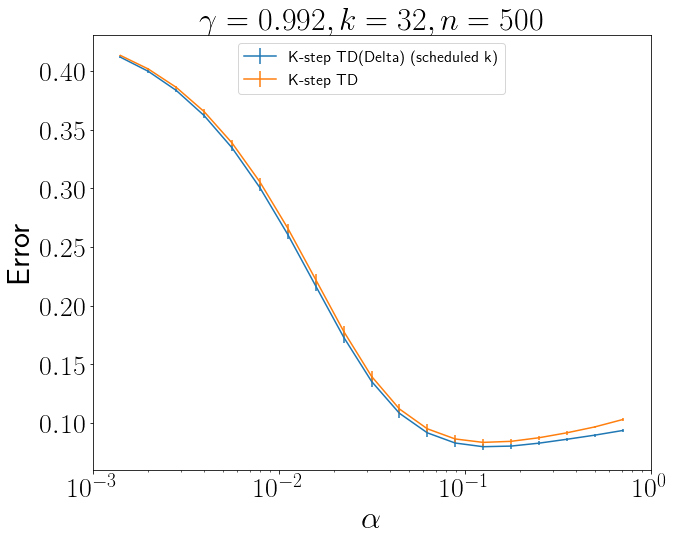}
    \includegraphics[width=.3\textwidth]{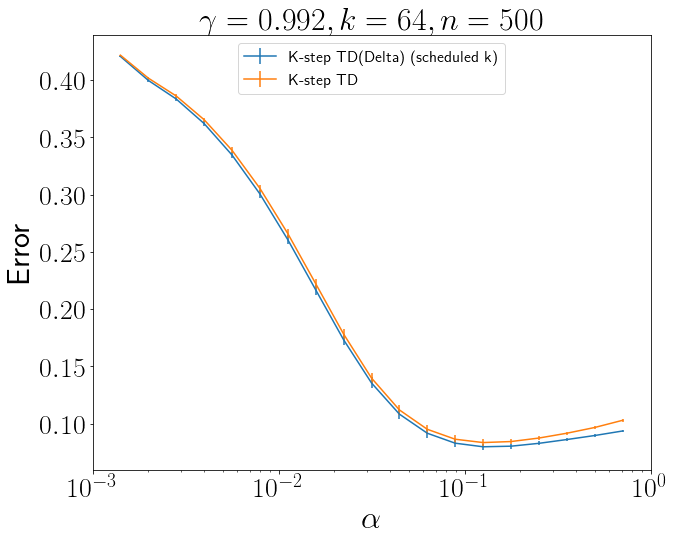}
    \includegraphics[width=.3\textwidth]{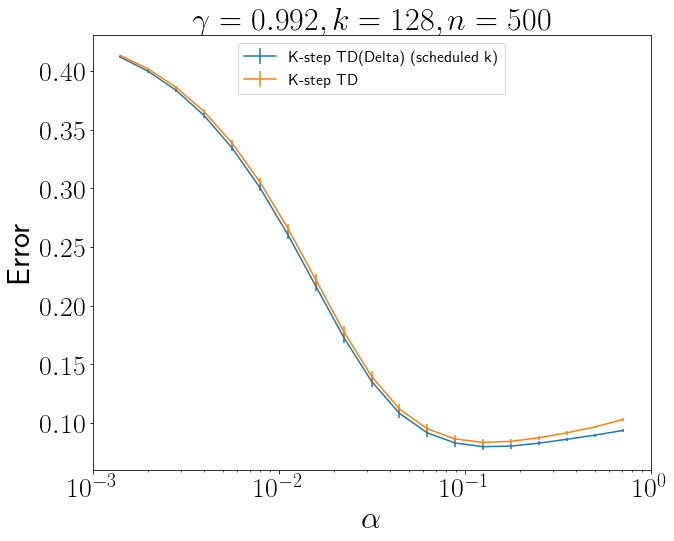}
    \includegraphics[width=.3\textwidth]{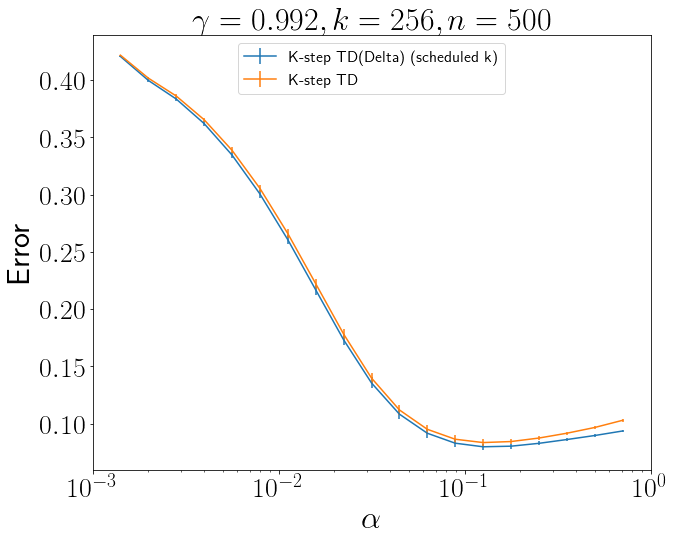}
    \caption{$\gamma=.992$ and different k values at different learning rates, where the number of timesteps $n=500$.}
    \label{fig:g992}
\end{figure}

\begin{figure}[!htbp]
    \centering
        \includegraphics[width=.3\textwidth]{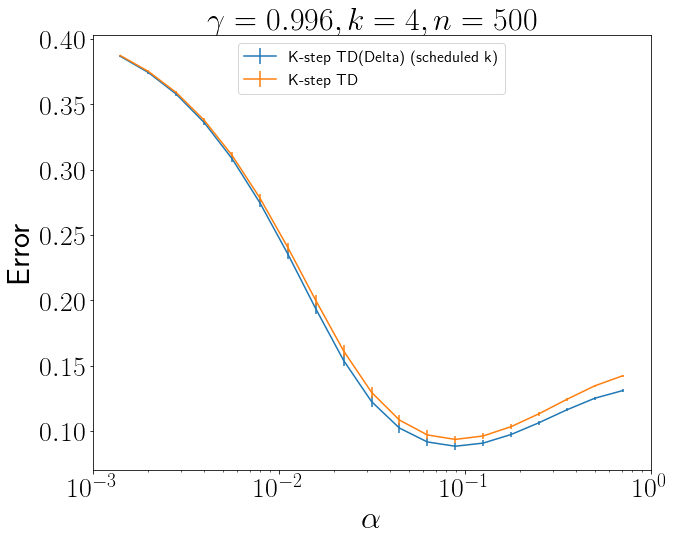}
    \includegraphics[width=.3\textwidth]{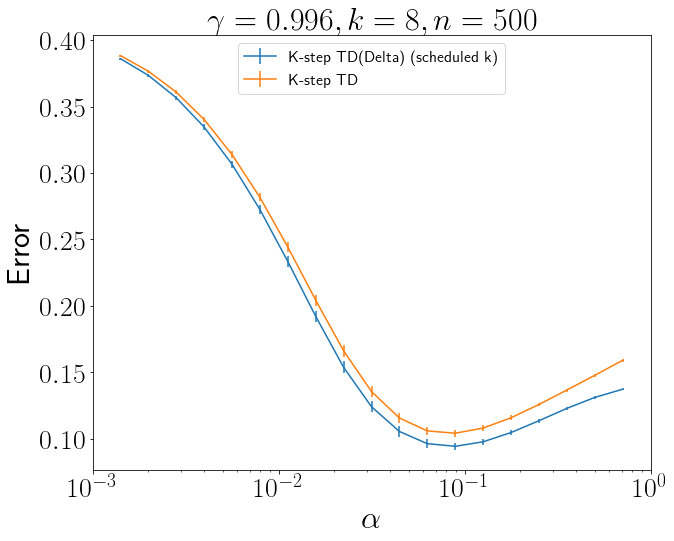}
    \includegraphics[width=.3\textwidth]{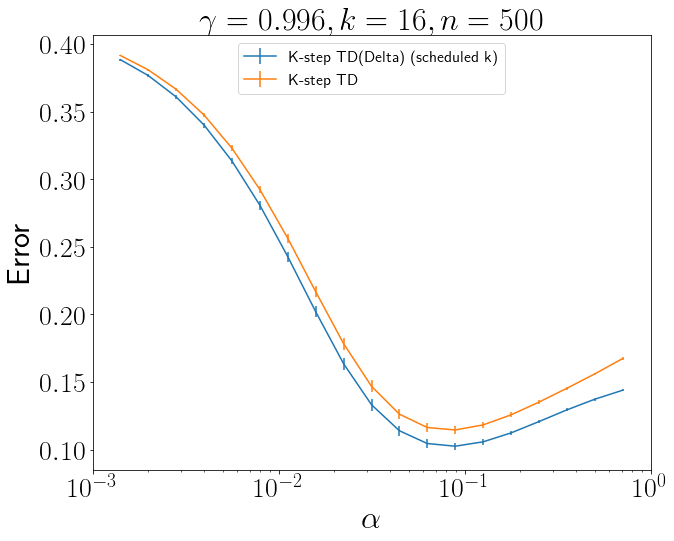}
    \includegraphics[width=.3\textwidth]{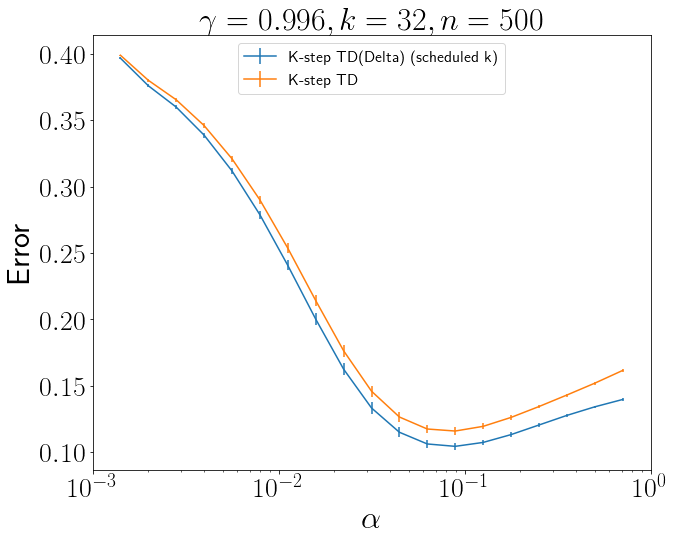}
    \includegraphics[width=.3\textwidth]{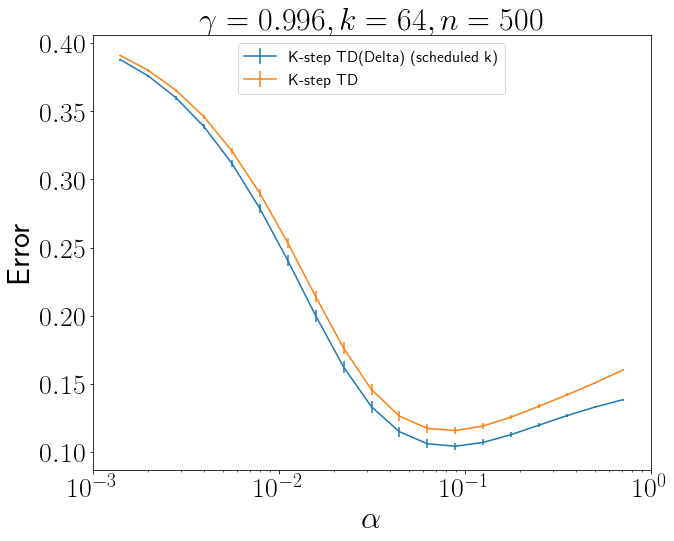}
    \includegraphics[width=.3\textwidth]{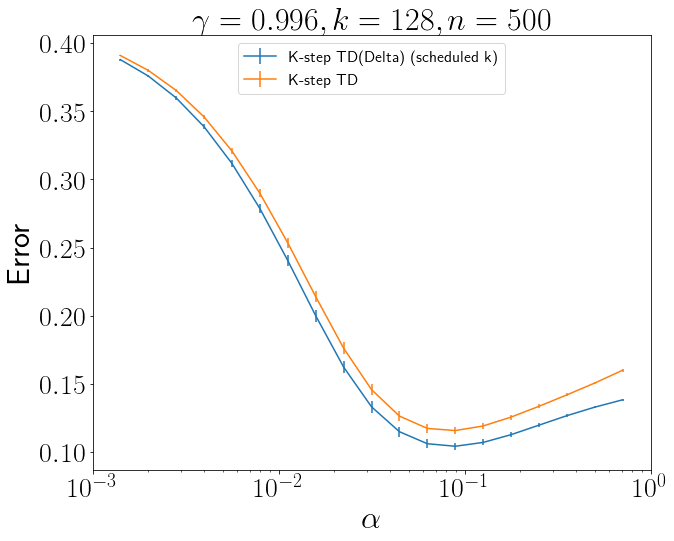}
    \includegraphics[width=.3\textwidth]{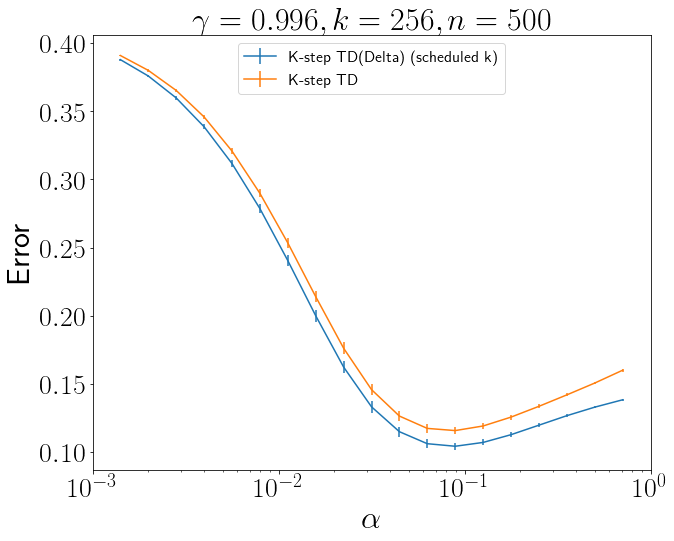}
    \caption{$\gamma=.996$ and different k values at different learning rates, where the number of timesteps $n=500$.}
    \label{fig:g996}
\end{figure}

\begin{figure}[!htbp]
    \centering
    \includegraphics[width=.3\textwidth]{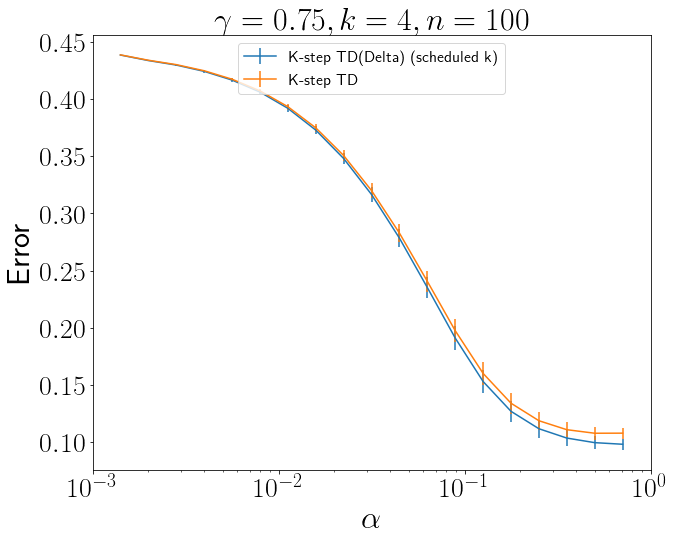}
    \includegraphics[width=.3\textwidth]{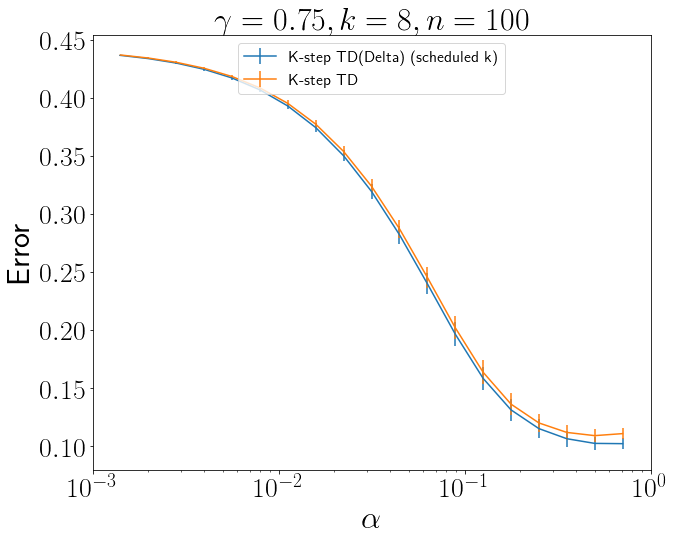}
    \includegraphics[width=.3\textwidth]{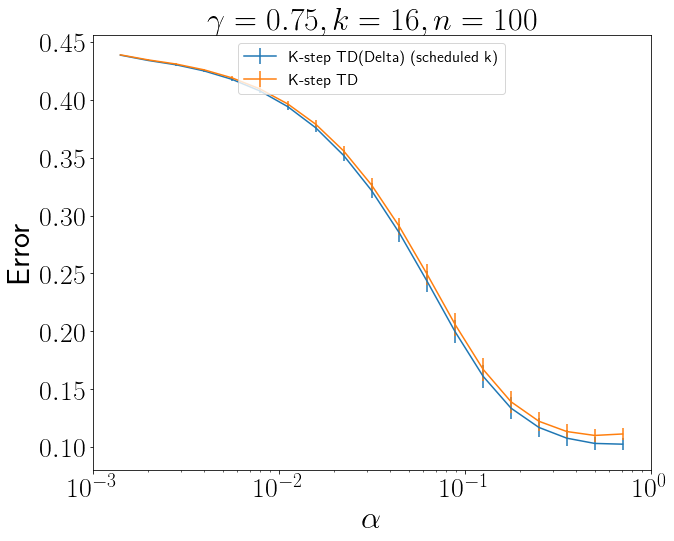}
    \includegraphics[width=.3\textwidth]{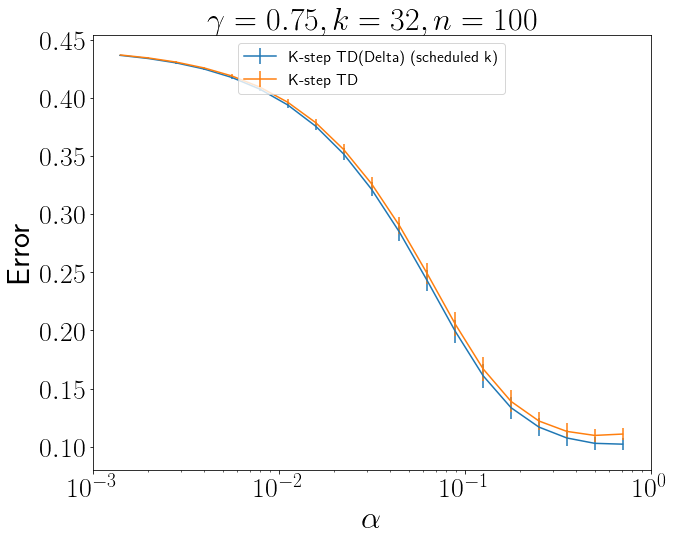}
    \includegraphics[width=.3\textwidth]{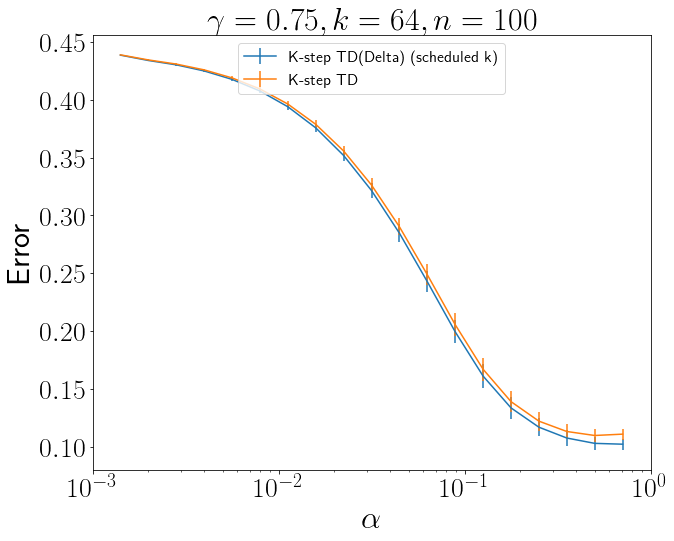}
    \includegraphics[width=.3\textwidth]{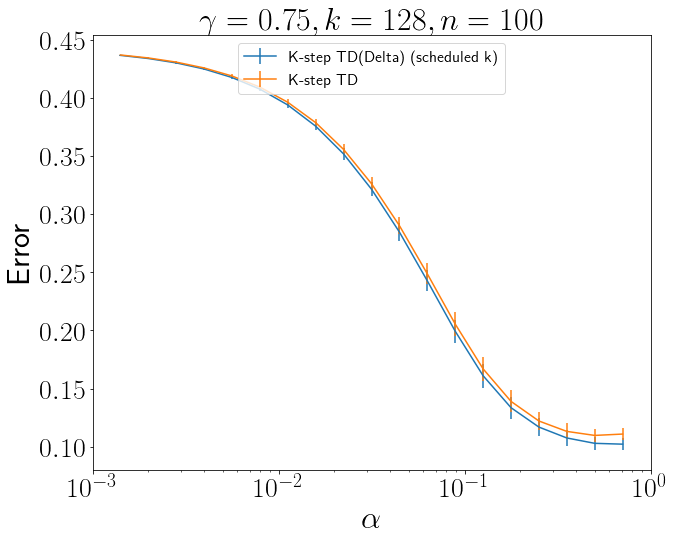}
    \includegraphics[width=.3\textwidth]{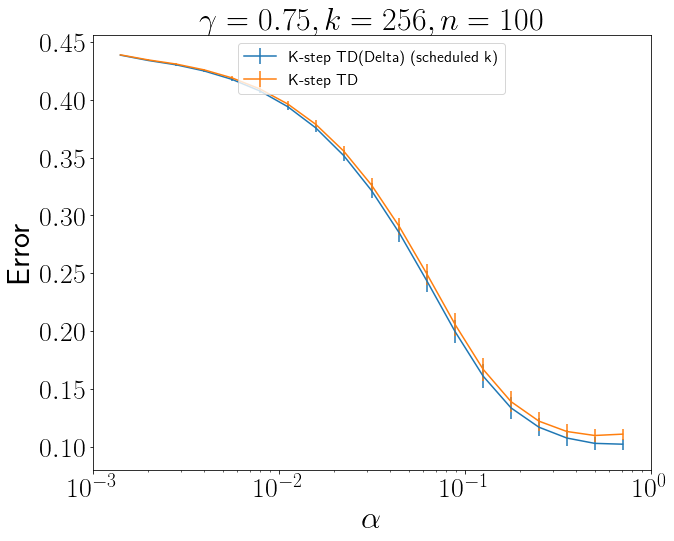}
    \caption{$\gamma=.75$ and different k values at different learning rates, number of timesteps in the environment $n=100$.}
    \label{fig:g_75}
\end{figure}

\begin{figure}[!htbp]
    \centering
    \includegraphics[width=.3\textwidth]{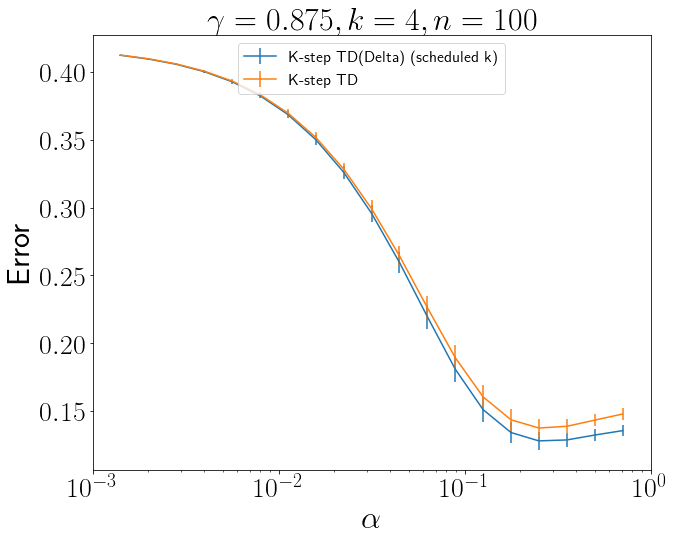}
    \includegraphics[width=.3\textwidth]{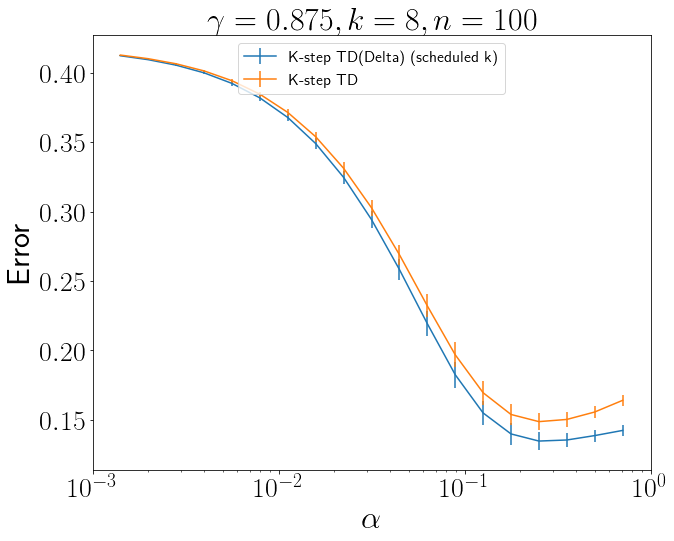}
    \includegraphics[width=.3\textwidth]{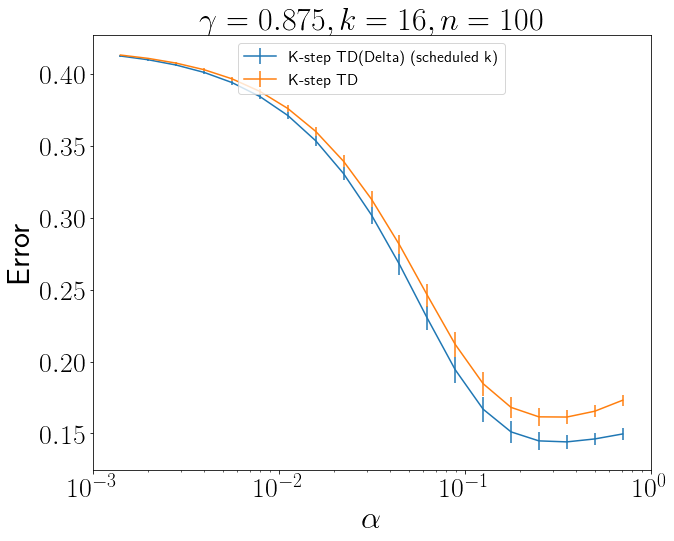}
    \includegraphics[width=.3\textwidth]{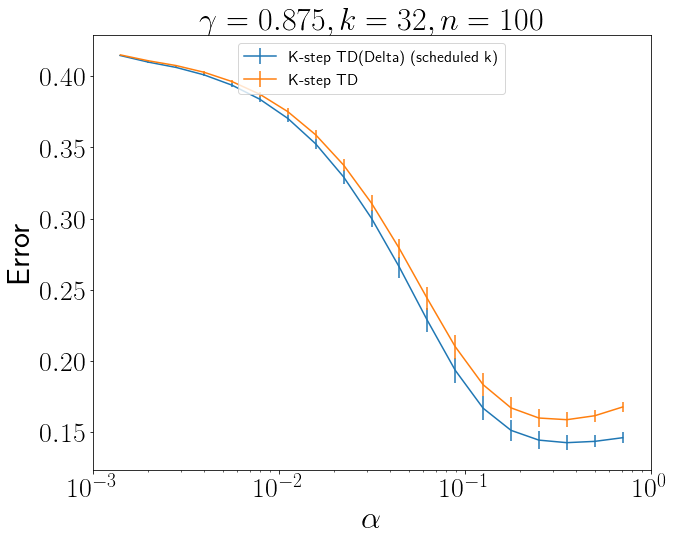}
    \includegraphics[width=.3\textwidth]{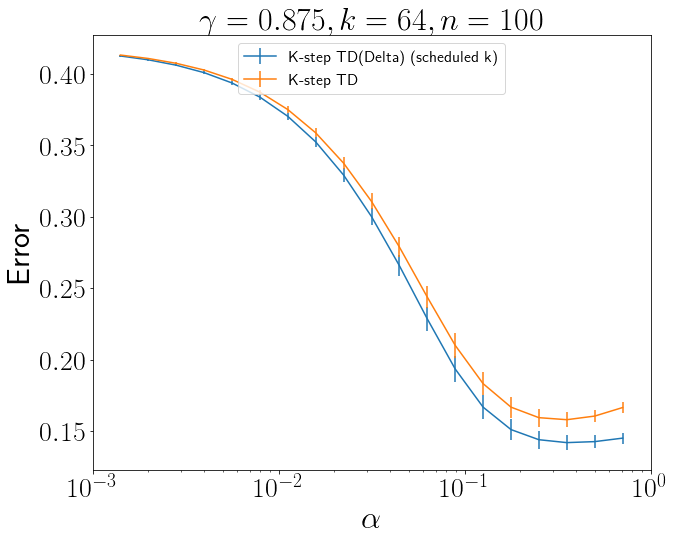}
    \includegraphics[width=.3\textwidth]{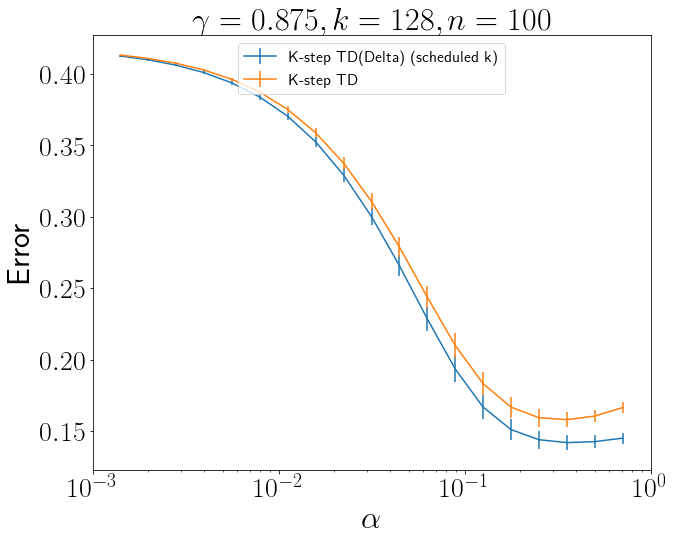}
    \includegraphics[width=.3\textwidth]{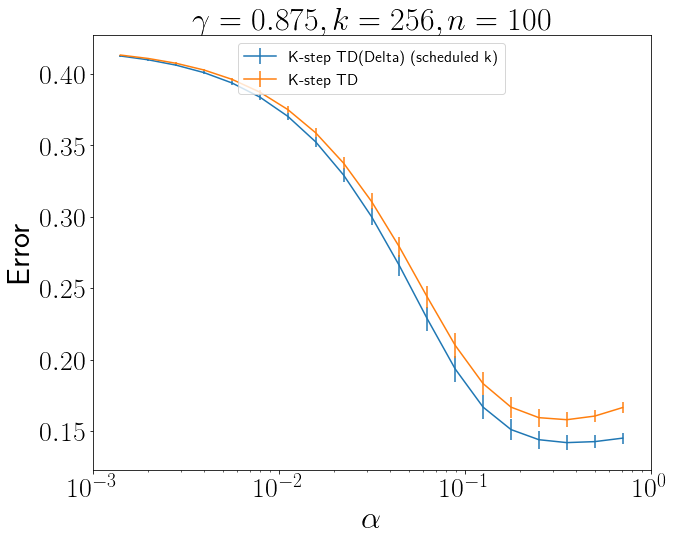}
    \caption{$\gamma=.875$ and different k values at different learning rates, number of timesteps in the environment $n=100$.}
    \label{fig:g875}
\end{figure}

\begin{figure}[!htbp]
    \centering
    \includegraphics[width=.3\textwidth]{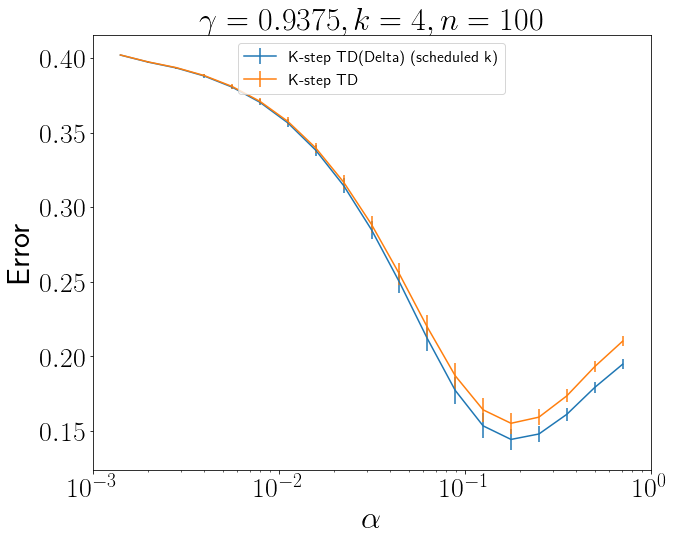}
    \includegraphics[width=.3\textwidth]{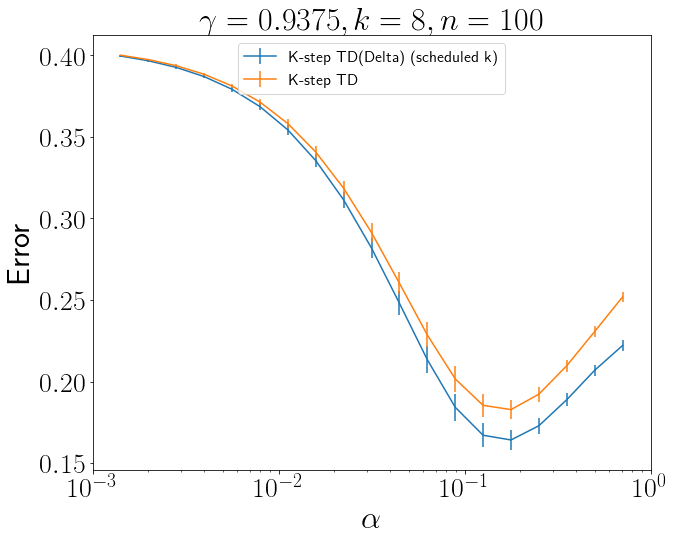}
    \includegraphics[width=.3\textwidth]{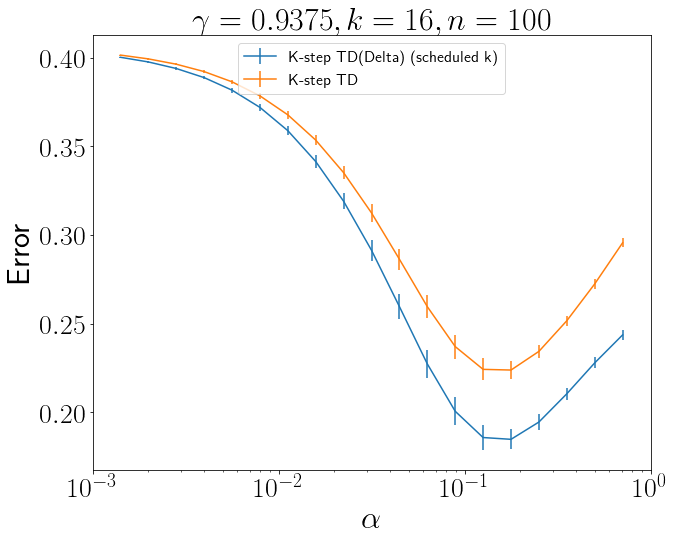}
    \includegraphics[width=.3\textwidth]{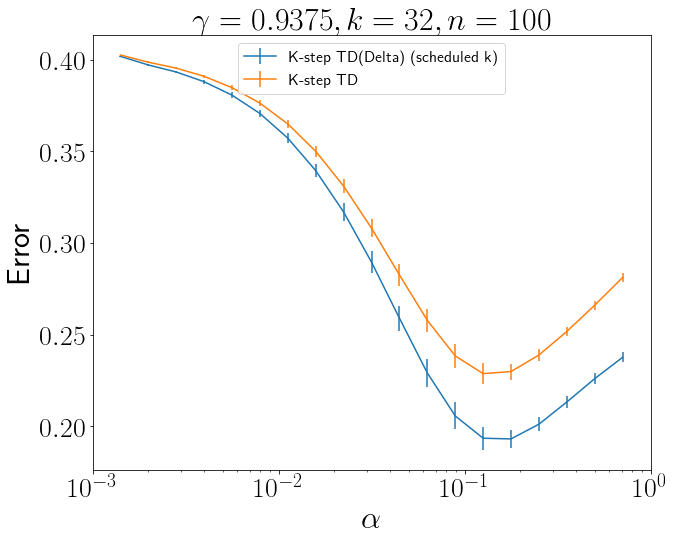}
    \includegraphics[width=.3\textwidth]{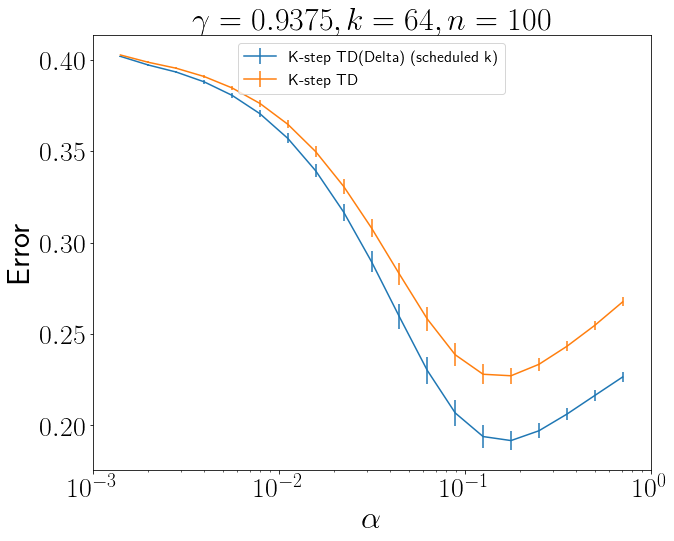}
    \includegraphics[width=.3\textwidth]{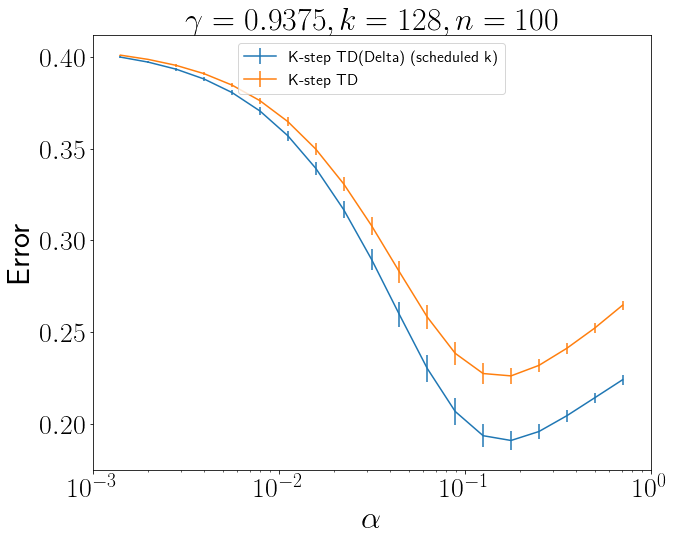}
    \includegraphics[width=.3\textwidth]{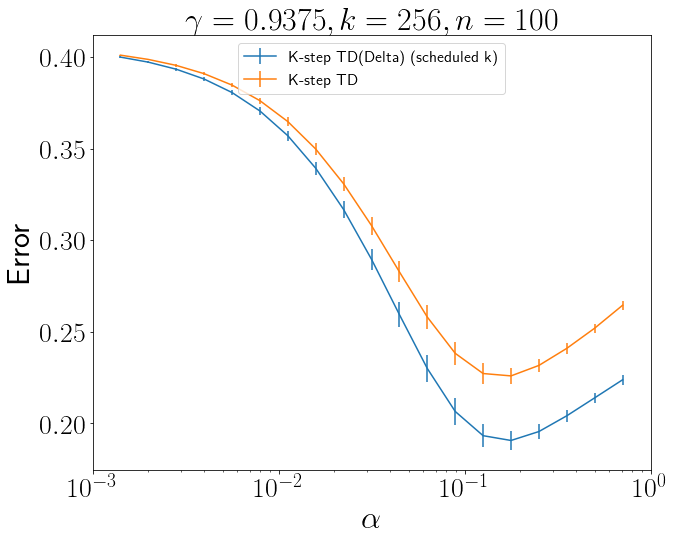}
    \caption{$\gamma=.93750$ and different k values at different learning rates, number of timesteps in the environment $n=100$.}
    \label{fig:g9375}
\end{figure}

\begin{figure}[!htbp]
    \centering
        \includegraphics[width=.3\textwidth]{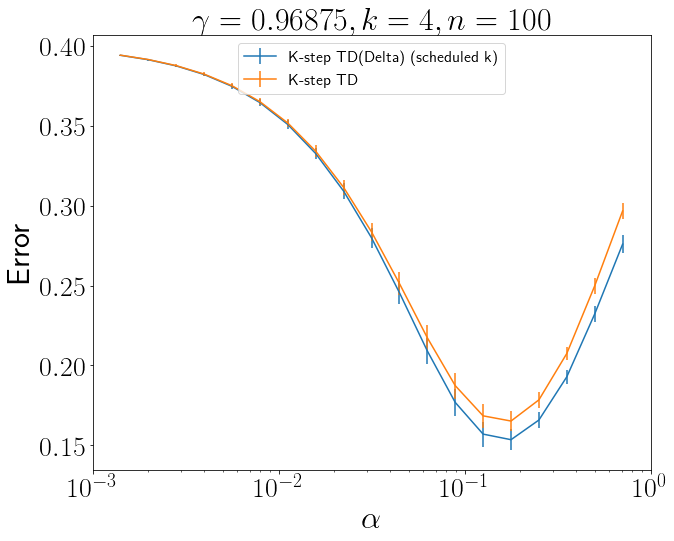}
    \includegraphics[width=.3\textwidth]{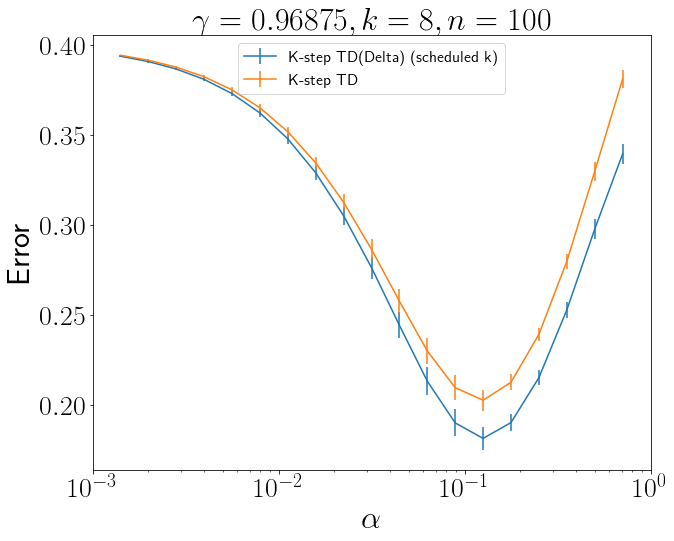}
    \includegraphics[width=.3\textwidth]{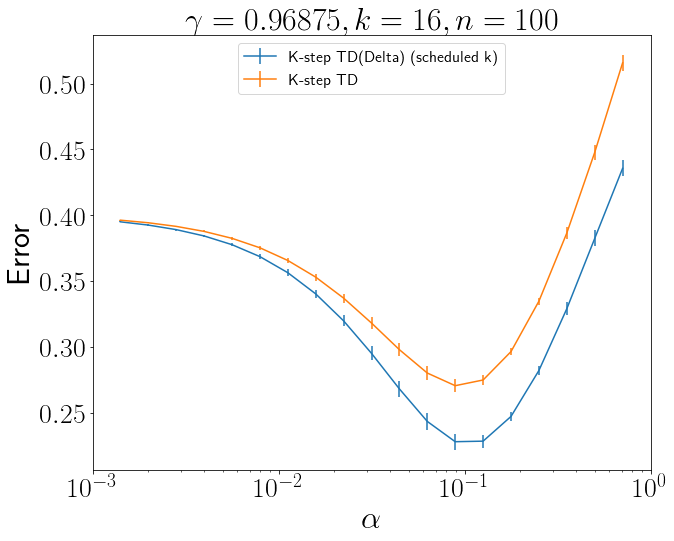}
    \includegraphics[width=.3\textwidth]{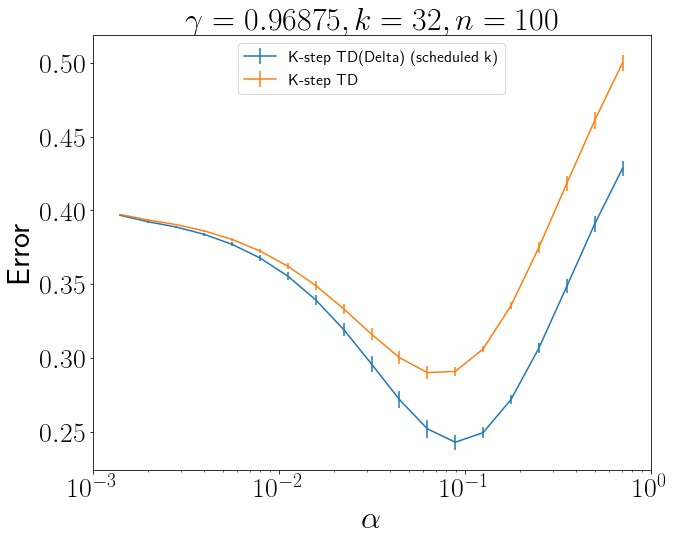}
    \includegraphics[width=.3\textwidth]{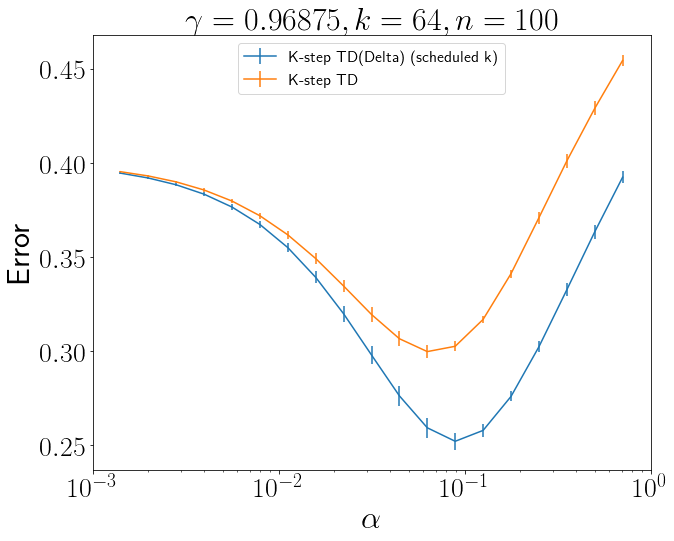}
    \includegraphics[width=.3\textwidth]{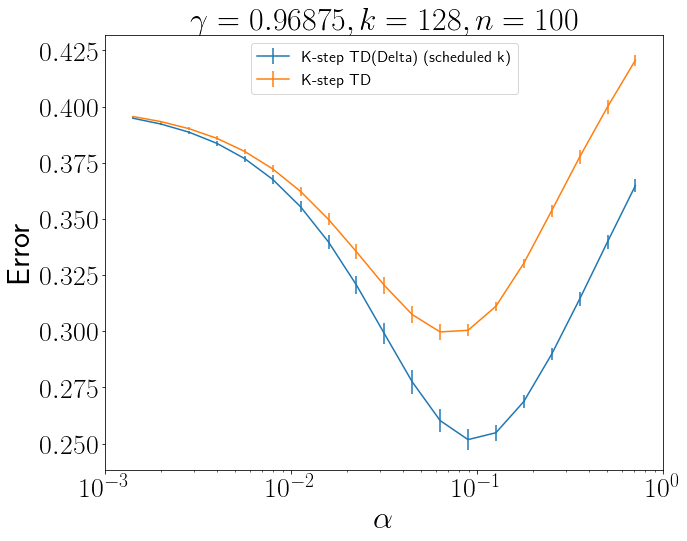}
    \includegraphics[width=.3\textwidth]{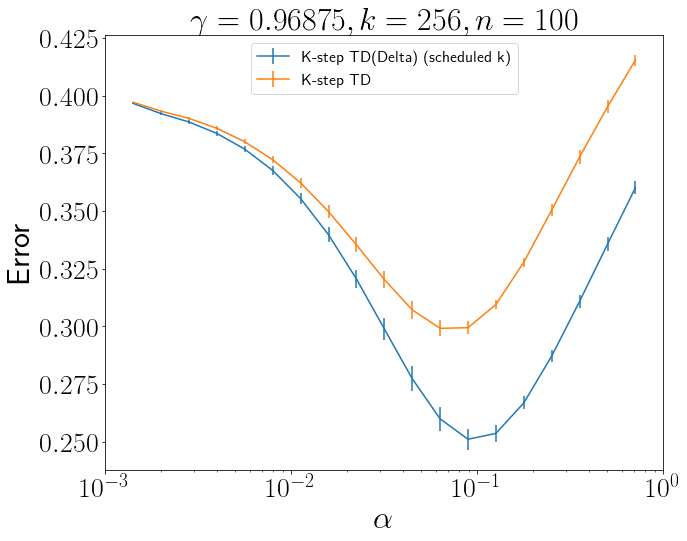}
    \caption{$\gamma=.96875$ and different k values at different learning rates, number of timesteps in the environment $n=100$.}
    \label{fig:g96875}
\end{figure}

\begin{figure}[!htbp]
    \centering
        \includegraphics[width=.3\textwidth]{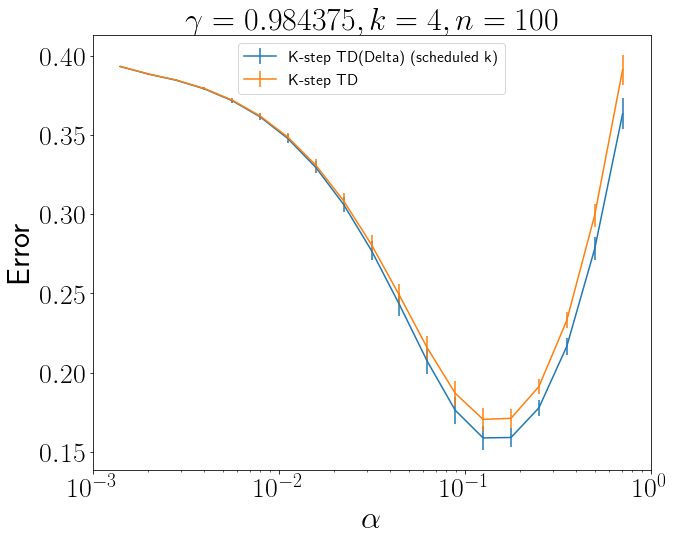}
    \includegraphics[width=.3\textwidth]{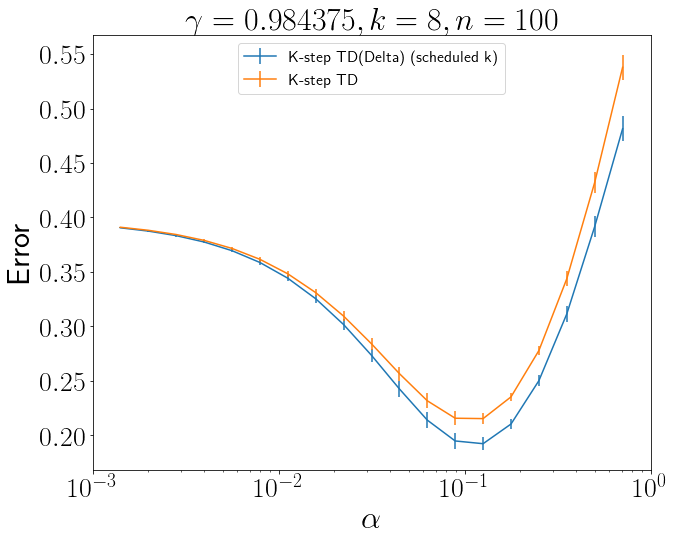}
    \includegraphics[width=.3\textwidth]{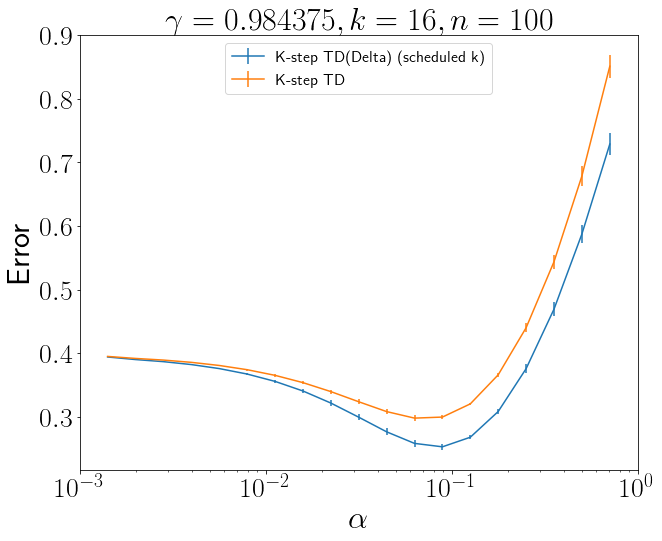}
    \includegraphics[width=.3\textwidth]{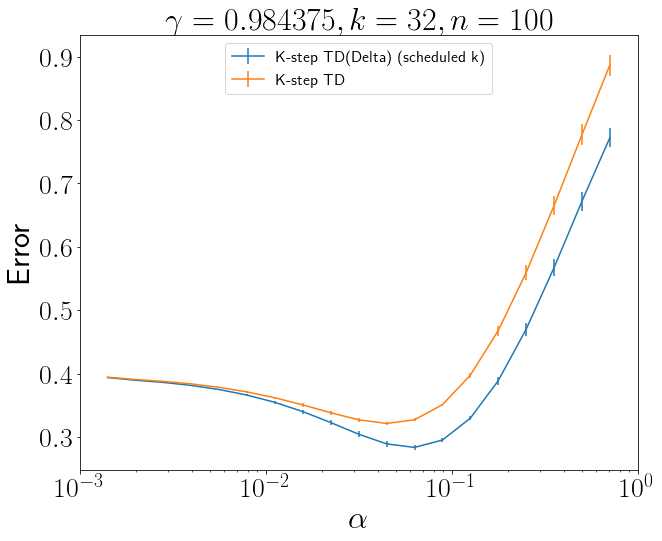}
    \includegraphics[width=.3\textwidth]{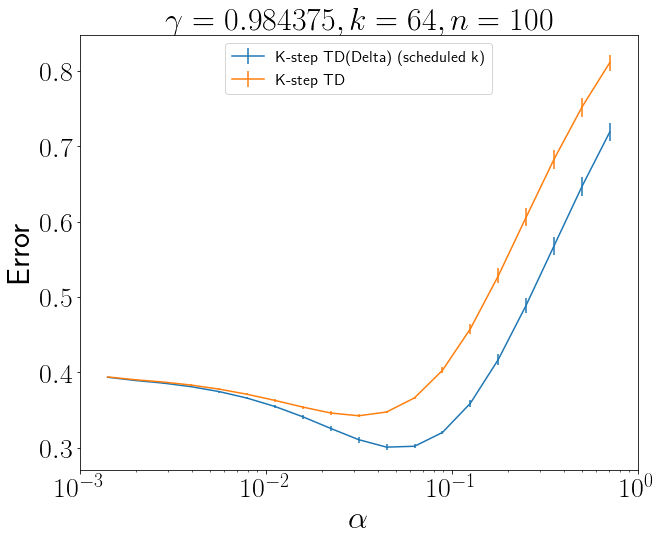}
    \includegraphics[width=.3\textwidth]{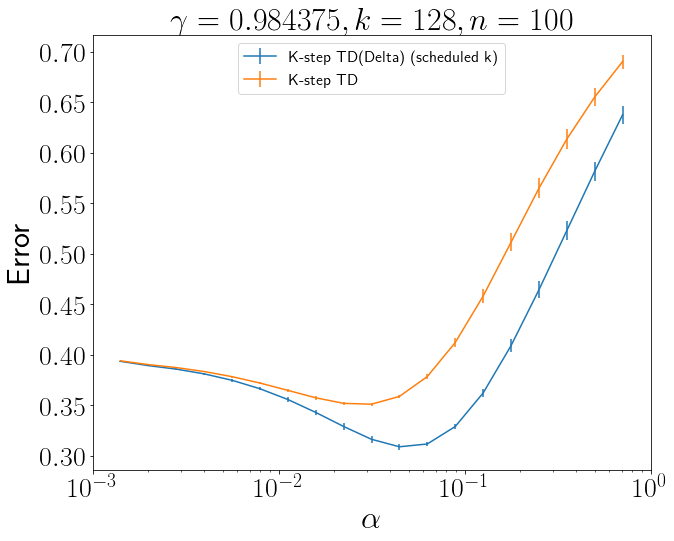}
    \includegraphics[width=.3\textwidth]{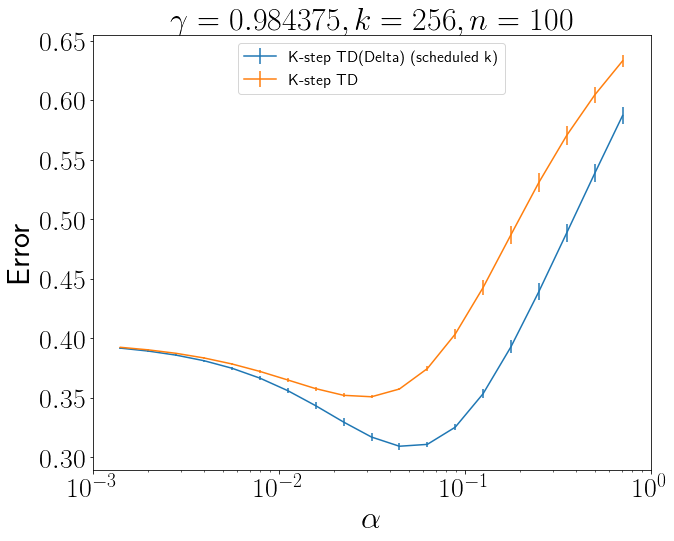}
    \caption{$\gamma=.98475$ and different k values at different learning rates, number of timesteps in the environment $n=100$.}
    \label{fig:g984375}
\end{figure}

\begin{figure}[!htbp]
    \centering
        \includegraphics[width=.3\textwidth]{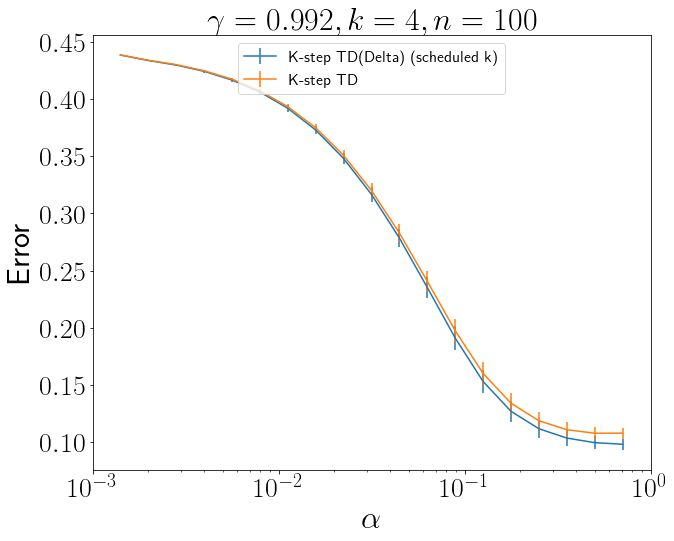}
    \includegraphics[width=.3\textwidth]{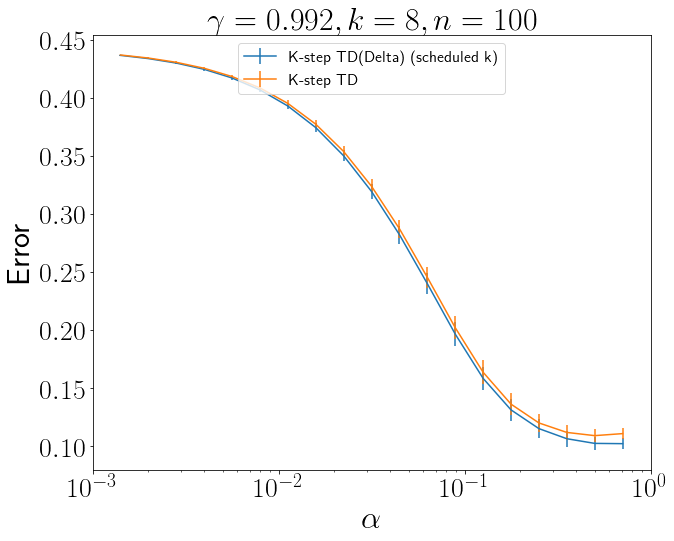}
    \includegraphics[width=.3\textwidth]{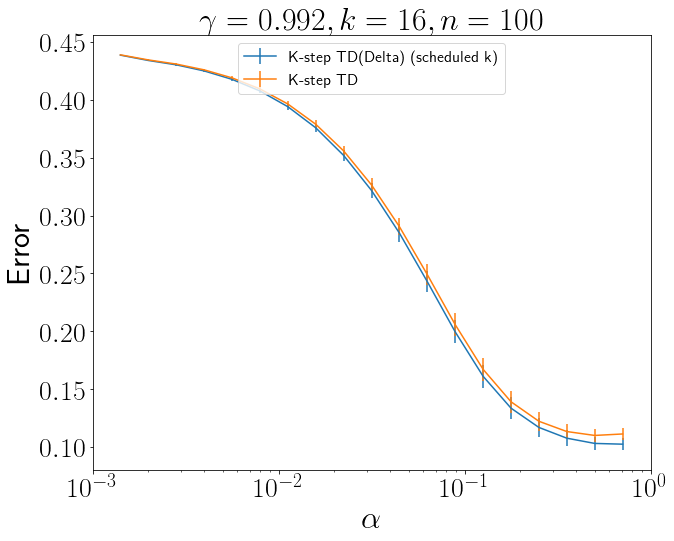}
    \includegraphics[width=.3\textwidth]{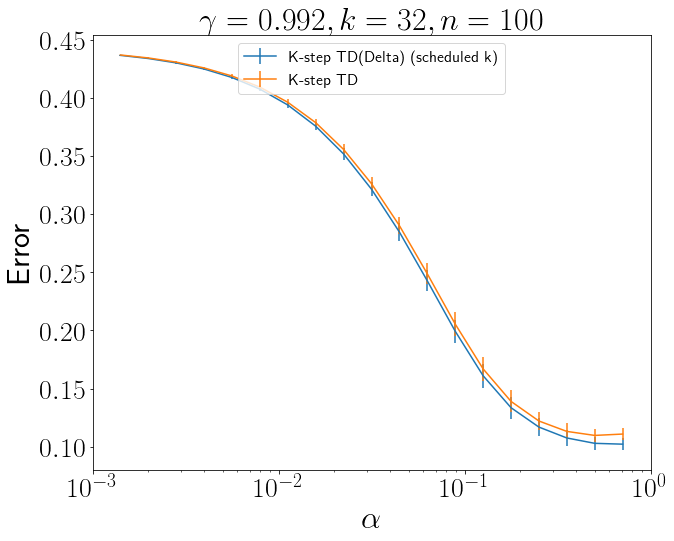}
    \includegraphics[width=.3\textwidth]{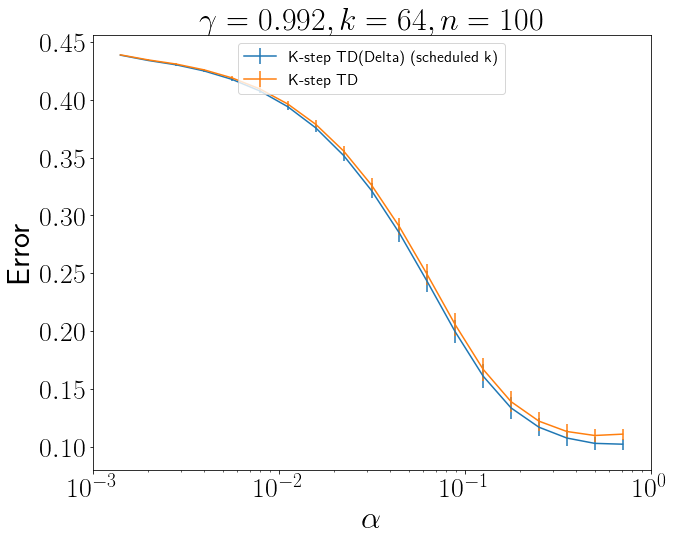}
    \includegraphics[width=.3\textwidth]{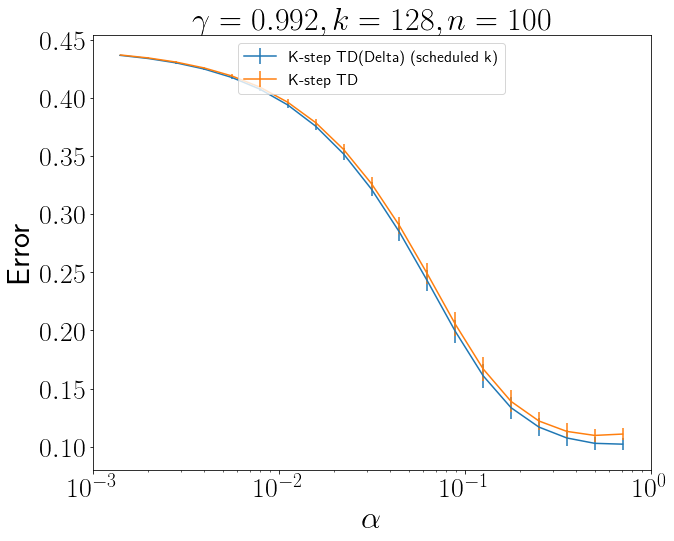}
    \includegraphics[width=.3\textwidth]{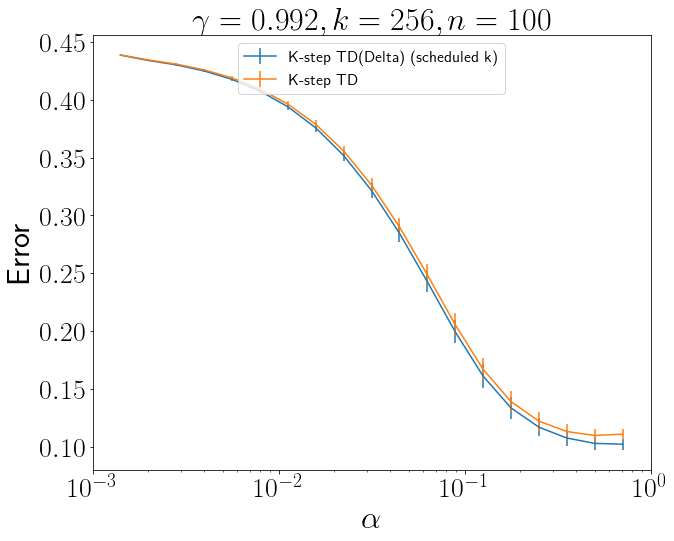}
    \caption{$\gamma=.992$ and different k values at different learning rates, number of timesteps in the environment $n=100$.}
    \label{fig:g992}
\end{figure}

\begin{figure}[!htbp]
    \centering
        \includegraphics[width=.3\textwidth]{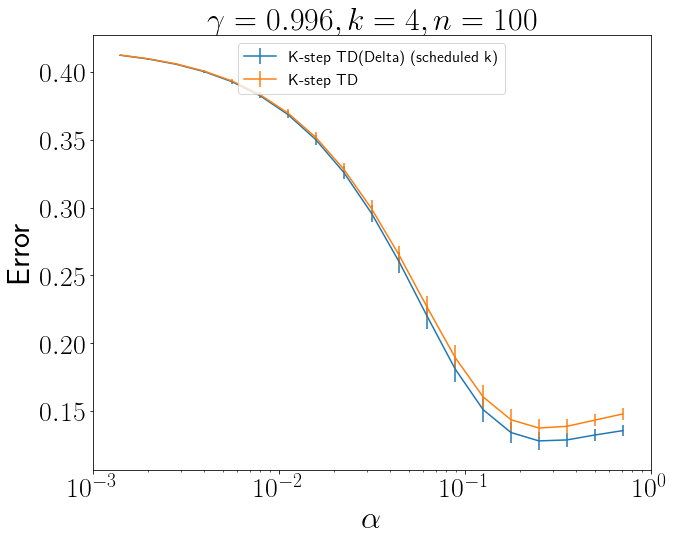}
    \includegraphics[width=.3\textwidth]{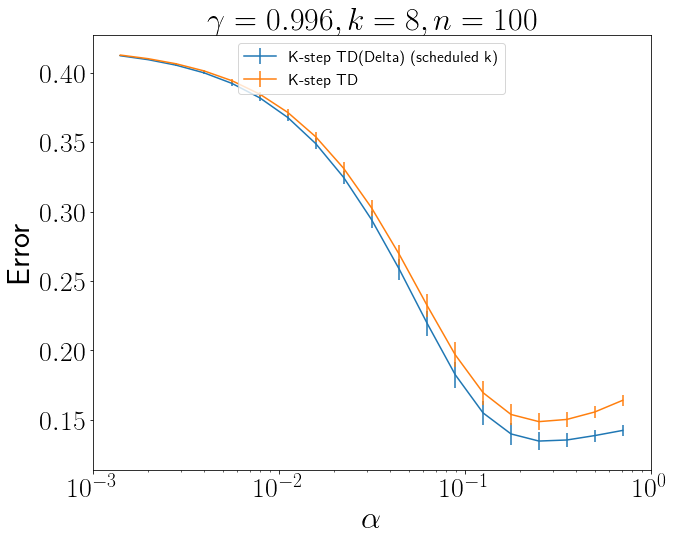}
    \includegraphics[width=.3\textwidth]{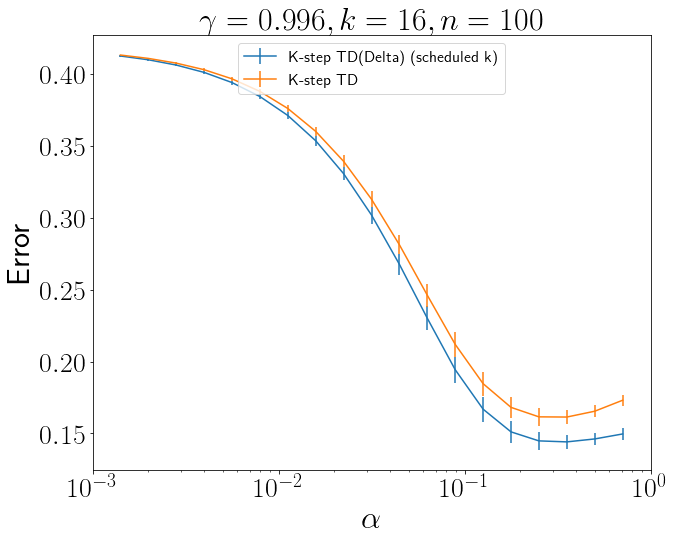}
    \includegraphics[width=.3\textwidth]{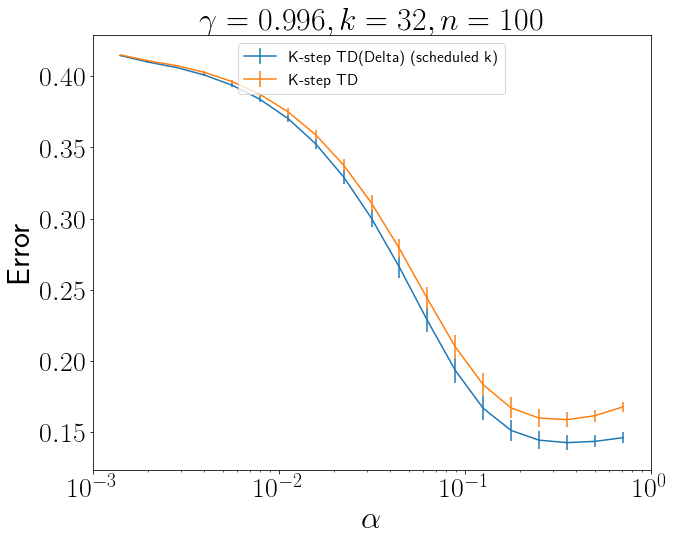}
    \includegraphics[width=.3\textwidth]{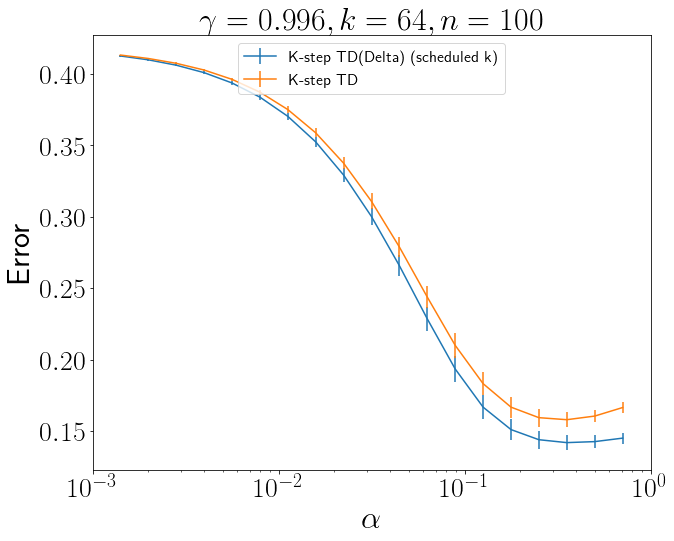}
    \includegraphics[width=.3\textwidth]{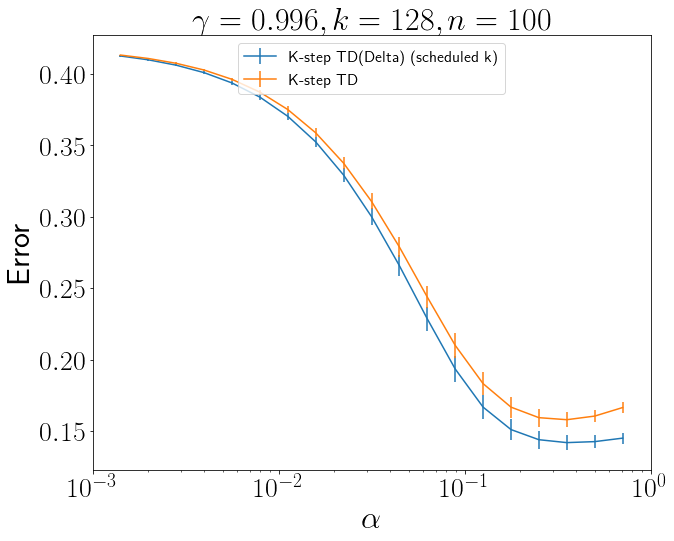}
    \includegraphics[width=.3\textwidth]{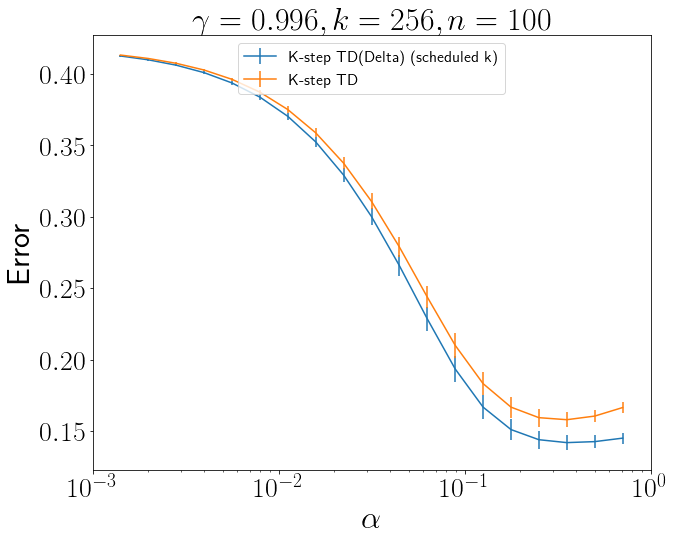}
    \caption{$\gamma=.996$ and different k values at different learning rates, number of timesteps in the environment $n=100$.}
    \label{fig:g996}
\end{figure}

\FloatBarrier
\subsection{Atari}
We use the NoFrameSkip-v4 version of all environments.

\subsubsection{Hyperparameters}

We use mostly the same hyperparameters as suggested by~\citet{schulman2017proximal,pytorchrl}. We use seeds $[125125, 513, 90135, 81212, 3523401, 15709, 17, 0, 8412, 1153780]$ generated randomly. For the TD($\Delta$) version of the algorithm we set $\gamma_z$ such that $\gamma_Z= \gamma$ and then we set $\gamma_{z-1} = \frac{1}{2}(\frac{1}{(1-\gamma_z)})$ while $\gamma_z > .5$. The rest of the hyperparameter settings can be found in Table~\ref{tab:hyperparameters}.

\begin{table}[!htbp]
    \centering
    \begin{tabular}{|c|c|}
    \hline
         Hyperparameter& Value  \\
         \hline
         Learning Rate & $2.5 \times 10^{-4}$\\
         Clipping parameter & $0.1$\\
         VF coeff & 1\\
         Number  of actors & 8\\
         Horizon(T) & 128\\
         Number of Epochs & 4\\
         Entropy Coeff. & 0.01\\
         Discount ($\gamma, \gamma_Z$)& .99\\
         \hline
    \end{tabular}
    \caption{Hyperparameters common to both PPO baseline and PPO with TD($\Delta$, GAE).}
    \label{tab:hyperparameters}
\end{table}

\subsubsection{Results}

\begin{table*}[!htbp]
    \centering
    \resizebox{\textwidth}{!}{    \begin{tabular}{|c|c|c|c|c|c|c|c|c|c|}
    \hline
        Algorithm & Zaxxon & WizardOfWor & \textbf{Qbert} & \textbf{MsPacman} & \textbf{Hero} & \textbf{Frostbite}  & BankHeist & \textbf{Amidar} & \textbf{Alien} \\
         \hline
          PPO-TD($\lambda, \Delta$) &451 $\pm$ 252&2069 $\pm$ 140 &13296 $\pm$ 576 $\dagger$&2336 $\pm$ 87 $\dagger$&29224 $\pm$ 742$\dagger$&267 $\pm$ 3&1176 $\pm$ 25&740 $\pm$ 31 $\dagger$&1508 $\pm$ 94\\
        \hline
        PPO-TD($\hat \lambda, \Delta$)  &3227 $\pm$ 797& 2402 $\pm$ 154&13328 $\pm$ 409 $\dagger$&2224 $\pm$ 75 $\dagger$&28913 $\pm$ 1036 $\dagger$&267 $\pm$ 10&1163 $\pm$ 12&668 $\pm$ 60 &1626 $\pm$ 109\\
              \hline
        PPO+  &6890 $\pm$ 269$\dagger$&2821 $\pm$ 351$\dagger$&10521 $\pm$ 622&1907 $\pm$ 94&23670 $\pm$ 961&269 $\pm$ 2& 1185 $\pm$ 10&605 $\pm$ 32& 1421 $\pm$ 96\\

                    \hline
        PPO  &7205 $\pm$ 208 $\dagger$& 3449 $\pm$ 176$\dagger$&11845 $\pm$ 301&1927 $\pm$ 111&21052 $\pm$ 1072&268 $\pm$ 2&1179 $\pm$ 9&604 $\pm$ 49 &1228 $\pm$ 63\\
          \hline
    \end{tabular}}
    \caption{Average returns on fully trained policy for Atari on 30 episodes each with a different amount of no-ops at the start as done by \citet{mnih2013playing}. Shows the mean across 10 seeds and the standard error. $\dagger$ denotes significantly better results over our algorithm in the case of baselines or over the best baseline in the case of our algorithm using Welch's t-test with a significance level of $.05$ per \citet{henderson2018deep,colas2018many} and bootstrap confidence intervals using the script by \citet{colas2018many} to process the data. Bold algorithms are where we perform as well as or significantly better than the baselines.}
    \label{tab:averageholdout}
\end{table*}

\begin{table*}[!htbp]
    \centering
    \resizebox{\textwidth}{!}{    \begin{tabular}{|c|c|c|c|c|c|c|c|c|c|}
    \hline
        Algorithm & Zaxxon & WizardOfWor & \textbf{Qbert} & \textbf{MsPacman} & \textbf{Hero} & \textbf{Frostbite}  & BankHeist & \textbf{Amidar} & \textbf{Alien} \\
         \hline
          PPO-TD($\lambda, \Delta$) &199 $\pm$ 63& 1653 $\pm$ 61&7812 $\pm$ 254 $\dagger$&1729 $\pm$ 50 $\dagger$&17805 $\pm$ 263 $\dagger$&276 $\pm$ 3&732 $\pm$ 25& 432 $\pm$ 18 $\dagger$& 1171 $\pm$ 64 $\dagger$\\
                    \hline
              PPO-TD($\hat \lambda, \Delta$)  & 1259 $\pm$ 286 & 1747 $\pm$ 47&8078 $\pm$ 201 $\dagger$&1704 $\pm$ 35 $\dagger$&16846 $\pm$ 289 $\dagger$&276 $\pm$ 7&717 $\pm$ 38&444 $\pm$ 20$\dagger$&1352 $\pm$ 65$\dagger$\\
              \hline
          PPO+ &3647 $\pm$ 288 $\dagger$& 2073 $\pm$ 90 $\dagger$&5493 $\pm$ 154&1302 $\pm$ 35&14116 $\pm$ 294&276 $\pm$ 0&953 $\pm$ 12 $\dagger$&367 $\pm$ 10&997 $\pm$ 49\\
                    \hline
          PPO &3776 $\pm$ 284 $\dagger$&2247 $\pm$ 70 $\dagger$ &5763 $\pm$ 217&1275 $\pm$ 35&13146 $\pm$ 370&267 $\pm$ 2&928 $\pm$ 13 $\dagger$& 395 $\pm$ 15&973 $\pm$ 34\\
          \hline
                    \end{tabular}}
    \caption{Average Atari performance (across all training episodes) with the mean and standard error across 10 seeds. See Table~\ref{tab:asymptotic} for more details.}
        \label{tab:average}
\end{table*}

\begin{figure}[!htbp]
    
    \includegraphics[width=.45\textwidth]{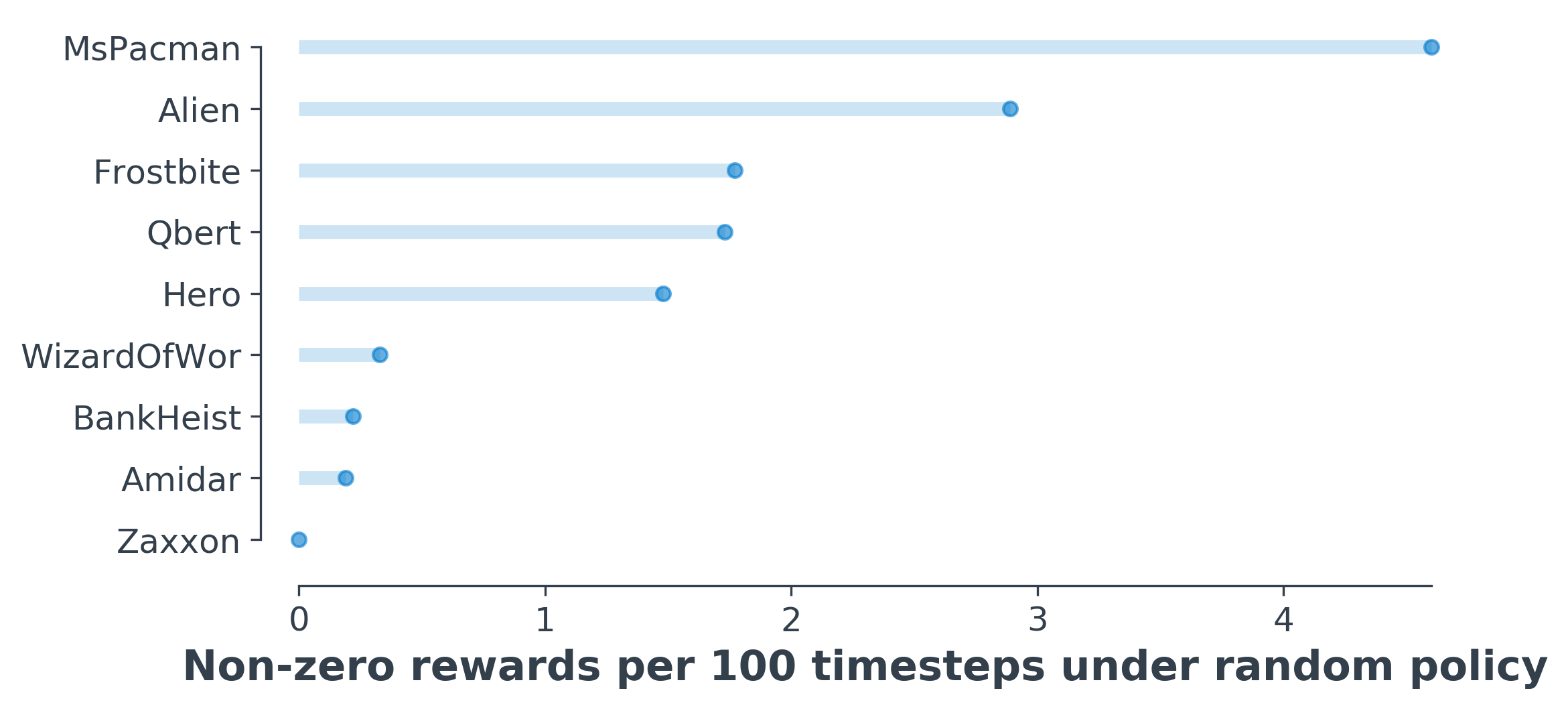}
        \includegraphics[width=.45\textwidth]{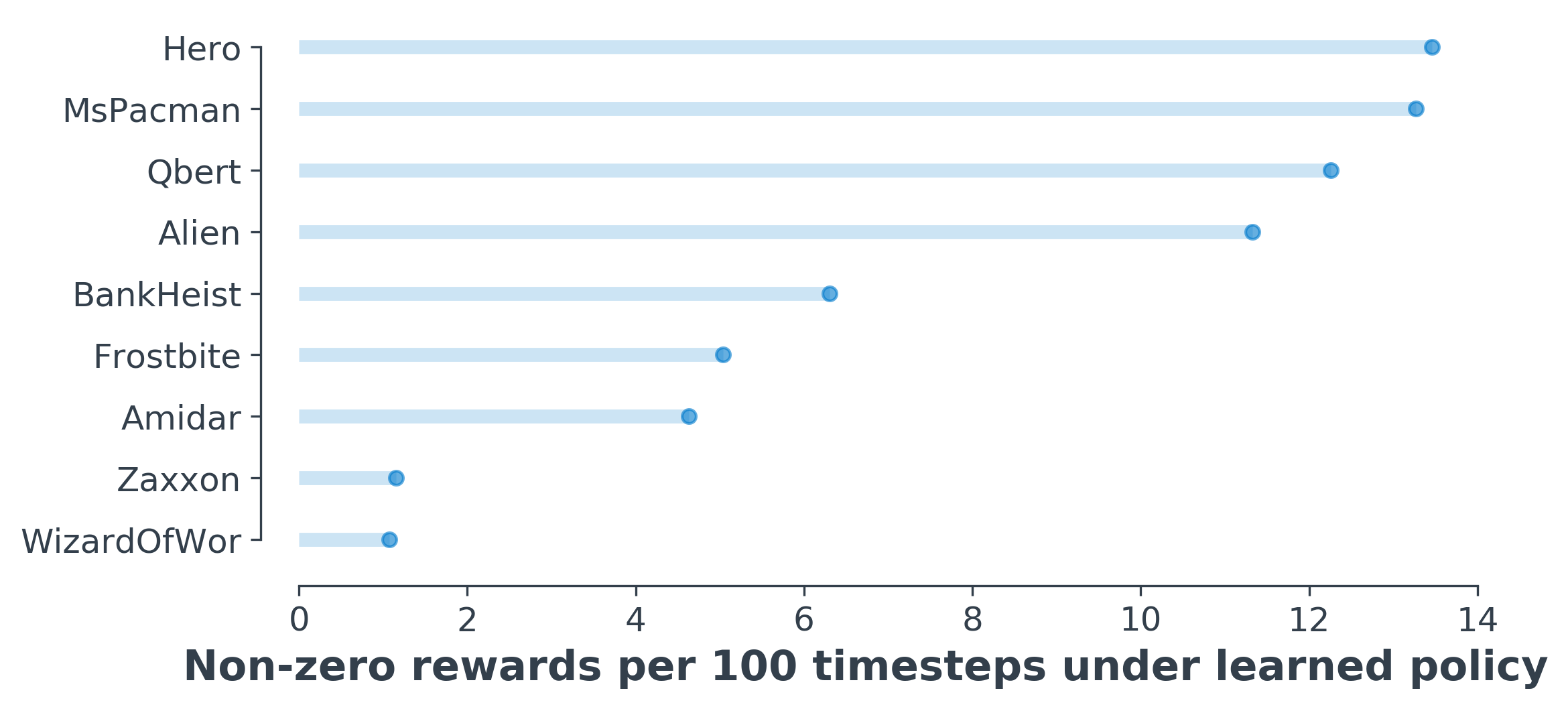}
    \caption{Frequency of rewards per $100$ time-steps averaged over $10000$ time-steps. Notice how the task `Zaxxon' has a much lower frequency (approximately 2 orders of magnitude) than the largest frequency task (Ms Pacman).}
    \label{fig:freq2}
\end{figure}

Frequencies of random policies vs learned policies can be seen in Figure~\ref{fig:freq2}. In the case of Frostbite, we achieve equal performance in all algorithm cases. We suspect this to be a limitation of on-policy learning with PPO and achieves similar results to the original codebase from \citet{schulman2017proximal}. Using more advanced exploration and off-policy learning may change the local minimum which seems to be found here.

\begin{figure}[!htbp]
    \centering
    \includegraphics[width=.32\textwidth]{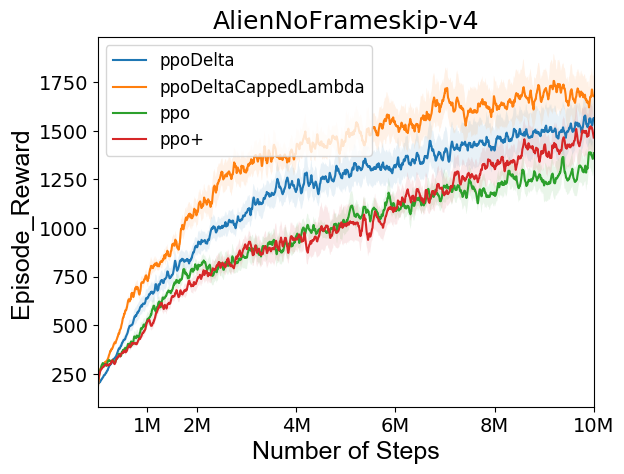}
    \includegraphics[width=.32\textwidth]{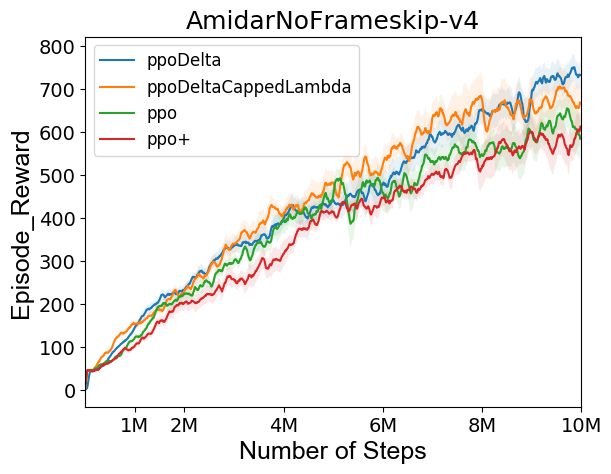}
    \includegraphics[width=.32\textwidth]{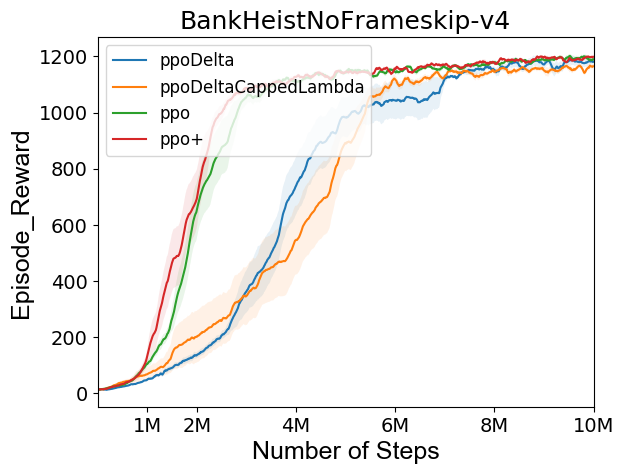}
    \includegraphics[width=.32\textwidth]{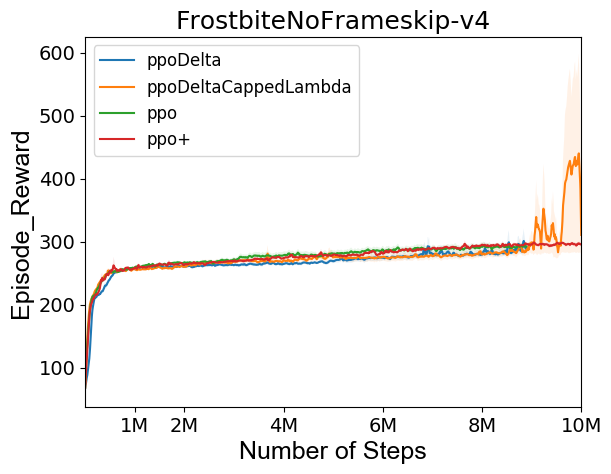}
    \includegraphics[width=.32\textwidth]{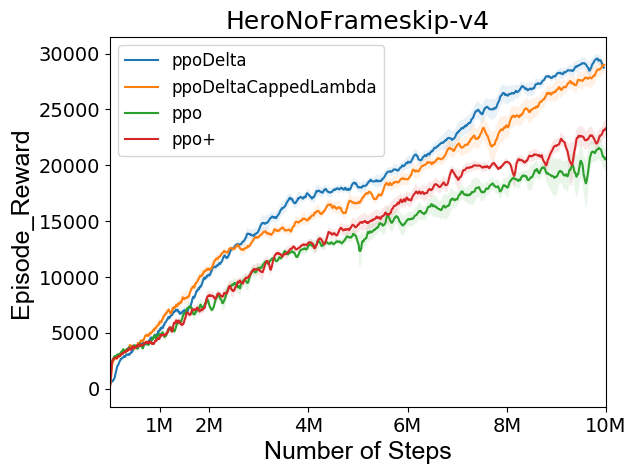}
    \includegraphics[width=.32\textwidth]{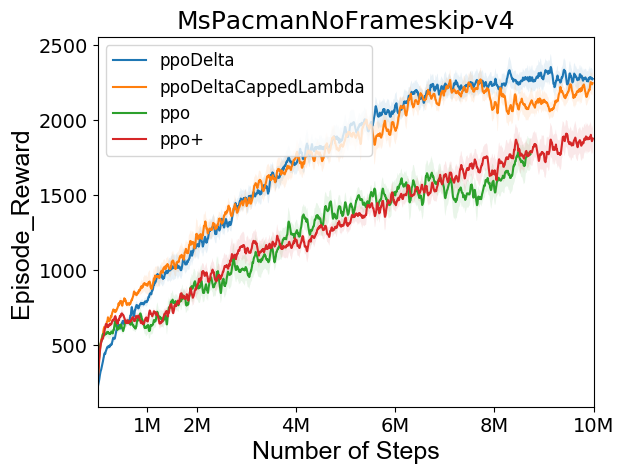}
    \includegraphics[width=.32\textwidth]{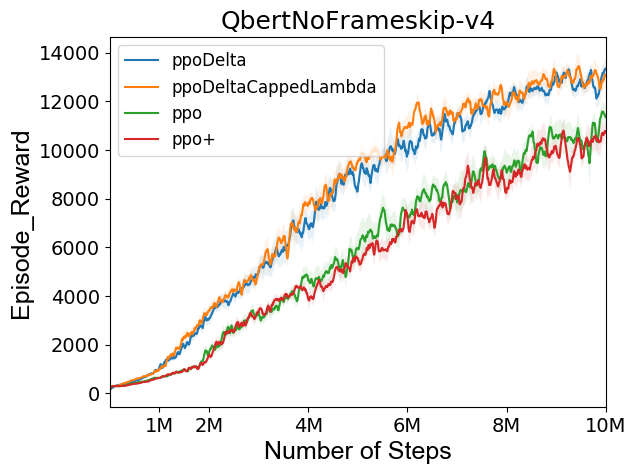}
    \includegraphics[width=.32\textwidth]{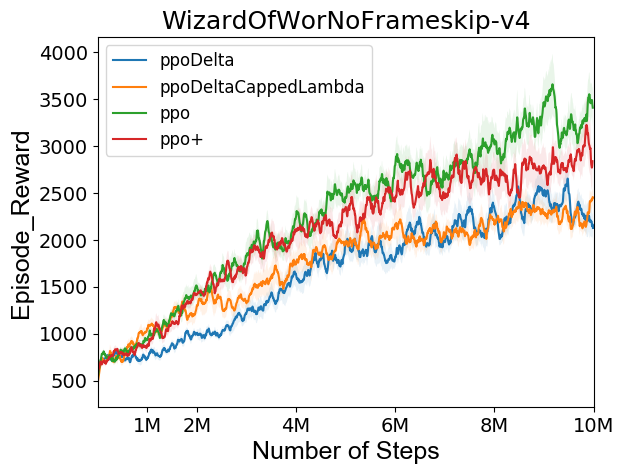}
    \includegraphics[width=.32\textwidth]{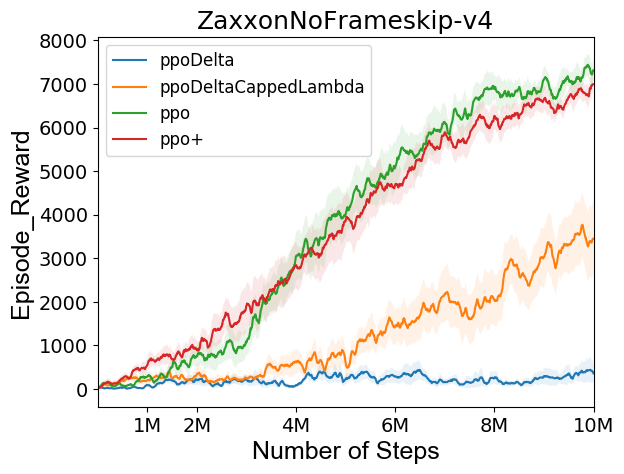}
    \caption{Performance of different TD($\Delta$) variations and baselines on all 9 Hard games with dense rewards. ppoDelta refers to setting $\gamma_z\lambda_z=\gamma\lambda$ $\forall z$. ppoDeltaCappedLambda uses the same $\gamma$s but caps all $\lambda$s at $1.0$ - introducing bias that helps in most cases. Standard error across random seeds is represented in shaded regions.}
    \label{fig:baselines_and_us}
\end{figure}

\begin{figure}[!htbp]
    \centering
    \includegraphics[width=.32\textwidth]{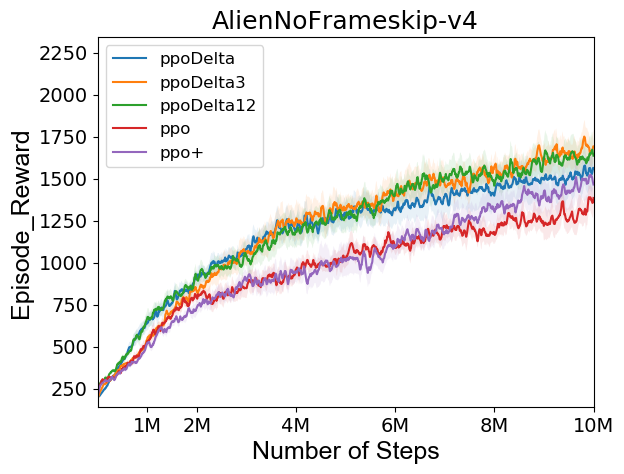}
    \includegraphics[width=.32\textwidth]{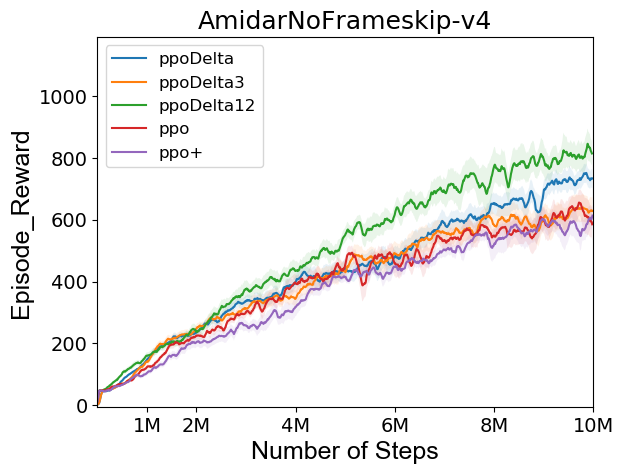}
    \includegraphics[width=.32\textwidth]{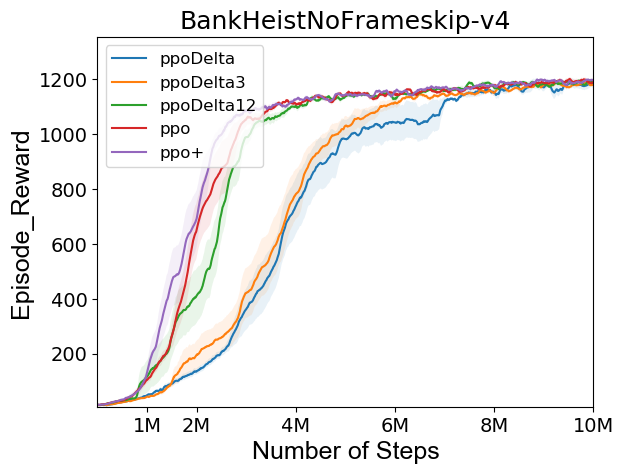}
    \includegraphics[width=.32\textwidth]{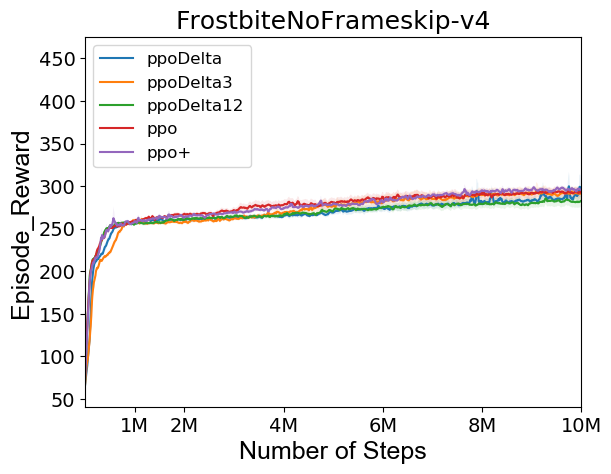}
    \includegraphics[width=.32\textwidth]{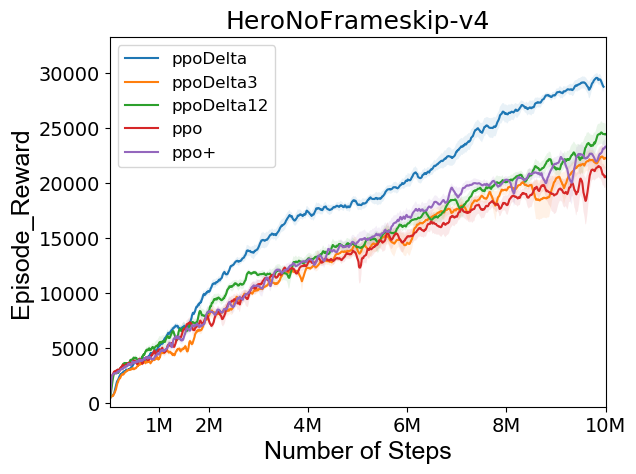}
    \includegraphics[width=.32\textwidth]{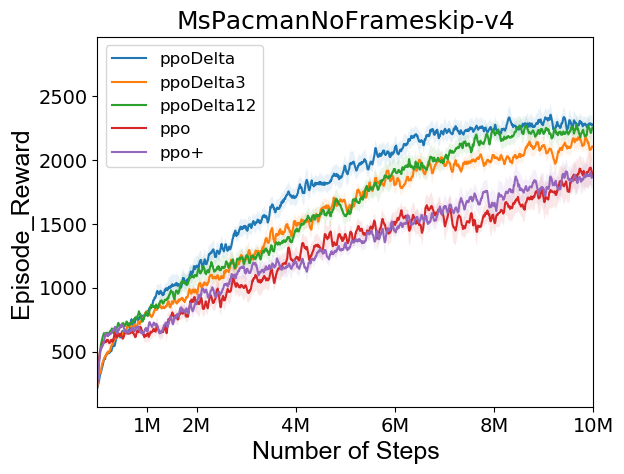}
    \includegraphics[width=.32\textwidth]{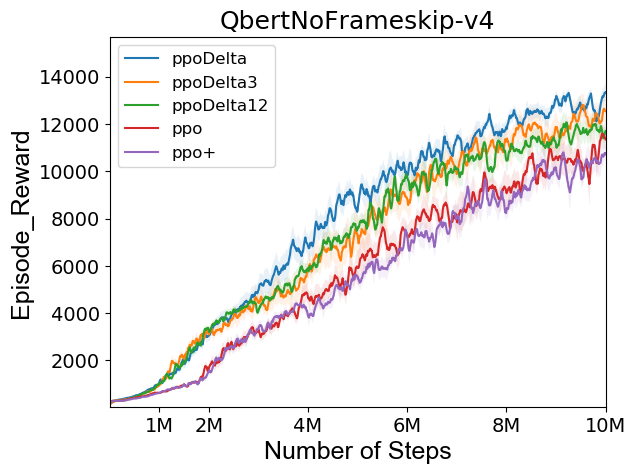}
    \includegraphics[width=.32\textwidth]{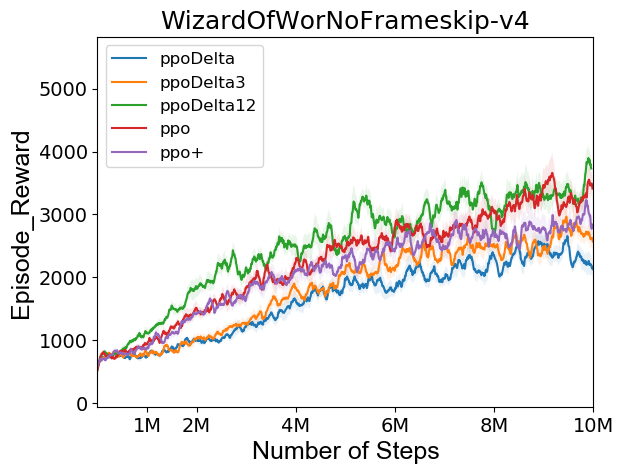}
    \includegraphics[width=.32\textwidth]{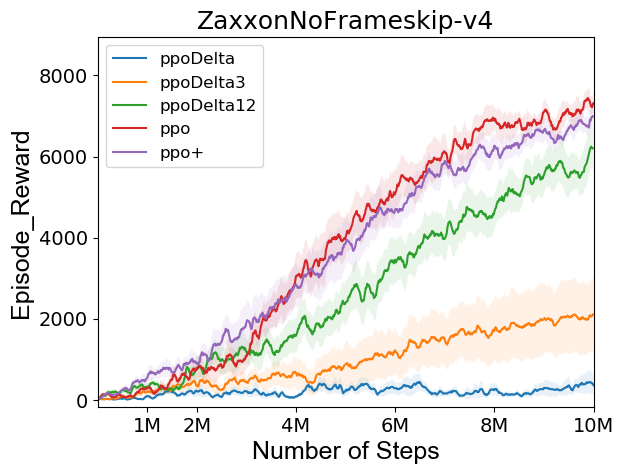}
    \caption{Performance of different TD($\Delta$) variations and baselines on all 9 Hard games with dense rewards. ppoDelta refers to setting $\gamma_z\lambda_z=\gamma\lambda$ $\forall z$. ppoDelta3 and ppoDelta12 only use two value functions - the first having a corresponding horizon of 3 and 12 respectively and the second 100. We see that bias is induced both from the number of estimators as well as the shortest horizon. Standard error across random seeds is represented in shaded regions. }
    \label{fig:baselines_and_usfull}
\end{figure}

\begin{figure*}[!htbp]
    \centering
    \includegraphics[width=.9\textwidth]{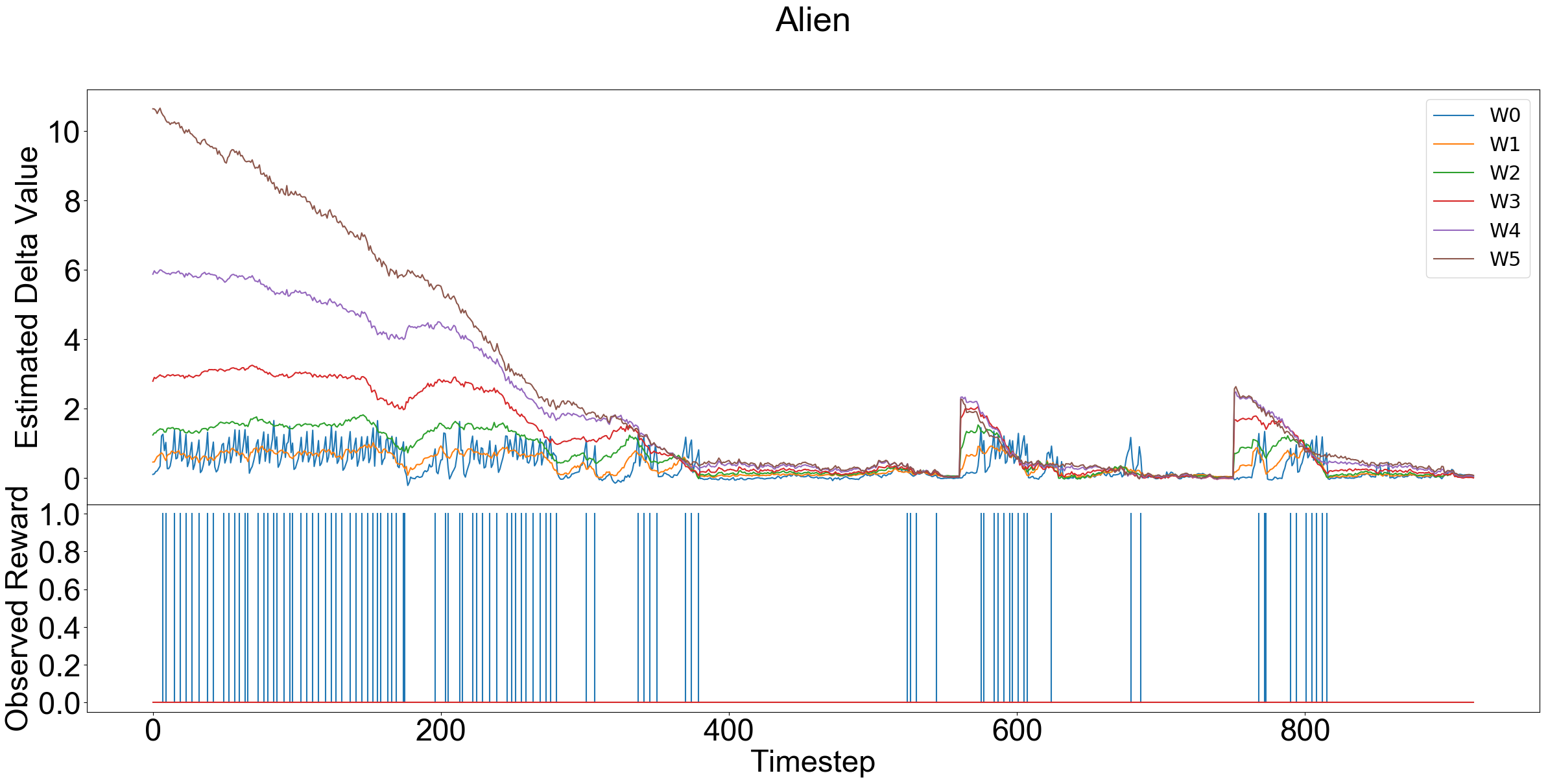}
    \includegraphics[width=.9\textwidth]{plots_valuefunction_Amidar.png}
    \includegraphics[width=.9\textwidth]{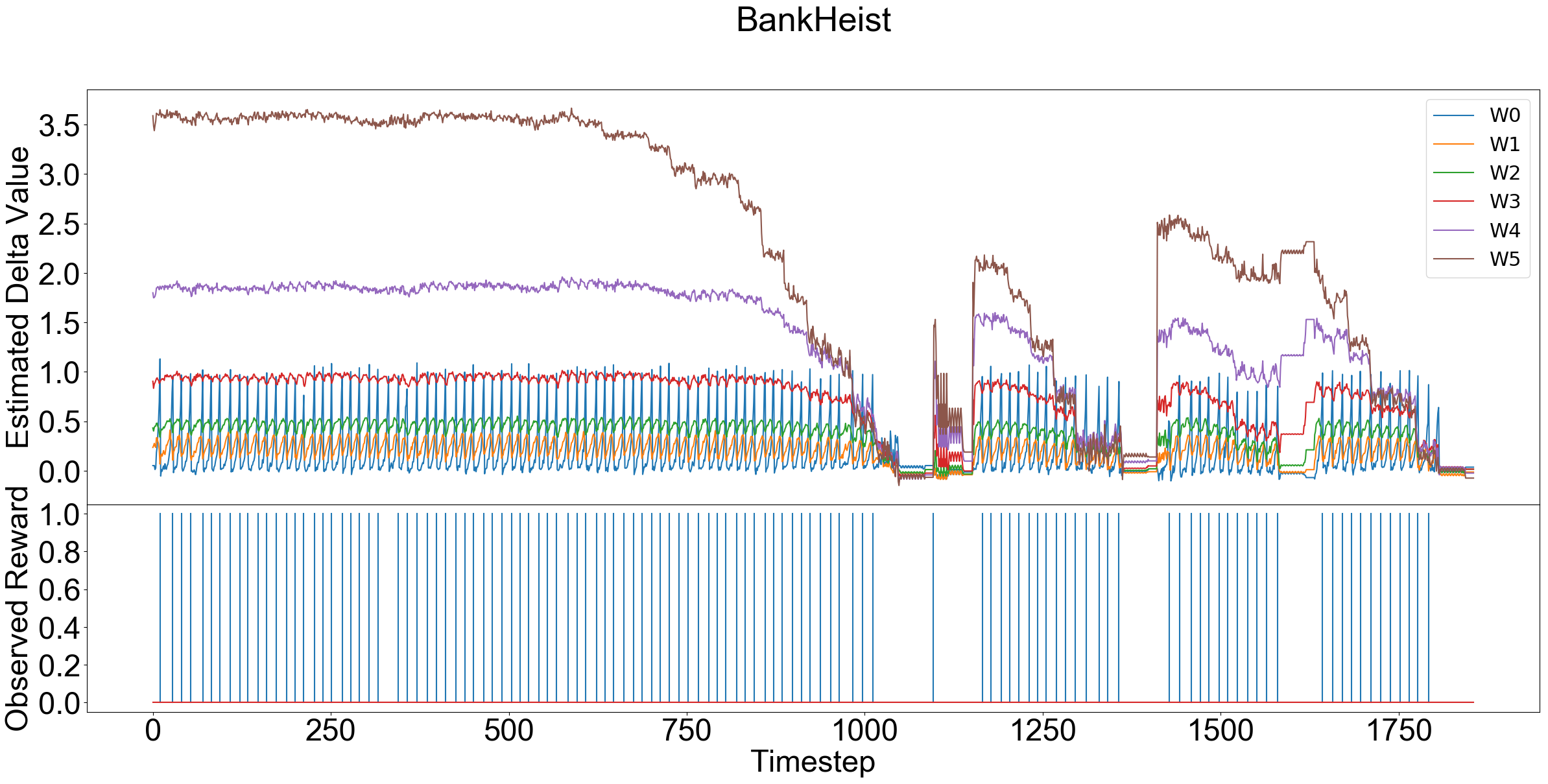}
    \caption{Each $W_z$ estimator for various games versus the reward of a policy on a single rollout trajectory.}
    \label{fig:trajectory2}
\end{figure*}

\begin{figure*}[!htbp]
    \centering
    \includegraphics[width=.9\textwidth]{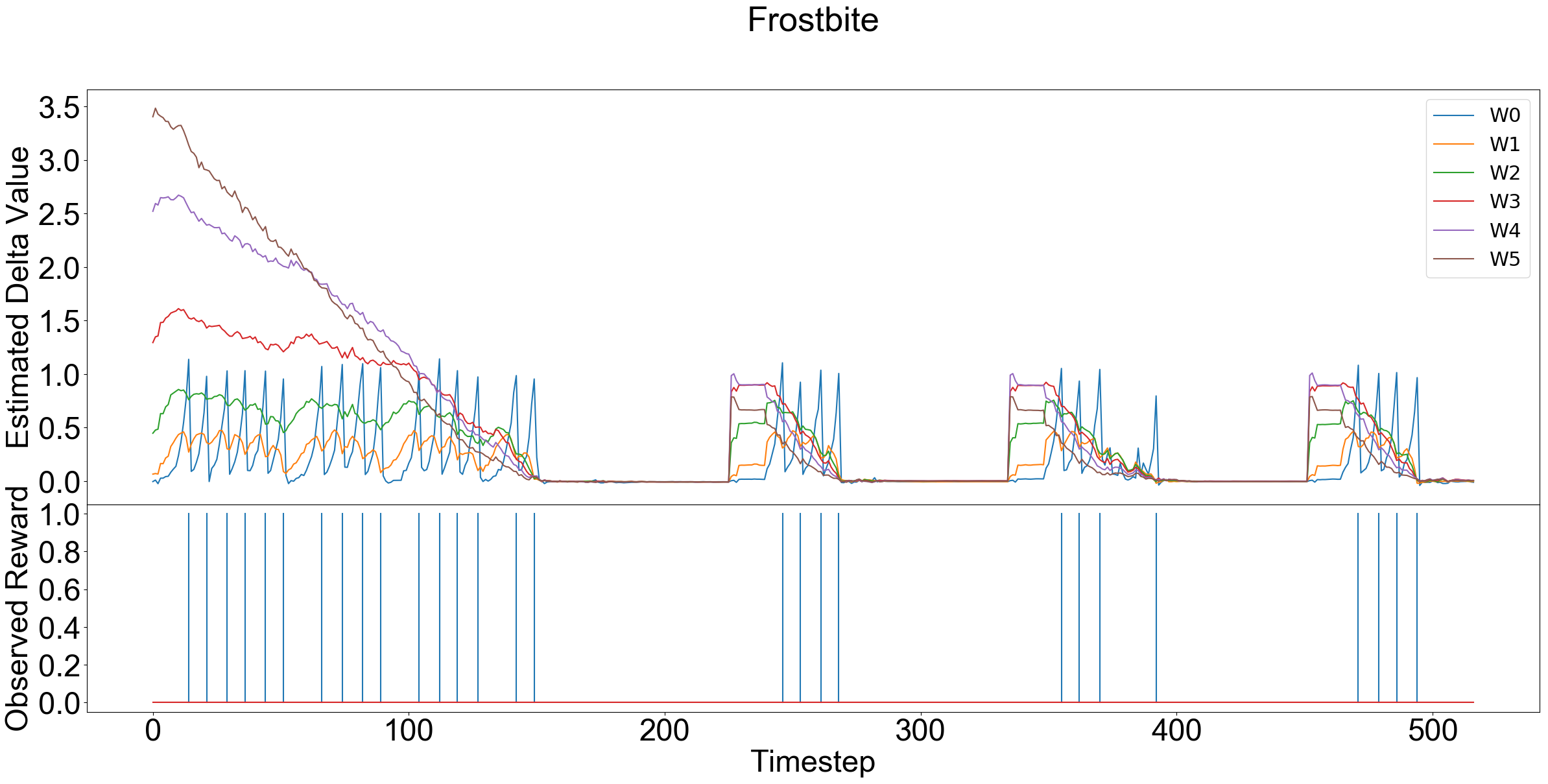}
    \includegraphics[width=.9\textwidth]{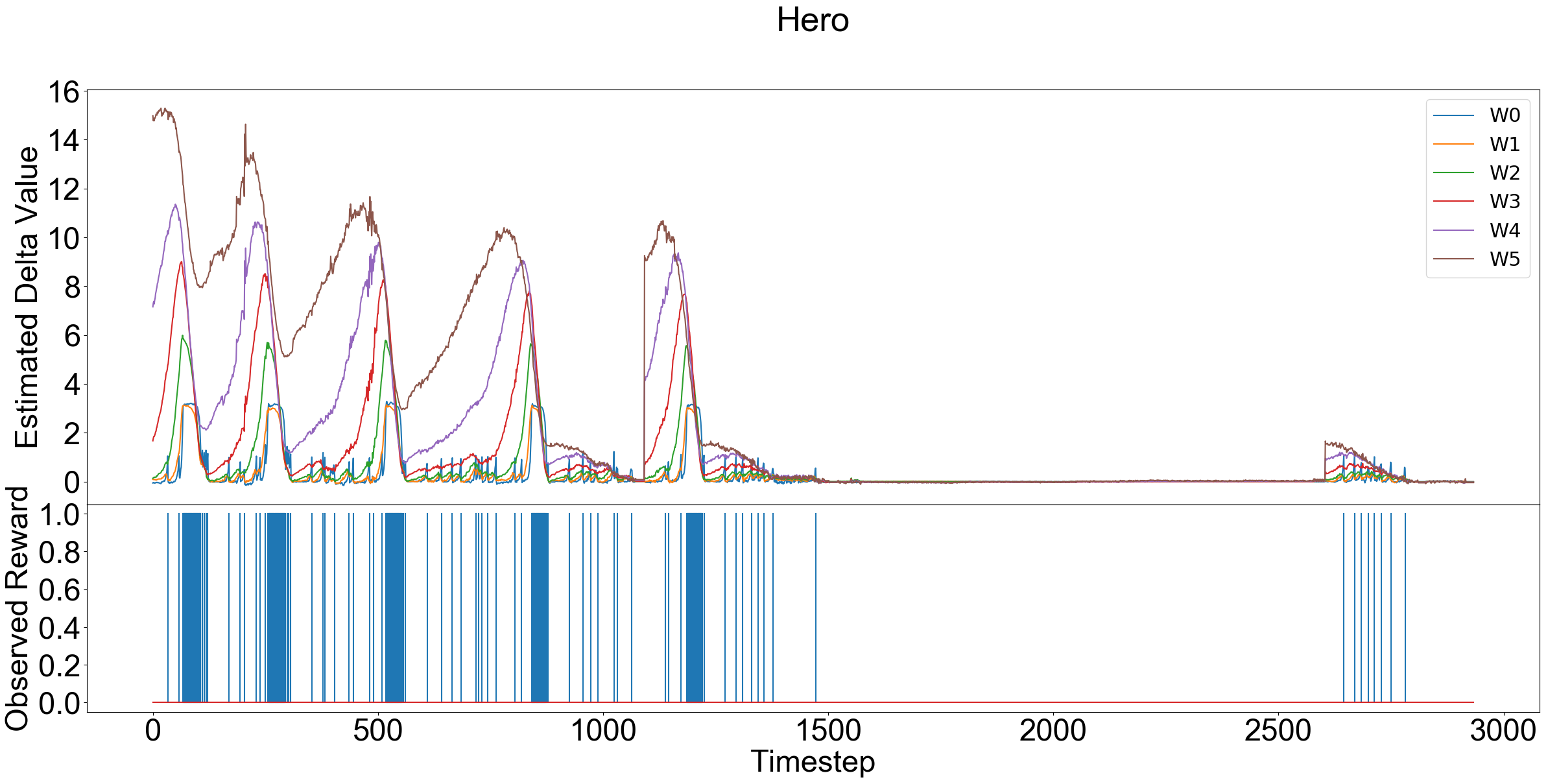}
    \includegraphics[width=.9\textwidth]{plots_valuefunction_MsPacman.png}
    \caption{Each $W_z$ estimator for various games versus the reward of a policy on a single rollout trajectory.}
    \label{fig:trajectory3}
\end{figure*}

\begin{figure*}[!htbp]
    \centering
    \includegraphics[width=.9\textwidth]{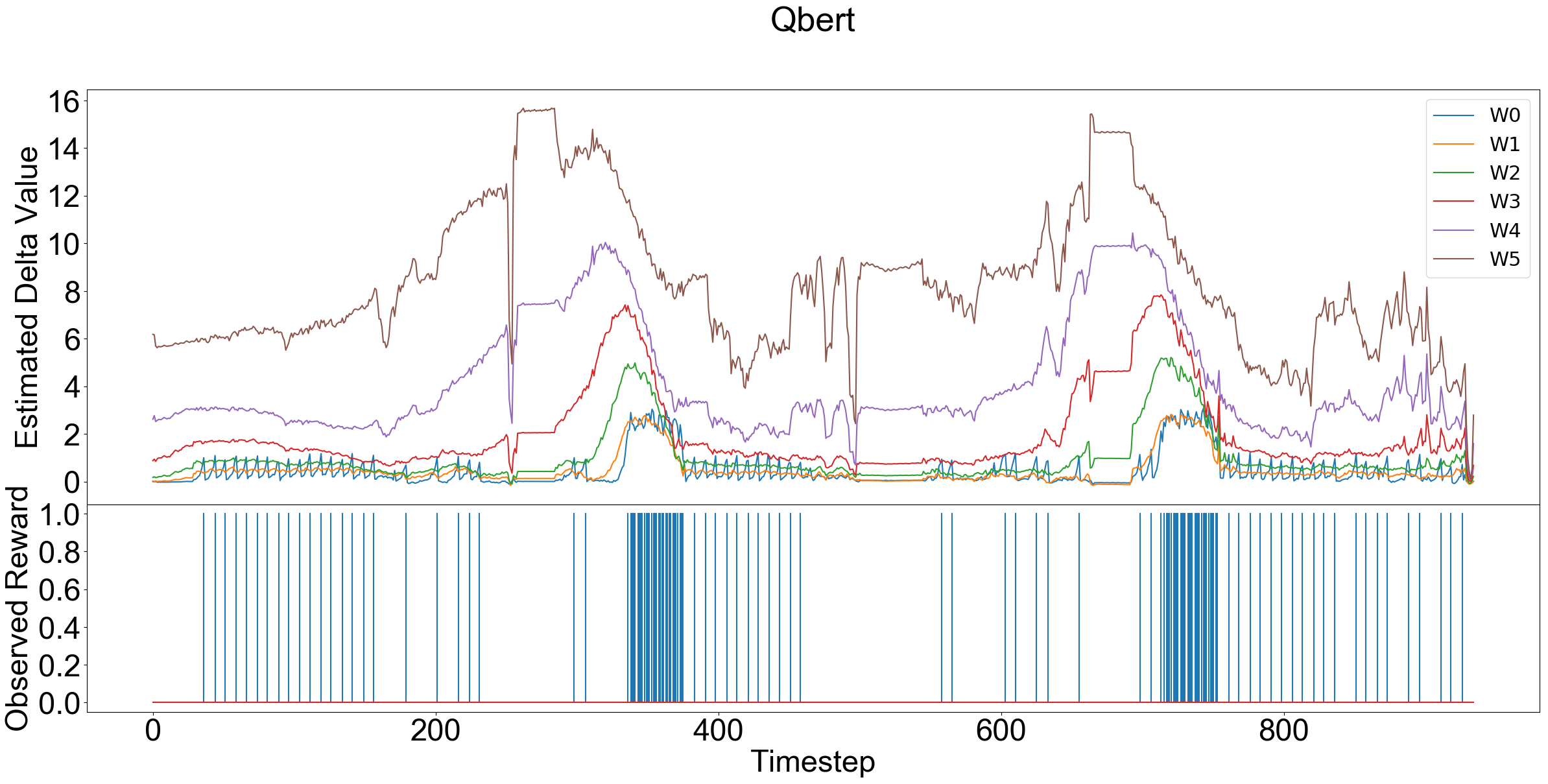}
    \includegraphics[width=.9\textwidth]{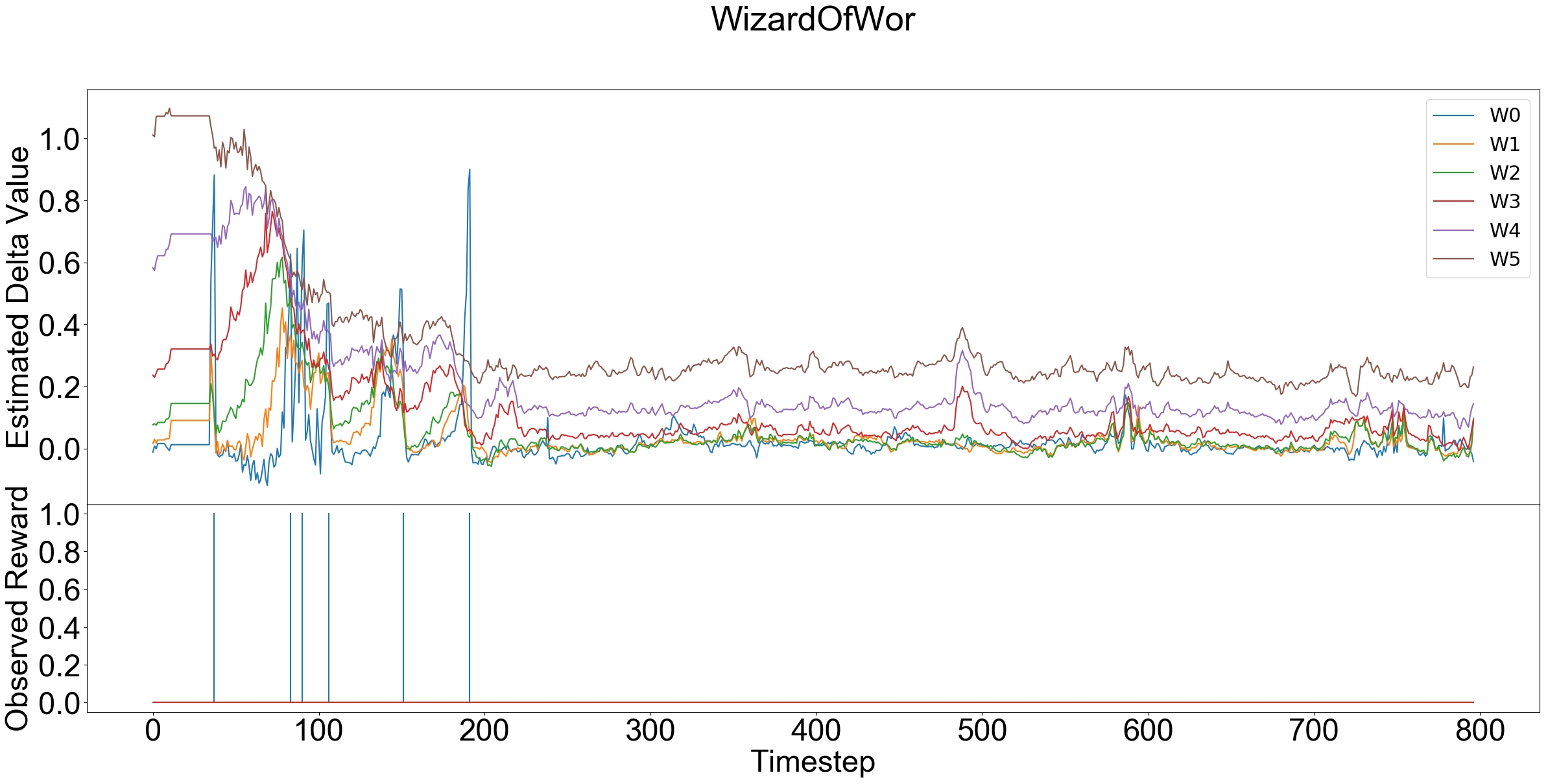}
    \includegraphics[width=.9\textwidth]{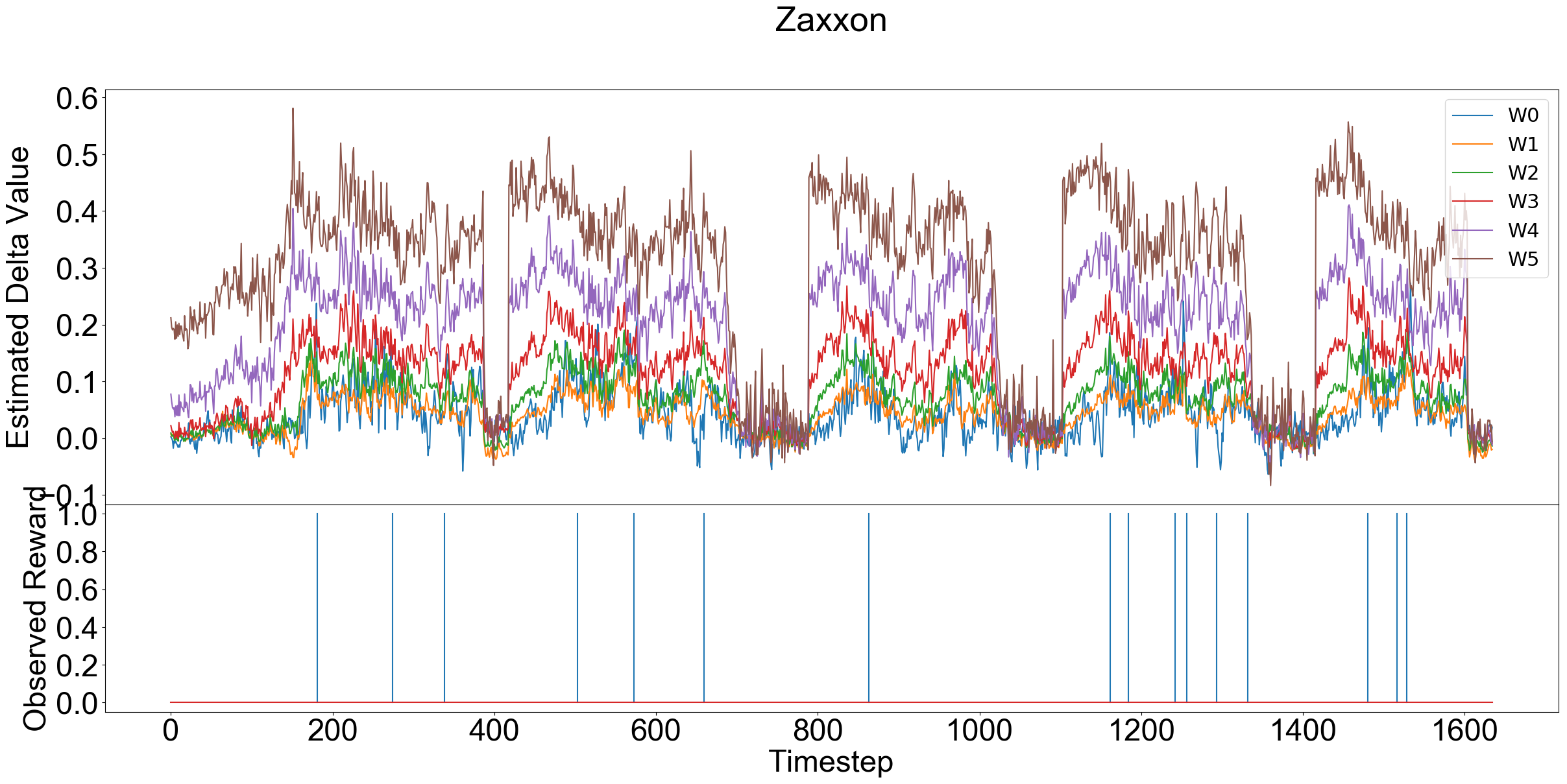}
    \caption{Each $W_z$ estimator for various games versus the reward of a policy on a single rollout trajectory.}
    \label{fig:trajectory4}
\end{figure*}

\end{document}